\documentclass[11pt,a4paper,onecolumn,thu]{thuc3i}

\usepackage[numbers,sort&compress,square]{natbib}

\usepackage{amsmath,amsfonts,amssymb}
\usepackage{algorithmic}
\usepackage{array}
\usepackage{textcomp}
\usepackage{url}
\usepackage{verbatim}
\usepackage{graphicx}
\usepackage{booktabs,makecell,multirow,tabularx,pifont}
\usepackage{fontawesome5}
\usepackage{longtable}
\usepackage{placeins}
\usepackage{etoolbox}
\usepackage{caption}
\makeatletter
\patchcmd{\LT@output}{\vss}
  {\vskip\z@\@plus1fil\@minus\normalbaselineskip}{}{}
\patchcmd{\LT@output}{\vss}
  {\vskip\z@\@plus1fil\@minus\normalbaselineskip}{}{}
\makeatother
\usepackage[linesnumbered,lined,boxed,commentsnumbered,ruled]{algorithm2e}
\usepackage{marginnote}
\usepackage{soul}
\usepackage{bigstrut}
\usepackage{diagbox}
\usepackage{stackengine}
\usepackage{siunitx}
\usepackage{cases}
\usepackage{xcolor}
\usepackage{colortbl}
\usepackage{hyperref}
\hypersetup{pdftitle={HandsOn: A Large-Scale Simulation Benchmark for Generalizable Dexterous Manipulation}}

\definecolor{palOrange}{HTML}{ED8D5A}

\definecolor{palMint}{HTML}{BFDFD2}
\definecolor{palSky}{HTML}{7BC0CD}
\definecolor{palBlue}{HTML}{4198AC}
\definecolor{palSand}{HTML}{DBCB92}
\colorlet{tabHead}{palMint!55}
\colorlet{tabBest}{palSky!50}
\colorlet{tabSolved}{palSky!28}
\colorlet{tabLow}{palOrange!25}
\colorlet{tabAside}{palMint!28}
\colorlet{tabOurs}{palSky!22}
\colorlet{tabOursHead}{palSky!50}
\makeatletter
\newcommand{\TabGapTop}{\noalign{\vskip-\belowrulesep}}
\newcommand{\TabGapBot}{\noalign{\vskip-\aboverulesep}}
\newcommand{\TabStrutTop}{\rule{0pt}{\dimexpr\ht\@arstrutbox+\belowrulesep\relax}}
\newcommand{\TabStrutBot}{\rule[-\dimexpr\dp\@arstrutbox+\aboverulesep\relax]{0pt}{\dimexpr\dp\@arstrutbox+\aboverulesep\relax}}
\makeatother
\newcommand{\TabStrut}{\TabStrutTop\TabStrutBot}
\newcommand{\TabHead}{\TabGapTop\rowcolor{tabHead}}
\newcommand{\TabHeadEnd}{\TabGapBot}
\newcommand{\TabBest}[1]{\cellcolor{tabBest}\textbf{#1}}
\makeatletter
\newcommand{\TabL}[1]{\gdef\CT@cell@color{\CT@color{#1}\@tempdimb\z@\global\let\CT@cell@color\relax}}
\newcommand{\TabR}[1]{\gdef\CT@cell@color{\CT@color{#1}\@tempdimc\z@\global\let\CT@cell@color\relax}}
\makeatother

\title{HandsOn: A Large-Scale Simulation Benchmark
for Generalizable Dexterous Manipulation}

\author{%
  Tianwei~Luo$^{1,3,*,\dagger}$, Yiming~He$^{1,*,\dagger}$, Hongjin~Wu$^{1,*}$, Ting~Pan$^{2,*}$, Debin~Ren$^{3,*}$, Haorui~Guo$^{3,*}$\\
  {\Authfont Junlin~Yang$^1$, Yuxin~Zuo$^1$, Zhehao~Kou$^1$, Kaisen~Yang$^1$, Zhuoqun~Lai$^3$, Yi~Ding$^3$}\\
  {\Authfont Jiale~Yu$^1$, Yuchen~Zhang$^4$, Zijie~Qiu$^1$, Haozhan~Li$^{1,3}$, Ning~Ding$^{1,3,\ddagger}$}\\
  {\Affilfont $^1$Tsinghua University\quad $^2$Beijing University of Posts and Telecommunications}\\
  {\Affilfont $^3$NatureWill\quad $^4$Peking University}\\
  \vspace*{2pt}
  {\Resourcefont \faEnvelope[regular]~\texttt{luotw23@mails.tsinghua.edu.cn, dingning@tsinghua.edu.cn}
   \quad\faGithub~{\fontfamily{ppl}\selectfont\textbf{Code:}}~\url{https://github.com/TsinghuaC3I/HandsOn}}
}
\role[*]{Equal contribution.}
\role[\dagger]{Project lead.}
\role[\ddagger]{Corresponding author.}
\hypersetup{pdfauthor={Tianwei Luo, Yiming He, Hongjin Wu, Ting Pan, Debin Ren, Haorui Guo, Junlin Yang, Yuxin Zuo, Zhehao Kou, Kaisen Yang, Zhuoqun Lai, Yi Ding, Jiale Yu, Yuchen Zhang, Zijie Qiu, Haozhan Li, Ning Ding}}
\setheadertext{HandsOn}
\reportnumber{}
\renewcommand{\today}{2026-10-13}
\makeatletter
\apptocmd{\ps@firststyle}{%
  \fancyhead[R]{%
    \raisebox{0pt}{\includegraphics[height=8pt]{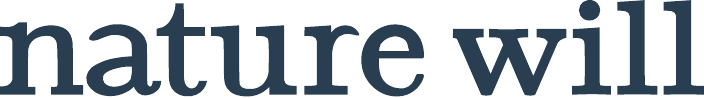}}\hspace{20pt}%
    {\footerfont\itshape\monthyeardate\today}}%
}{}{\PackageError{HandsOn}{Cannot customize the title-page header}{}}
\makeatother

\begin{abstract}
General-purpose manipulation requires reliable behavior across embodiments, objects, and task demands.
We introduce HandsOn, a composable simulation platform and benchmark for cross-embodiment dexterous manipulation.
It supports five robotic arms, six dexterous hands, and approximately 200 rigid, articulated, and deformable assets in randomized scenes.
The benchmark pairs 18,000 curated human demonstrations with 66 human-verified tasks: Core-50 supports demonstration-based learning and execution tests, while Transfer-16 evaluates asset and relation transfer, skill composition, and deformable manipulation without target-task training demonstrations.
Qualitative real-robot rollouts illustrate sim-to-real transfer.
Across diffusion, vision-language-action, world-action, and prompted VLM-based methods, the best trained multi-task policy achieves 29.1\% Core-50 success.
Without benchmark finetuning, GPT-6 using \texttt{move\_eef} for palm-and-finger control achieves 5.2\% on Core-50 and 1.9\% on Transfer-16 over 660 trials.
Trained-policy stage diagnostics expose failures in contact establishment, action coordination, and maintaining achieved goals, with additional sequencing errors in composed tasks.
HandsOn provides a shared basis for investigating manipulation generalization and the limits of reliable physical interaction.\looseness=-1
\end{abstract}

\begin{document}
\maketitle

\par\smallskip
\noindent\begin{minipage}{\linewidth}
  \centering
  \includegraphics[width=\linewidth]{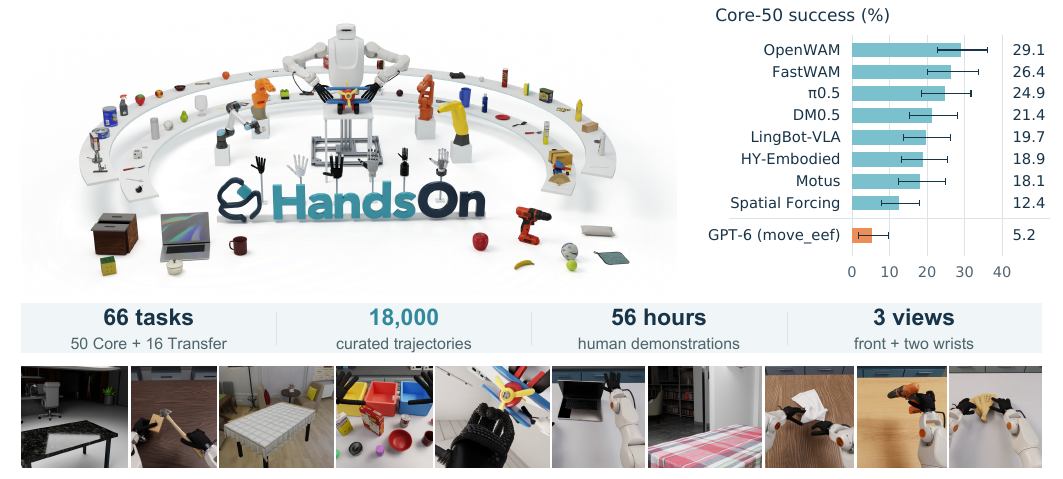}
  \captionsetup{font=small,skip=4pt,hypcap=false}
  \captionof{figure}{\textbf{Overview of HandsOn.} Representative assets,
  environments and interactions accompany 66 tasks and 18,000 curated
  demonstrations (56 hours; three synchronized views).
  Right: Core-50 success of eight trained checkpoints (100 trials/task)
  and GPT-6 with \texttt{move\_eef} (ten). Bars: 95\% task-and-episode bootstrap
  intervals for every displayed method; conditions differ
  (Section~\ref{sec:gpt6-results}). Poses and cables are illustrative,
  not control-compatibility evidence; the caf\'e background is panoramic.}
  \label{fig:overview}
\end{minipage}\par

\clearpage
\begingroup
\setcounter{tocdepth}{3}
\renewcommand{\contentsname}{Contents}
\tableofcontents
\endgroup
\clearpage

\section{Introduction}
\label{sec:introduction}
Dexterous manipulation requires controlling how contacts change during an
operation. Moving a mouse must not trigger a premature click; setting two
knobs requires both knobs to remain at their target settings simultaneously;
and placing a lid requires preserving alignment while releasing it. Evaluating
these capabilities requires demonstrations of varied interactions and diagnostics
that distinguish initial progress from reliable functional completion.
Recent dexterous benchmarks broaden task and embodiment coverage
\cite{wang2026dexjoco,yao2026dexverse,bench2dex,labdex}, but their
human-demonstration sets are small and their tasks rarely include ones
reserved for evaluation.

HandsOn provides 18,000 curated simulation demonstrations, approximately
56 hours of interaction with three synchronized camera views. This is
5.7--16.4 times the human trajectory count of representative dexterous
benchmarks under the convention in Table~\ref{tab:benchmark-comparison}.
High-fidelity re-rendering produces varied observations of recorded
behavior while preserving state/action supervision. Visual domain
randomization and meaning-preserving prompt randomization broaden
appearance and language coverage around task-specific configurations.
Figure~\ref{fig:overview} summarizes the asset collection, environment
and interaction diversity, demonstration scale, and Core-50 results.
Figure~\ref{fig:protocol} connects this data pipeline to evaluation and
illustrates correspondence with the physical system.

\par\addvspace{\intextsep}
\noindent\begin{minipage}{\linewidth}
  \centering
  \includegraphics[width=\linewidth]{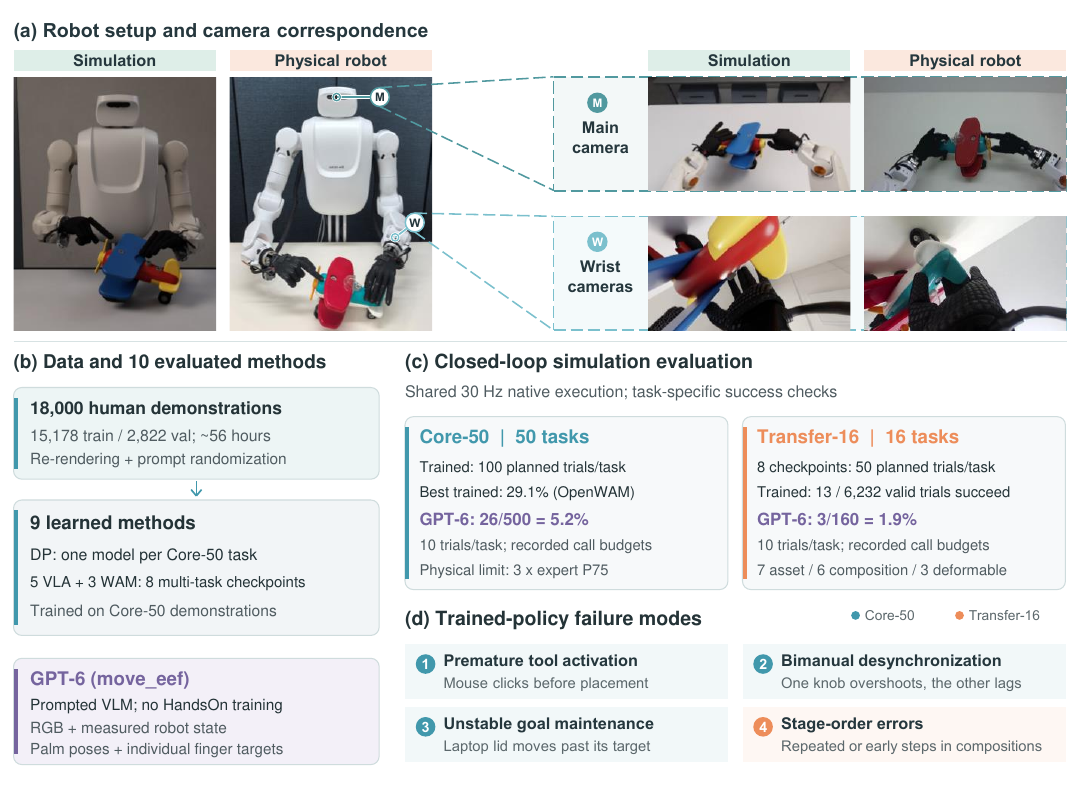}
  \captionsetup{hypcap=false}
  \captionof{figure}{Platform and evaluation protocol. (a) Toy-aircraft views:
  setup (not a policy input), wrist (W), and main (M). Real images show
  appearance; real rollouts are qualitative videos. (b) Nine trained methods
  and GPT-6 (\texttt{move\_eef}); GPT-6 uses no HandsOn training.
  (c) Separate suite-specific counts and results.
  (d) Trained-policy failures (Section~\ref{sec:main-failures}).}
  \label{fig:protocol}
\end{minipage}\par\addvspace{\intextsep}

The benchmark comprises two complementary suites: \textbf{Core-50},
50 tasks with demonstrations for policy learning and fresh-rollout evaluation,
and \textbf{Transfer-16}, 16 tasks reserved for generalization evaluation.
Core-50 tasks combine functional and numerical understanding,
articulation, spatial relations, precise contacts, and bimanual execution
(Figure~\ref{fig:taxonomy}). Transfer-16 tests new asset relations and
skill combinations, including three deformable interactions without
deformable-task demonstrations in this dataset. The simulated embodiment
follows a physical platform, and real-robot rollouts of
simulation-trained checkpoints are shown in the accompanying videos.

Our Core-50 study compares ten methods: five VLA models, three WAM models,
single-task Diffusion Policy, and the prompted VLM controller GPT-6.
The best trained multi-task score is 29.1\%. Across the trained checkpoints,
tool use and precision fits are nearly unsolved, and skill-group gaps persist after
adjusting for demonstration count and expert duration. Two VLA checkpoints
barely condition on which object the instruction names. Failures concentrate
where closing the grasp fires a button, where both hands must hold their
targets together, and where reached goals must be kept. These findings
motivate evaluating what happens after acquisition, alongside aggregate
success.

We additionally study GPT-6 as a prompted multimodal controller, using
RGB observations, measured robot state, and the \texttt{move\_eef} palm-and-finger
control interface without HandsOn finetuning. This supplementary experiment
achieves 5.2\% success on Core-50 and 1.9\% on Transfer-16 over 660 trials.
Success appears on only 12 of the 66 tasks. Qualitative action traces
suggest task-level planning potential and motivate richer interfaces
between that planning and high-dimensional dexterous execution. Its
observation interface, trial count, and model-call budgets are disclosed
alongside the trained-policy comparison.

Our contributions are:
\begin{enumerate}\setlength{\itemsep}{1pt}\setlength{\parskip}{0pt}
  \item A large human-demonstration dataset, re-rendered with visual and
  prompt randomization.
  \item Core-50 and Transfer-16: 50 demonstration-backed tasks and 16
  evaluation-only asset, skill-composition, and deformable tasks, scored by
  native stage checkers.
  \item A comparison of DP, five VLA, and three WAM policies with a
  covariate-adjusted skill analysis and a language-conditioning probe,
  locating failures at three hand transitions and, in compositions, at
  step order and completion, together with a 660-trial GPT-6 \texttt{move\_eef}
  evaluation without benchmark finetuning.
\end{enumerate}

\section{Related Work}
DexJoCo, DexVerse, Bench2Dex, and LabDex cover functional and bimanual
manipulation with diverse embodiments
\cite{wang2026dexjoco,yao2026dexverse,bench2dex,labdex}.
DexMimicGen expands human seeds through automated trajectory generation,
whereas SPD studies simulation pretraining and real-data adaptation
\cite{dexmimicgen,kamat2026spd}. Table~\ref{tab:benchmark-comparison}
compares data scale, randomization, policy coverage, and evaluation protocols.
HandsOn evaluates DP, five VLA models, three WAM models, and a prompted
VLM controller, GPT-6 with \texttt{move\_eef}, across a demonstrated task
suite and a separate dexterous task-transfer suite.

LIBERO, CALVIN, DexArt, RoboDojo, and RoboTwin motivate transfer,
capability evaluation, and randomization
\cite{liu2023libero,mees2022calvin,bao2023dexart,chen2026robodojo,chen2025robotwin2}.
Earlier dexterous suites learn mainly by reinforcement or from small
demonstration sets \cite{rajeswaran2017adroit,chen2022bidexhands,liu2025vtdexmanip};
large simulated and real manipulation corpora are dominated by grippers
\cite{gu2023maniskill2,nasiriany2024robocasa,li2022behavior1k,oxe2024,khazatsky2024droid}.
Dexterous teleoperation and motion capture \cite{qin2023anyteleop,iyer2024openteach,wang2024dexcap}
collect human data outside a benchmark; HandsOn records it in the
evaluation simulator. Dexterous sim-to-real learning precedes HandsOn
\cite{openai2018dactyl,handa2023dextreme}.
Section~\ref{sec:dataset-identity} describes the dataset counting conventions.
Appendix~\ref{sec:sim_real_complementarity} presents qualitative real-robot
rollouts, and Appendix~\ref{app:physical-transfer} specifies the proposed
matched evaluation protocol.

\begin{table}[t]
  \centering\footnotesize\setlength{\tabcolsep}{4pt}
  \renewcommand{\arraystretch}{1.22}
  \caption{Dexterous learning resources (DMG: DexMimicGen; NR: not reported).
  Human and generated trajectories
  are counted separately; catalogs and evaluation scopes differ; HandsOn shaded.\looseness=-1}
  \label{tab:benchmark-comparison}
  \begin{tabularx}{\linewidth}{@{}l*{5}{>{\raggedright\arraybackslash\hsize=.94\hsize}X}>{\columncolor{tabOurs}[\tabcolsep][0pt]\raggedright\arraybackslash\hsize=1.3\hsize}X@{}}
    \toprule
    \TabHead\TabStrut\TabL{tabHead}& DexJoCo & DexVerse & Bench2Dex & DMG & SPD & \TabR{tabOursHead}\textbf{HandsOn}\\\TabHeadEnd
    \midrule\TabGapTop
    \TabStrutTop Task catalog & 11 & 100 & 26 & 9 & 6 scenes & \textbf{50 + 16}\\
    Human demos & 1,100 & 3,180 & $\sim$1,300 & 60 & $\sim$2,000 & \textbf{18,000}\\
    Generated demos & --- & --- & --- & 21,000 & --- & ---\\
    Language variation & Digits / words & NR & NR & NR & No input & \textbf{32 / 8 per goal}$^a$\\
    Test-only tasks & NR & NR & NR & NR & NR & \textbf{16 tasks}\\
    Transfer setting & Visual; language & Visual; embodiment & Visual / geometric & Initial states; objects & Real-task finetuning & \textbf{Assets; skills; soft objects}\\
    Policy coverage & ACT, DP, VLA & DP, 3D DP, VLA & ACT, DP, VLA & BC-RNN, DP & Diffusion transformer & \textbf{DP; 5 VLA; 3 WAM; GPT-6}$^c$\\
    Extra diagnostics & Failures; grounding & Category SR & Stages; time; safety; grasp & Generation yield & Stage progress & \textbf{Transitions; retention; contact}\\
    Real evaluation & NR & NR & NR & Real demos via sim\TabStrutBot & Real finetuning\TabStrutBot & Qualitative$^b$\TabStrutBot\\\TabGapBot
    \bottomrule
  \end{tabularx}
  \par\smallskip\raggedright
  $^a$32 train / 8 held-out paraphrases for each of 91 task--goal pairs;
  separate from Transfer-16. $^b$Recorded real-robot rollout videos; no physical success rate.
  $^c$GPT-6 is a prompted VLM controller using \texttt{move\_eef}, with
  ten trials/task on both suites and no HandsOn demonstrations or finetuning;
  its inputs and call budgets differ from the trained-policy conditions.
  Extra views do not add human demonstrations.
  Dataset counting conventions: Section~\ref{sec:dataset-identity}.
  Comparisons use the cited sources and their reported evaluation scopes.
\end{table}

\section{The HandsOn Benchmark}
\label{sec:benchmark}

HandsOn has two suites. \textbf{Core-50} provides demonstrations for policy
learning and fresh-rollout evaluation; \textbf{Transfer-16} is reserved for
generalization evaluation and contributes no target-task demonstration to
the training set. Both suites are scored by native, task-specific success
predicates rather than by a shared proxy, and every task was verified in
person by the data-collection operators before release
(Section~\ref{sec:task-design}).

\begin{figure}[t]
  \centering
  \includegraphics[width=\linewidth]{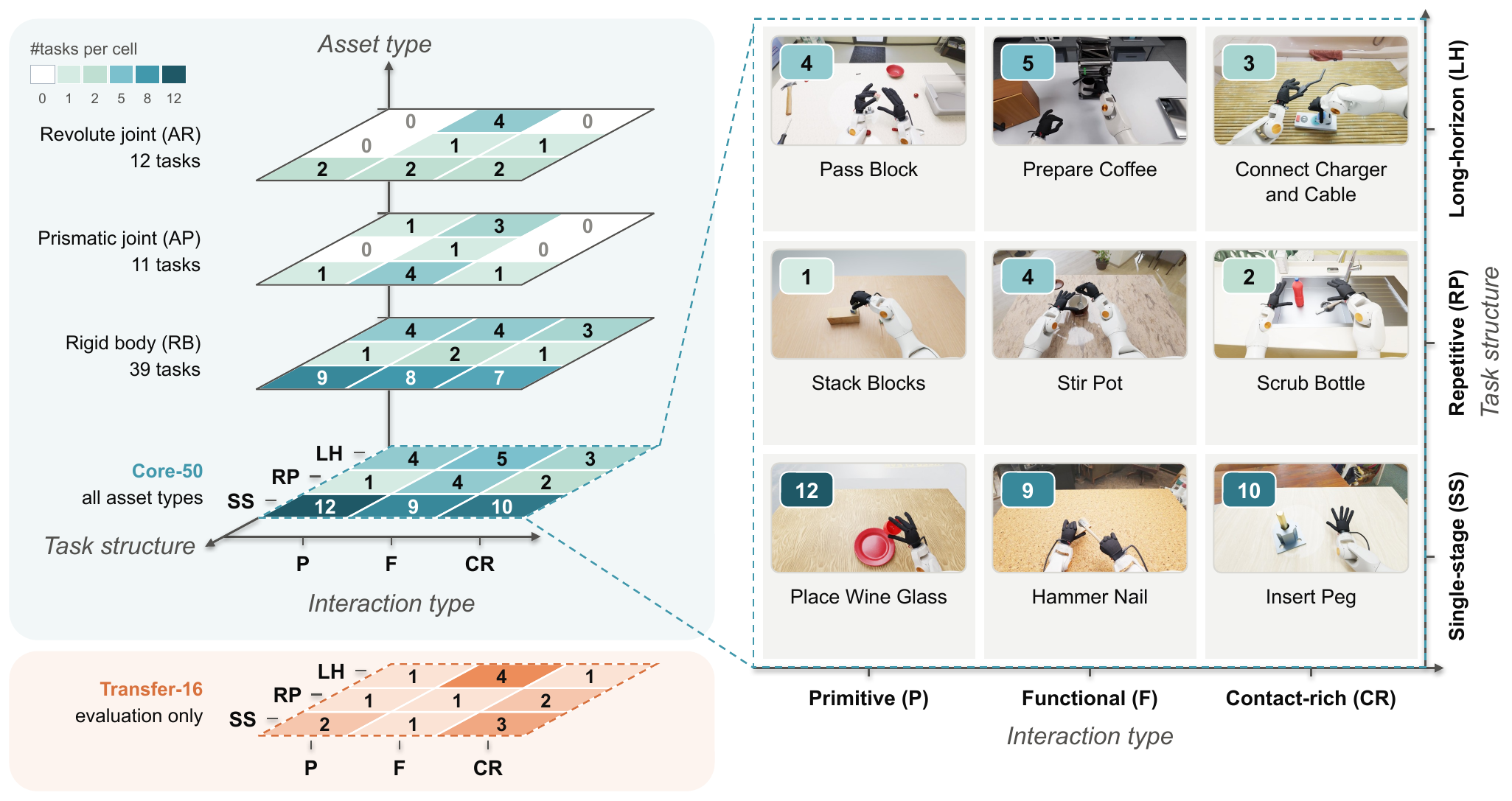}
  \caption{Three-dimensional task taxonomy of HandsOn. Tasks are classified
  by interaction type (primitive, functional, contact-rich), task structure
  (single-stage, repetitive, long-horizon), and asset type (rigid bodies,
  assets with revolute joints, assets with prismatic joints, deformable
  objects). A task may have multiple asset-type labels. Each cell gives the
  number of tasks at the intersection of the indicated categories. The right
  panel illustrates the nine interaction--structure combinations in Core-50;
  badges show the corresponding task counts. Evaluation-only Transfer-16
  tasks are shown separately in orange, with deformable manipulation reserved
  for this suite. Table~\ref{tab:full-task-inventory} provides the full
  per-task taxonomy.\looseness=-1}
  \label{fig:taxonomy}
\end{figure}

\subsection{Core-50 and Transfer-16}
\label{sec:suites}
Core-50 tasks combine functional and numerical understanding, articulation,
spatial relations, precise contacts, and bimanual execution
(Figure~\ref{fig:taxonomy}). Transfer-16 tests new asset relations and skill
compositions, including three deformable interactions for which this dataset
contains no deformable-task demonstrations. Because the two suites answer
different questions, we report them separately throughout: a Core-50 score
measures execution on tasks with demonstrations, whereas Transfer-16 tests
held-out tasks. The learned multi-task methods train on Core-50 before
both evaluations; GPT-6 is evaluated on both suites without HandsOn
demonstrations or finetuning.

\begin{table}[!hbp]
  \centering\small
  \caption{The two HandsOn suites. Core-50 carries demonstrations;
  Transfer-16 is evaluation-only.}
  \label{tab:suites}
  \begin{tabularx}{\linewidth}{@{}lXX@{}}
    \toprule
    \TabHead\TabStrut\TabL{tabHead} & \textbf{Core-50} & \TabR{tabHead}\textbf{Transfer-16}\\\TabHeadEnd
    \midrule
    Purpose & Policy learning and execution evaluation &
      Asset, relation, and skill-composition generalization\\
    Tasks & 50 & 16 (6 EVAL-Comb, 7 EVAL-Gen, 3 EVAL-Flex)\\
    Demonstrations & 46--463 training trajectories per task
      (median 268) & None used for training; 13 tasks use a
      described duration proxy\\
    Scope & Functional tool use, numerical and spatial grounding,
      articulation, bimanual manipulation & New asset relations, skill
      composition, three deformable interactions\\
    Scoring & Task-macro success & Task-macro success and ordered
      progress (trained multi-task methods); success only for GPT-6\\
    \bottomrule
  \end{tabularx}
\end{table}

\FloatBarrier
\subsection{Task Design and Feasibility Verification}
\label{sec:task-design}
Tasks and demonstrations pass a written collection-and-review procedure with
separated roles: data-collection operators record and self-check episodes,
independent reviewers judge them, and a third person arbitrates
disagreements; nobody reviews their own data. Each task has a task card that
fixes the goal, the standard initial and final states, completion evidence
independent of the program's success flag (object poses, joint states, or
images), allowed and forbidden actions with known shortcuts, the time limit,
and the environment, task, and checker versions with their code hashes.

A new task or major revision is first piloted with at least 20 operator
episodes, all reviewed by hand and covering success, failure, time-out,
reset, and boundary states; it enters collection only with no severe defect,
at least 98\% agreement between labels and the objective evidence, and
unambiguous rules. During collection the environment and checker are frozen
by hash. A checker change passes reproduction, impact analysis, two-person
review, regression tests on normal, failing, boundary, and known-defect
cases, and a new fully reviewed pilot; affected episodes are replayed,
relabeled, or quarantined. Every recorded episode then passes automatic
checks of completeness, version hashes, action plausibility, agreement
between the success flag and the final physical state, and duplication.
Stratified random blind review covers about 20\% of each batch (all of the
first batch of a new task or version), and a further 5--10\% of passed
episodes are blindly re-reviewed. A success flag without objective completion
quarantines the task as a suspected checker defect until it is repaired and
piloted again.\looseness=-1

This procedure matters for how the results should be read. Every Core-50 and
Transfer-16 task was verified in person by the data-collection operators in
the released environment before evaluation. Each Core-50 task is also
completed by its curated demonstrations. For 47 of the 50 Core-50 tasks,
at least one evaluated multi-task checkpoint succeeds in at least one episode.
Human verification establishes task feasibility under the verified
configurations. The low policy success rates indicate substantial
performance gaps under the evaluated protocol
(Section~\ref{sec:main-failures}). Sources adapted from prior
benchmarks, and the modifications made to them, are recorded in
Table~\ref{tab:task-provenance}.

\subsection{Evaluation Protocol}
\label{sec:evaluation}
Simulation uses the fixed native end-effector execution protocol. Learned
policies retain checkpoint-owned representations, preprocessing, and
normalization; GPT-6 uses the adapter in Appendix~\ref{app:gpt6-motion}. Actions
are executed at 30~Hz, with each action advancing the simulation by exactly
$1/30$~s; integration steps are recorded separately. The small source cohort
collected with another controller is identified rather than assumed
equivalent.

\paragraph{Time budgets.}
Physical-time budgets are fixed before evaluation and shared across methods;
GPT-6 additionally has the recorded model-call budgets described below. For a
Core-50 task $i$, let $F_{i,(k)}$ denote its demonstration lengths in frames,
sorted in ascending order with zero-based index $k$. The frozen duration
table uses the non-interpolated quantile
\begin{equation}
 q_i=F_{i,(\min(n_i-1,\lfloor0.75n_i\rfloor))},\qquad
 H_i=3q_i/30,\qquad B_i=\lceil30H_i\rceil.
 \label{eq:time-budget}
\end{equation}
Here $H_i$ is the budget in simulated physical seconds and $B_i$ the maximum
number of executed actions. Source demonstrations are recorded at 30~fps and
evaluation advances exactly $1/30$~s per action, verified from the physics
timestep and substeps. Inference latency, rendering time, and action-chunk
length do not redefine this physical budget. The released table is computed
from the curated package's source cohorts, including its trajectory
validation partition, without using evaluated policy outcomes.

For the 13 non-Flex Transfer-16 tasks no target-task demonstration
distribution exists. We instead use
$H_i=3\sum_{j\in\mathcal{S}_i}q_j/30$, where $\mathcal{S}_i$ is a
predeclared mapping to supporting atomic tasks. This is a duration proxy,
not the measured P75 of composed trajectories. For example,
\emph{Open Laptop and Click Mouse} uses $(196+125)$ source frames, giving
32.1~s or 963 actions. The three Flex tasks retain explicit native budgets:
240~s for sponge compression and 180~s each for towel folding and pillow
packing. Table~\ref{tab:eval-budgets} lists all Transfer-16 budgets and their
mappings.

\paragraph{Stopping and success.}
After every executed action the native environment checks success,
task-defined failure, and timeout. The rollout stops at the first terminal
outcome, including within an action chunk; unused predicted actions are
discarded. Success requires the complete native predicate, including any
release, stability, or dwell requirement. Native failure terminates with zero
success; exhausting the budget without success also scores zero. Success
reached on the final permitted action is accepted. There is no generic early
stop for apparent stagnation or for completing only an intermediate stage.
Task-specific irreversible violations retain their documented native rules.

The GPT-6 \texttt{move\_eef} experiment also uses a model-call budget,
which can end an unsuccessful trial before its physical-time limit.
Its recorded context and budget settings, outcome acceptance, and
comparability are described in Appendix~\ref{app:gpt6-protocol}.
This prompted controller uses no HandsOn finetuning and is reported
separately from the trained-checkpoint aggregates.

\paragraph{Validity and attribution.}
An invalid policy output, such as a nonfinite action, is recorded as a policy
failure under the declared adapter. Simulator defects, rejected resets,
process interruption, and infrastructure failures are recorded separately
and do not become numerical policy failures. Wall-clock watchdogs bound
infrastructure execution only. Results retain executed steps, physical time,
terminal reason, validity, and protocol revision; historical longer-budget
successes are never relabeled as successes within a later budget.
A repaired environment requires fresh target-task trials, and missing tasks
and model/task coverage remain explicit in aggregates.

\subsection{Diagnostics and Statistical Reporting}
\label{sec:diagnostics}
Success and failure alone do not say \emph{where} a policy stops working.
We record native-stage diagnostics where supported by the task checker and
report their coverage explicitly.

\paragraph{Task success and ordered progress.}
Primary scores are per-task success and macro-averages over declared task
sets. Native executable success predicates and physical-duration holds
determine completion. For a task with $K_i$ stages, ordered progress is
$P_e=L_e/K_i$, where $L_e$ is the longest valid ordered prefix completed in
episode $e$. Stages are defined by semantic operations; merely approaching or
touching an object does not complete a functional stage. Progress never
overrides the final success predicate. Ordered progress is reported for the
trained multi-task methods; GPT-6 has no graded-progress measurement.\looseness=-1

\paragraph{Capability-specific diagnostics.}
Functional completion separates acquiring a tool from operating it.
Stability measures include drops, slip, and displacement of an object
supported by the other hand when these quantities are observable. Deformable
tasks additionally report task-specific geometric progress. Contact-force
measures require validated sensors or simulator measurements; command
magnitude is never treated as measured force. We distinguish tool
acquisition from functional activation, requested numeric setting from any
actuator movement, and articulated travel from mere object contact.
Capability labels overlap and use disclosed membership; their averages are
not added into a new overall score.

\paragraph{Trials and uncertainty.}
Trials preserve task-local resets, assets, source/goal weights, state/action
semantics, and checkpoint-owned preprocessing and normalization. Independent
evaluation seeds are shared across compatible model comparisons. Binary
success receives Wilson intervals; aggregate uncertainty is reported with
the resampling unit specified. Missing or environment-invalid trials are not
numerical policy failures. Every aggregate identifies its task and trial
coverage, and checkpoint selection, benchmark-development exposure, and
historical data exposure are disclosed in the run record.\looseness=-1

\paragraph{Reading the diagnostics.}
Native event-arrival rates and ordered completion scores are different
quantities. A conditional transition rate requires joint, order-consistent
events, not division of unrelated stage marginals. Retention diagnostics
identify both the achieved condition and the subsequent observation interval;
success-triggered termination can give episodes unequal exposure. Recorded
action commands are not measured wrist motion or contact force.

\section{The HandsOn Data Platform}
\label{sec:data-platform}

HandsOn pairs a teleoperated demonstration corpus with a re-rendering
pipeline. Collection produces exact state and action supervision at the cost
of a fixed set of recorded appearances; re-rendering then varies the
observations \emph{around} that recorded behaviour while the interaction and
its labels stay fixed. The two halves are described in turn.

\begin{figure}[t]
  \centering
  \includegraphics[height=1.55in]{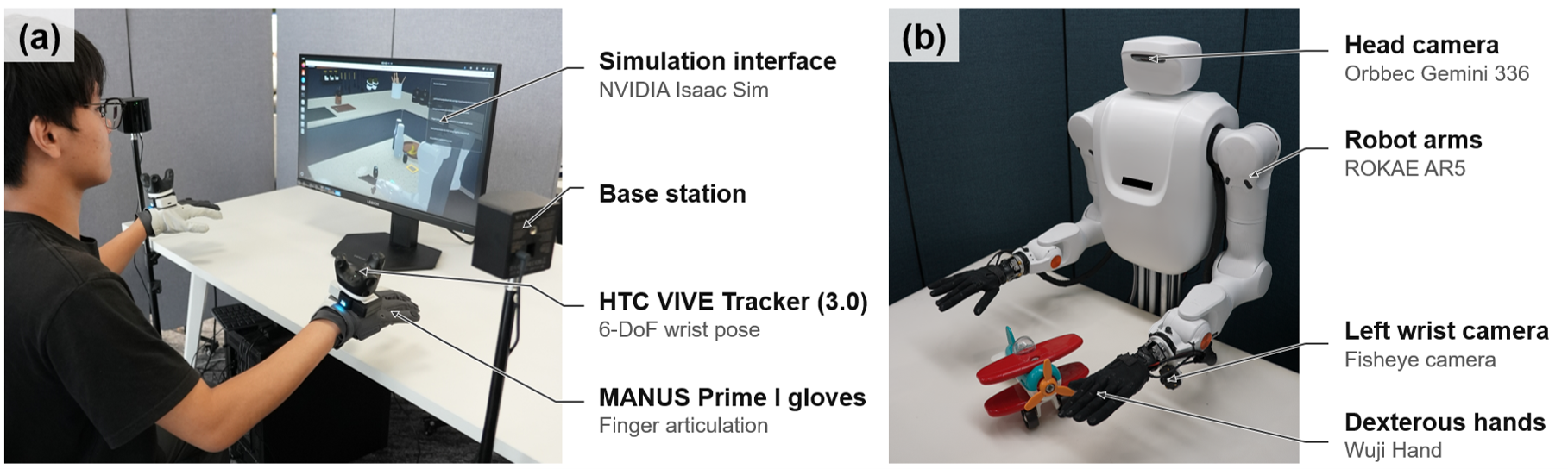}
  \caption{Human demonstration collection. Gloves record finger articulation
  and wrist trackers record hand motion. Retargeted commands drive the
  simulated arm--hand system while the \mbox{operator observes the scene.}}
  \label{teleop-setup}
\end{figure}

\subsection{Collection System and Operators}
\label{sec:collection}
Demonstrations are collected in NVIDIA Isaac Sim 5.1~\cite{isaacsim51}.
MANUS Prime I gloves capture finger articulation, and wrist-mounted HTC VIVE
Trackers provide global wrist poses
\cite{manus_prime_series,vive_tracker3}. After frame registration and temporal
alignment, wrist motion and a palm-centred 21-keypoint hand representation
are converted to commands. The arm branch produces feasible end-effector
targets through robot kinematics; Wuji retargeting maps hand keypoints to
joint targets, with 20 outputs for the first-generation Wuji
Hand~\cite{wujiretargeting}. Simulation records synchronized observations,
measured state, commanded actions, language, and source metadata
(Figure~\ref{teleop-setup}).

At least ten operators contributed over 1,050 collection sessions, which is
why the feasibility procedure of Section~\ref{sec:task-design} separates
recording from review. All demonstrations are simulated: no physical robot
recordings enter the training set, and the physical platform of
Section~\ref{sec:real-platform} is used only for the qualitative rollouts of
Appendix~\ref{sec:sim_real_complementarity}.

\subsection{Dataset Composition and Identity}
\label{sec:dataset-identity}
The released dataset contains 18,000 curated trajectories: 15,178 training
and 2,822 validation. Selection uses task-group quotas of 500, 300, or 100
trajectories, favouring shorter eligible demonstrations within each quota.
This is a curated distribution, not an unbiased sample of all collected
behaviour. Three stages of one multi-stage source task share a quota, giving
52 quota groups. Three simultaneous camera views do not multiply the
interaction duration. Across the package the median trajectory has 262 steps
(10th--90th percentile 96--705); per Core-50 task, training trajectories
range from 46 to 463 (median 268) and expert P75 durations from 2.5 to
49.2~s (median 12.7~s). Split membership is defined at trajectory or session
level and re-rendered derivatives inherit it; normalization fitted for this
package uses training data only, and a checkpoint that inherits another
normalization or has historical exposure retains that provenance.

Table~\ref{tab:identity-units} lists the counting units that recur through
the paper. Keeping them separate matters: the 54,000 video streams are three
views of the same 18,000 interactions, and the raw collection before
deduplication and curation is not the training size of any checkpoint.

\begin{table}[!hbp]
  \centering\small
  \caption{Distinct counting units in the dataset and catalog. The video
  count is three views of the same interactions, not three times the data.}
  \label{tab:identity-units}
  \begin{tabularx}{\linewidth}{>{\raggedright\arraybackslash}Xr}
    \toprule
    \TabHead\TabStrut Unit & Count\\\TabHeadEnd
    \midrule
    Curated demonstration trajectories & 18,000\\
    Training / validation trajectories & 15,178 / 2,822\\
    Recorded time steps & 6,053,555\\
    Camera videos (three per trajectory) & 54,000\\
    Native task identities / quota groups & 54 / 52\\
    Core-50 / Transfer-16 tasks & 50 / 16\\
    \bottomrule
  \end{tabularx}
\end{table}

The evaluation catalog contains the Core-50 tasks and excludes one source
task, Chinese Chess, for every model. This is a scope exclusion: it does not
remove demonstrations from the 18,000-trajectory package and does not rewrite
any checkpoint's training provenance. A mapping record links each paper task
to its source task or cohort, goal, asset identities, episodes, and executable
evaluation task.

\subsection{High-Fidelity Re-rendering}
\label{sec:randomization}
Re-rendering varies appearance while replaying the recorded interaction.
Final-package capture records use 32 samples per pixel; higher-sampling
presentation renders are identified separately. The implemented bank includes
four table materials, three background configurations, and three ceiling
styles, with texture scale and orientation, surface roughness and tint, and
illumination direction, intensity, and colour-temperature variations. One
verified source camera configuration samples each camera once per trajectory
within $\pm2$~mm translation and $\pm0.3^\circ$ rotation per axis, retaining
intrinsics. Episode records identify source trajectory and runtime, render
revision, appearance seed and configuration, and camera configuration.
Appearance is sampled per trajectory and remains fixed during replay;
independently resampling appearance every frame is outside this
static-domain protocol, and any temporal sensor noise would require a
separate model.

Not every factor can be varied by replay alone. Table~\ref{tab:randomization}
states what each factor must preserve. Background and appearance changes can
be applied during replay because the recorded actions remain valid; geometry
and dynamics changes alter the reachable interaction and require valid
re-execution rather than frozen-state replay. We use the taxonomy of
Table~\ref{tab:randomization} as the rule for adding variation: a factor is
admitted only with the evidence its row requires.

\begin{table}[!hbp]
  \centering\small
  \caption{Visual, language, and physical variation and the labels each must
  preserve. Geometry and dynamics changes require evidence beyond visual
  state replay.}
  \label{tab:randomization}
  \begin{tabularx}{\linewidth}{@{}p{0.17\linewidth}XX@{}}
    \toprule
    \TabHead\TabStrut\TabL{tabHead}Factor & Examples & \TabR{tabHead}Must preserve\\\TabHeadEnd
    \midrule
    Appearance & Material, lighting, background, ceiling & Preserve recorded
      interaction and task-relevant visual semantics.\\
    Camera & Extrinsic perturbation about calibration & Preserve timing;
      transform camera-frame labels if affected.\\
    Language & Canonical instruction and paraphrases & Preserve object,
      numerical goal, direction, order, and constraints.\\
    Task geometry & Object pose, goal relation, native reset & Retain each
      source task's joint sampling distribution and compatible actions.\\
    Dynamics & Mass, friction, compliance, actuation & Requires valid
      re-execution; not established by frozen-state re-rendering.\\
    \bottomrule
  \end{tabularx}
\end{table}

Camera perturbations are composed with task-specific nominal calibration.
Within the trained-checkpoint Core-50 evaluation, which of the four table materials, three colour
palettes, or two roughness levels is drawn does not measurably change success
(Section~\ref{sec:reliability}). Those factors lie inside the training
distribution, so this is an invariance check rather than a
visual-generalization result.

\subsection{Language Bank and Prompt Sampling}
\label{sec:language-bank}
The frozen language bank contains 108 task--instruction entries, of which 91
occur in the final 18,000-trajectory manifest. For these 91 active entries,
there are 32 training and eight held-out paraphrases per entry, yielding
2,912 training and 728 held-out paraphrases. For example, ``Pick the apple and
place it into the bucket'' and ``Grab the apple and put it in the bucket''
preserve the same goal. Replacing apple with banana, changing a knob angle,
or changing operation order instead requires an episode with the
corresponding executable goal, which is why those variations are task
variations rather than paraphrases. Held-out strings have no exact overlap
with any active canonical or training string; this is a lexical disjointness
check, and semantic equivalence additionally requires preserving the
episode's target and constraints. Inactive bank entries are not counted as
demonstrated goals.\looseness=-1

The packaged sampler hashes the language seed, epoch, episode identity, frame
index, draw index, and split to choose an instruction, and offers training,
canonical, and unseen modes. The reference training mixture is 20\% canonical
and 80\% training paraphrases; fixed validation uses the run's declared
language mode. Language sampling does not resample the scene, cameras, or
physics. A model record must identify the language bank revision, the actual
sampling policy, prompt preprocessing, and whether conditioning is text, a
structured goal, or a fixed task identifier. Section~\ref{sec:language-probe}
examines whether policies condition their actions on the object specified in
the instruction.

\subsection{Physical Platform and Simulated Correspondence}
\label{sec:real-platform}
The platform uses ROKAE AR5-08 arms (seven degrees of freedom, 5~kg rated
payload, 683~mm reach) and first-generation Wuji Hands with 20 actively
controlled joints per hand~\cite{rokae_ar5,wuji_hand_v1}. The simulated
kinematics, asset scale, cameras, state and action frames, and timing follow
this physical system, and the correspondence record is given in
Appendix~\ref{app:physical-system}.

\subsection{Portability Across Arms, Hands, and Physics Backends}
\label{sec:portability}
HandsOn is organized around a task and asset layer that is independent of the
particular hardware that executes it, so that a task definition, its assets,
its success predicate, and its demonstration set survive a change of arm,
hand, or physics back-end. The released code supports the substitution of
several commercial arms, including the ROKAE arms used in the reported
experiments and additional supported arms from KUKA, UR, ABB, and
FANUC; several dexterous hands, including the first- and second-generation
Wuji Hands and the Sharpa Hand; and two physics back-ends, NVIDIA PhysX and
Newton. The benchmark therefore measures a manipulation capability rather
than a specific piece of hardware: the same Core-50 and Transfer-16
definitions can be instantiated against any supported arm--hand pair, and
the demonstrations can be re-collected on that pair without editing the task
itself.

All experiments reported in this paper use one configuration: the ROKAE
AR5-08 arms and first-generation Wuji Hands of
Section~\ref{sec:real-platform}, with NVIDIA PhysX as the physics back-end.
This is the configuration of the 18,000-trajectory corpus, of every
collection session, and of every Core-50 and Transfer-16 result in
Section~\ref{sec:results}. The other arms, hands, and back-ends are
supported by the platform and are the natural basis for studying transfer
across embodiments, but they do not contribute episodes to the results
reported here, and the numbers in this paper should not be read as
\mbox{performance on those configurations.}

\section{Experiments and Analysis}
\label{sec:results}
We evaluate five VLA models, three WAM models, single-task DP, and GPT-6
with \texttt{move\_eef} on Core-50. Section~\ref{sec:experimental-setup} describes the
experimental setup. Section~\ref{sec:benchmark-results} presents the main
results on Core-50 and Transfer-16, and Section~\ref{sec:detailed-analysis}
examines demonstration scale and task duration, instruction conditioning,
failure modes, task grouping, and evaluation reliability.
GPT-6 has ten trials per task on both suites.
Tables~\ref{tab:seen-results} and~\ref{tab:unseen-results}
include this prompted reference, with its full results in
Section~\ref{sec:gpt6-results}. Skill-group and per-task comparisons and
outcome summaries also include GPT-6 where measured results are available.
Trained-policy averages and analyses of learning, stage transitions, and
checkpoint reliability retain their original populations.

\subsection{Models and Experimental Setup}
\label{sec:experimental-setup}
We evaluate ten methods: single-task DP, five VLA models, three WAM models,
and GPT-6 as a prompted VLM controller. The eight VLA/WAM checkpoints are
trained on the Core-50 demonstrations. The
five vision-language-action (VLA) models are $\pi_{0.5}$~\cite{pi05},
DM0.5 of the DM0 family~\cite{dm0}, LingBot-VLA~\cite{lingbot},
HY-Embodied-0.5-VLA~\cite{hyembodied}, and Spatial
Forcing~\cite{spatialforcing}, which aligns the visual features of
$\pi_{0.5}$ with a pretrained 3D foundation model during training. The three
world-action models (WAMs), OpenWAM~\cite{openwam}, FastWAM~\cite{fastwam},
and Motus~\cite{motus}, combine action prediction with modeling of future
observations. Each checkpoint executes its integration's default action
chunk per query (Table~\ref{tab:seen-results}, Steps/plan). The selected
checkpoints are $\pi_{0.5}$ at 120,000 updates, DM0.5 at 86,000, OpenWAM at
70,000, Motus and FastWAM at 100,000, Spatial Forcing at 90,000,
HY-Embodied at 125,000, and LingBot-VLA at 151,000; they are fixed, not
chosen by benchmark success. The single-task reference, Diffusion Policy
(DP)~\cite{chi2023diffusion}, trains one 30,000-update model per Core-50
task. Without a predeclared routing protocol it is not run on
Transfer-16; this is not a measured zero-shot failure.
Pretraining data and compute differ across models, so the comparison does
not isolate architecture.

\paragraph{GPT-6 with \texttt{move\_eef}.}
We additionally evaluate GPT-6 Astra (API identifier \texttt{gpt-6-astra},
reasoning effort \texttt{medium}) without benchmark finetuning or
demonstrations in its prompt. At each turn it receives three RGB views,
measured palm poses and finger joints, the task instruction, and execution
feedback. Its only callable function, \texttt{move\_eef}, specifies absolute palm poses
and 20 independent finger-joint targets per hand. A kinematic adapter
converts these targets to coordinated 30~Hz trajectories for the shared
native executor, with no learned grasp primitive; simulation pauses during
API inference. The batch assigns ten fixed trials to each of the 66 tasks,
500 on Core-50 and 160 on Transfer-16. We use ten trials per task because
of API-cost and wall-clock evaluation-time constraints, allocating the
available budget across the full task catalog. These trial counts, the prompted
interface, and the additional call budget differ from the trained-policy
experiments. Appendix~\ref{app:gpt6-protocol} specifies the prompts,
controller, recorded budgets, and interruption handling;
Section~\ref{sec:gpt6-results} reports the completed 660-trial experiment.

\Needspace{6\baselineskip}
\subsection{Benchmark Results}
\label{sec:benchmark-results}

\subsubsection{Performance on Core-50}
\label{sec:seen-results}
\begin{figure}[t]
  \centering
  \includegraphics[width=.97\linewidth]{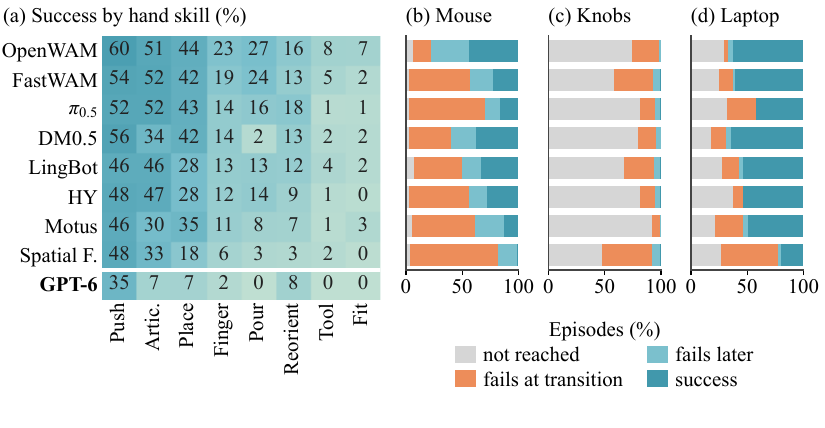}
  \caption{Core-50 diagnostics. (a)~Task-macro success by hand-skill group
  for eight multi-task checkpoints and GPT-6 (\texttt{move\_eef}); ten
  trials/task for GPT-6 versus 100 for trained checkpoints (one invalid
  HY-Embodied episode). Task groups are listed in
  Table~\ref{tab:skill-groups}. (b)--(d)~Episodes at the three transitions of
  Section~\ref{sec:main-failures}, measured for the eight trained checkpoints;
  grey: transition never reached, orange:
  failed at it, light blue: failed after it. (b)~Mouse moved 8~cm onto its
  pad before any click. (c)~Both knobs within $5^\circ$ of target at once
  (grey: at most one knob). (d)~Laptop display ever at the requested angle,
  and still there at the end.}
  \label{fig:main-analysis}
\end{figure}

Table~\ref{tab:seen-results} reports all 50 Core-50 tasks for the eight
multi-task checkpoints and the DP and GPT-6 references (per task in
Appendix~\ref{app:seen-results}). For
analysis we group the tasks by the hand skill they demand
(Section~\ref{sec:reliability}; Figure~\ref{fig:main-analysis}a).
\begin{table}[!hbp]
  \centering\small\setlength{\tabcolsep}{6.6pt}
  \caption{Core-50 task-macro success (\%), 100 episodes per task and learned checkpoint (GPT-6: ten); 95\% intervals in Table~\ref{tab:seen-intervals}. Steps/plan: control steps (30~Hz) per policy query. Tasks $\geq$50\%: number of the 50 tasks with at least 50\% success. P/F/CR: primitive (17), functional (18), contact-rich (15) tasks. @1$\times$: success within the expert p75 duration. Early: episodes ended by an irrecoverable failure. Stall: the percentage of instrumented time-outs with no new checker state after the expert p75 duration (Section~\ref{sec:main-failures}). Shaded: best learned checkpoint (lowest for Early, Stall).}
  \label{tab:seen-results}
  \begin{tabularx}{\linewidth}{>{\raggedright\arraybackslash}Xccccccccc}
    \toprule
    \TabHead\TabStrut Model & Steps/plan & SR & Tasks $\geq$50\% & P & F & CR & @1$\times$ & Early & Stall\\\TabHeadEnd
    \midrule\TabGapTop
    \TabStrutTop OpenWAM & 16 & \TabBest{29.1} & \TabBest{11} & 40.8 & \TabBest{23.8} & \TabBest{22.1} & 8.3 & \TabBest{10.9} & \TabBest{65}\\
    FastWAM & 16 & 26.4 & 9 & \TabBest{42.7} & 20.2 & 15.3 & 10.2 & 13.0 & 72\\
    $\pi_{0.5}$ & 30 & 24.9 & 10 & 40.4 & 14.6 & 19.7 & \TabBest{14.6} & 14.1 & 78\\
    DM0.5 & 30 & 21.4 & 9 & 37.7 & 10.0 & 16.7 & 10.9 & 13.9 & 80\\
    LingBot-VLA & 30 & 19.7 & 8 & 26.9 & 14.3 & 17.9 & 12.0 & 12.7 & 83\\
    HY-Embodied & 30 & 18.9 & 8 & 27.1 & 13.3 & 16.4 & 11.0 & 14.4 & 81\\
    Motus & 16 & 18.1 & 5 & 30.1 & 11.5 & 12.3 & 5.4 & 15.6 & 70\\
    Spatial Forcing & 16 & 12.4 & 3 & 17.9 & 8.5 & 10.9 & 5.6 & 16.5 & 75\\
    \midrule
    DP (single-task)$^\dagger$ & --- & 9.7 & 3 & \multicolumn{6}{l}{reference; 9.9 on 11 matched tasks}\\
    \midrule
    GPT-6 (\texttt{move\_eef})$^\ddagger$ & 1--300 & 5.2 & 1 & 7.6 & 1.1 & 7.3 & --- & --- & ---\\
    \bottomrule
  \end{tabularx}
  \par\smallskip\footnotesize
  $^\dagger$One 30k-step model/task. Earlier goal sampler: 38 tasks; matched: ten retrained single-arm tasks and Open Latched Door ($\pi_{0.5}$: 29.1); separate: Lift Carton. All 50 contribute to SR. HY-Embodied: 99 valid episodes on Place Wine Glass.
  $^\ddagger$Prompted reference, ten trials/task due to API-cost and wall-clock evaluation-time budgets; variable \texttt{move\_eef} trajectory length, differing state inputs and recorded call budgets (Appendix~\ref{app:gpt6-protocol}). GPT-6 intervals are in Tables~\ref{tab:seen-intervals} and~\ref{tab:gpt6-results}; unreported diagnostics are dashes, not zeros.
\end{table}

\paragraph{Policies move objects and mechanisms; tools and precision fits remain unsolved.}
Among the eight trained multi-task checkpoints, OpenWAM has the highest
success, 29.1\%, and picking the best checkpoint
per task would give 35.4\%. Success is ordered by what the task asks of
the fingers beyond closing them. Pushing or flicking without a grasp
(two tasks, 51\% pooled over checkpoints), moving a hinge, drawer, lever, or lid
(43\%), and grasp--transport--release including handovers and object
selection (35\%) are the relatively better-performing groups in the suite; all nine
tasks with mean success above 40\% are of these kinds. Operating a trigger, button, knob,
slider, or shutter with the fingers drops to 14\%, pouring to 13\%, and
regrasping or reorienting an object in or between the hands to 11\%. Two
groups are essentially unsolved: acting on a third object through a held
tool (137 of 4,800 episodes, 2.9\%; hammer, spoon, set square, and spatula
together 9 of 3,200) and precision fits such as peg insertion, stacking, or
face alignment (107 of 4,800, 2.2\%). The best checkpoint on any of these
twelve tasks reaches 21\% (OpenWAM on Stir Pot); on ten of them none
exceeds 14\%. Skill-group gaps persist after adjusting for demonstration
count and expert duration:
precision fits have more demonstrations per task (median 413) than
grasp--transport--release (392), and with log demonstration count and log
expert duration as covariates the tool, precision, in-hand, and
finger-actuation groups stay 20--25 points below grasp--transport--release,
with task-bootstrap intervals excluding zero (Section~\ref{sec:ablations}).

\paragraph{Checkpoints differ in using a grasped object, not in grasping it.}
On grasp--transport--release tasks four checkpoints are within two
points (OpenWAM 44, $\pi_{0.5}$ 43, FastWAM 42, DM0.5 42), and the
OpenWAM--$\pi_{0.5}$ gap on primitive tasks is $+0.5$ points
(paired task bootstrap 95\% interval $-6.9$ to $+7.6$). The same gap on
functional tasks is $+9.2$ ($+4.7$ to $+13.9$): Pour Seasonings 60 versus
34\%, Move and Click Mouse 44 versus 16, Press Soap Pump 33 versus 9, Stir
Pot 21 versus 4. Section~\ref{sec:main-failures} locates this gap at
specific transitions: OpenWAM clicks before the mouse is ready in 16 of
100 episodes and $\pi_{0.5}$ in 67, OpenWAM keeps the laptop
display at the requested angle in 67 and $\pi_{0.5}$ in 42, and OpenWAM
completes the handover to the receiving hand in 41 and $\pi_{0.5}$ in 20.
Two VLA checkpoints place fixed objects as well as the leaders but
collapse when the instruction selects the object: HY-Embodied and LingBot-VLA
score 55 and 61\% on Place Cup on Plate ($\pi_{0.5}$ 62) but at most 22\%
on Place Ball in Bowl, Place Fruit in Bucket, and Place Largest Fitting
Ball, where $\pi_{0.5}$ scores 35--75\%. They barely read the object name:
in an offline probe (Section~\ref{sec:language-probe}), naming the other
ball moves the 1-s palm endpoint of $\pi_{0.5}$ and DM0.5 by twice the
resampling change (15 and 17 versus 7 and 8~cm) but that of HY-Embodied
and LingBot-VLA by no more (7--9~cm), and with the swapped name they still
head for the original ball in 57 and 71\% of first frames (DM0.5 10\%). Motus is a
strong picker (76\% on Place Ball in Bowl) but near zero wherever a
mechanism must be held open or a load carried (Open Micropipette Box 8\%,
Open Drawers 4\%, Lift Carton 8\%). Spatial Forcing is weakest in every
group and, unlike all other checkpoints, worse on single-arm (3.8\%) than
on bimanual tasks (14.5\%).

\begin{figure}[!htbp]
  \centering
  \includegraphics[width=.6\linewidth]{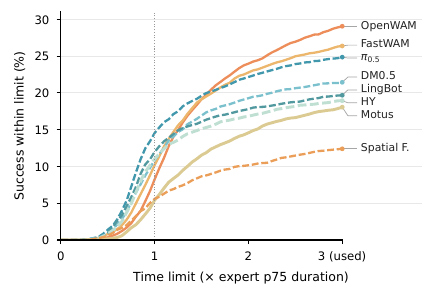}
  \caption{Task-macro success counting only episodes completed within a given
  multiple of the expert 75th-percentile duration. Shorter limits are prefixes
  of the evaluated $3\times$ rollouts, not reruns, so the curves do not
  estimate what a longer limit would add. Solid: world-action models; dashed:
  VLA checkpoints. Success at $1\times$/$1.5\times$/$2\times$/$3\times$ is
  OpenWAM 8.3/19.1/24.0/29.1; FastWAM 10.2/19.1/22.8/26.4; $\pi_{0.5}$
  14.6/19.6/22.2/24.9; DM0.5 10.9/16.9/19.3/21.4; LingBot-VLA
  12.0/16.1/17.8/19.7; HY-Embodied 11.0/15.1/16.8/18.9; Motus
  5.4/11.5/14.6/18.1; Spatial Forcing 5.6/8.7/10.2/12.4. Twelve of the 28
  checkpoint pairs change order between $1\times$ and $3\times$.}
  \label{fig:time-limit}
\end{figure}

\paragraph{World-action models finish later but finish more.}
The three WAM
checkpoints complete their successes at a median of 1.13--1.27 p75
durations versus 0.88--0.99 for the four 30-step VLA checkpoints, and more
of their time-outs are still making progress after $1\times$ (14.5--18.7\%
of episodes versus 10.1--11.8\%). Counting only completions within $1\times$
(Table~\ref{tab:seen-results}, @1$\times$), $\pi_{0.5}$ leads with 14.6\%
and OpenWAM ranks sixth with 8.3\%; 12 of the 28 checkpoint pairs change
order between the two limits (Figure~\ref{fig:time-limit}). The WAMs
execute 16 control steps per query, but Spatial Forcing, a VLA that also
executes 16, is not slow (median 1.09), so the query cadence alone does not
explain the difference; @1$\times$ is therefore the secondary score
(Section~\ref{sec:evaluation}).

\paragraph{Reliability and feasibility.}
The 95\% interval over tasks and episodes (22.8--36.0\% for OpenWAM,
Table~\ref{tab:seen-intervals}) mainly reflects the task mix (episodes
alone: 28.0--30.1\%). Paired task resampling orders
$\pi_{0.5}$ above DM0.5 in 99\% of draws but HY-Embodied above Motus in
only 61\% (Section~\ref{sec:reliability}). Leaving out any one task
swaps at most HY-Embodied and Motus and moves no score by more than 1.4
points. Every task is completed by verified human demonstrations
(Section~\ref{sec:task-design}). For 47 of the 50 Core-50 tasks, at least one
evaluated multi-task checkpoint succeeds in at least one episode.

\paragraph{Single-task reference.}
DP averages 9.7\% over 50 tasks: 38 use an earlier goal sampler,
11 use the current protocol, and Lift Carton is evaluated separately
(Appendix~\ref{app:seen-results}). On the 11 matched tasks it reaches
9.9\%, against 22.7--34.5\% for seven multi-task
checkpoints and 9.3\% for Spatial Forcing. This does not separate
multi-task data sharing from pretraining or model scale.

\FloatBarrier
\subsubsection{Generalization to Transfer-16}
\label{sec:transfer}
Table~\ref{tab:unseen-results} groups Transfer-16 results by asset/relation
transfer, skill composition, and deformable manipulation, with ordered
progress crediting the longest completed prefix of stages. Transfer-16 is
much harder than Core-50, and we rank trained checkpoints by ordered progress, which
also credits partial completion. OpenWAM's 12.9\% exceeds HY-Embodied,
LingBot-VLA, and Spatial Forcing by 6.0, 3.7, and 2.6 points (paired
intervals exclude zero); the other four are within 1.7 points.
\begin{table}[!htbp]
  \centering\small
  \caption{Transfer-16 results (\%): task-macro success (SR) and ordered
  progress (Prog.) within the declared time limit over the 7 Gen, 6 Comb, and
  3 Flex tasks; 50 episodes per task and checkpoint (Stack Boxes 37 and Spray
  Shoe 43, whose missing resets fail for every checkpoint; at most two invalid
  episodes elsewhere). Gen02 logs no graded hinge progress and is excluded
  from Prog. DP is not run (Section~\ref{sec:experimental-setup}). Shaded:
  best trained checkpoint per column. GPT-6 is a prompted reference with
  ten trials/task and different inputs/call budgets; graded progress was
  not evaluated for this condition.}
  \label{tab:unseen-results}
  \begin{tabularx}{\linewidth}{l*{8}{>{\centering\arraybackslash}X}}
    \toprule
    \TabHead\TabStrutTop Model & \multicolumn{2}{c}{Asset/relation} & \multicolumn{2}{c}{Skill composition} & \multicolumn{2}{c}{Deformable} & \multicolumn{2}{c}{All 16}\\
    \rowcolor{tabHead}\TabStrutBot & SR & Prog. & SR & Prog. & SR & Prog. & SR & Prog.\\\TabHeadEnd
    \midrule
    $\pi_{0.5}$ & 0.0 & 14.0 & 0.3 & 11.7 & 0.0 & 8.3 & 0.1 & 11.9\\
    DM0.5 & \TabBest{1.4} & 9.2 & \TabBest{0.7} & \TabBest{14.4} & 0.0 & 8.8 & \TabBest{0.9} & 11.2\\
    LingBot-VLA & 0.0 & 13.3 & 0.0 & 4.2 & 0.0 & 11.3 & 0.0 & 9.3\\
    HY-Embodied & 0.0 & 9.6 & 0.0 & 4.6 & 0.0 & 6.2 & 0.0 & 6.9\\
    OpenWAM & 0.9 & 14.8 & 0.0 & 12.5 & 0.0 & 10.2 & 0.4 & \TabBest{12.9}\\
    Motus & 0.3 & \TabBest{15.8} & 0.0 & 9.5 & 0.0 & \TabBest{11.7} & 0.1 & 12.4\\
    FastWAM & 0.0 & 12.5 & 0.0 & 11.3 & 0.0 & 11.3 & 0.0 & 11.8\\
    Spatial Forcing & 0.3 & 12.7 & 0.0 & 7.9 & 0.0 & 10.5 & 0.1 & 10.3\\
    \midrule
    GPT-6 (\texttt{move\_eef}) & 4.3 & --- & 0.0 & --- & 0.0 & --- & 1.9 & ---\\
    \bottomrule
  \end{tabularx}
\end{table}

Human verification establishes the feasibility of Transfer-16 tasks under
the verified configurations (Section~\ref{sec:task-design}). The low success
rates indicate substantial policy performance gaps under the evaluated
protocol. For the trained checkpoints, success is close to zero on all three families, which makes
aggregate success a weak discriminator: ordered progress separates the
checkpoints where success cannot. The three deformable tasks are unsolved by
every checkpoint, and the compositions, which require completed skills to be
chained, are the hardest family after them (Section~\ref{sec:main-failures}).

\FloatBarrier
\Needspace{6\baselineskip}
\subsubsection{GPT-6 with \texorpdfstring{\texttt{move\_eef}}{move\_eef}}
\label{sec:gpt6-results}
GPT-6 achieves 26 successes in 500 Core-50 trials (5.2\%) and three in
160 Transfer-16 trials (1.9\%), for 29/660 overall (4.4\%;
Table~\ref{tab:gpt6-results}). All 66 tasks have ten scored outcomes.
No demonstrations or HandsOn policy training are provided. The model
chooses palm poses and individual finger targets from RGB observations,
measured robot state, the task instruction, and execution feedback.

\begin{table}[!htbp]
\centering\small
\caption{GPT-6 with \texttt{move\_eef}: ten scored trials per task. SR is the unweighted task mean; equal trial counts make it identical to pooled success. Intervals are conditional 95\% within-task episode bootstrap intervals (20,000 resamples), with the task catalog fixed. Cap: unsuccessful outcomes ended by the applicable model-call budget. Recorded budget settings are described in Appendix~\ref{app:gpt6-budget}.}
\label{tab:gpt6-results}
\begin{tabularx}{\linewidth}{@{}Xrrrrr@{}}
\toprule
Suite & Tasks & Successes / trials & SR (\%) & 95\% CI & Cap (\%)\\
\midrule
Core-50 & 50 & 26 / 500 & 5.2 & [3.6, 6.8] & 39.8\\
Transfer-16 & 16 & 3 / 160 & 1.9 & [0.0, 3.8] & 38.1\\
\midrule
All & 66 & 29 / 660 & 4.4 & [3.2, 5.6] & 39.4\\
\bottomrule
\end{tabularx}
\end{table}

\begin{figure}[!htbp]
  \centering
  \includegraphics[width=\linewidth]{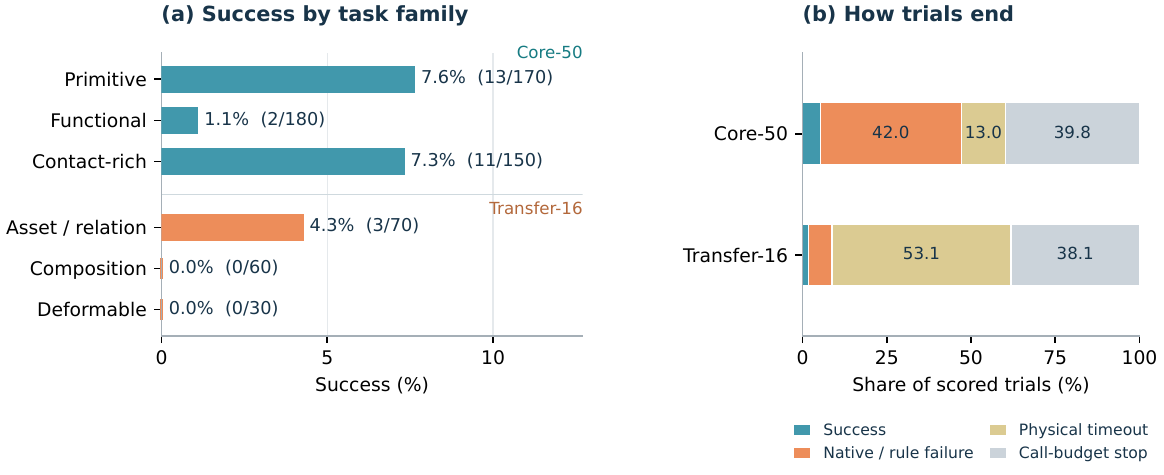}
  \caption{GPT-6 with \texttt{move\_eef} on all 660 scored trials. (a) Success by the declared task
  families; labels give successes/trials. Each task contributes ten trials.
  (b) Trial outcomes within each suite. A call-budget stop is an unsuccessful
  outcome under the recorded budget, not a native physical failure.
  Native/rule failures include the ten forbidden direct-hand-contact
  endings on Line Up Blocks. Physical time does not advance while the
  model is queried.}
  \label{fig:gpt6-results}
\end{figure}

\paragraph{Success is concentrated in a small part of the catalog.}
Only ten Core-50 tasks and two Transfer-16 tasks have any successful trial
(Appendix~\ref{app:gpt6-tasks}). Rotate Propeller has the highest observed
success, 7/10; Orient Cup Upright and Lift Carton each have 4/10, followed
by Open Micropipette Box and Place Cup on Plate at 3/10. The remaining
five solved Core-50 tasks each have 1/10. Across the primitive, functional,
and contact-rich categories, success is 7.6\%, 1.1\%, and 7.3\%,
respectively (Figure~\ref{fig:gpt6-results}a). The contact-rich score comes
entirely from Rotate Propeller and Lift Carton; it does not indicate broad
competence in precision fitting. Transfer-16 successes occur only on
Open Trash Bin (2/10) and Orient Bottle Upright (1/10), giving 4.3\% on
asset/relation transfer and no observed success on skill composition or
deformable manipulation.

\paragraph{Call budgets and physical failures both limit completion.}
Of the 660 trials, 260 (39.4\%) end at their applicable model-call budget,
150 (22.7\%) reach the physical-time limit, and 221 (33.5\%) end through a
native termination or task-rule violation without success
(Figure~\ref{fig:gpt6-results}b). These are distinct from the 29 successes.
Budget stops are particularly common in the composition and deformable
families (31/60 and 24/30). They show that the prompted controller often
cannot complete within its available interaction budget; they do not
establish what an unlimited-call controller would achieve.

\paragraph{Zero-shot task-level planning signals.}
Saved action notes suggest task-aware strategies despite low completion.
For example, in Open Laptop and Click Mouse, GPT-6 cautions that
``neither object has yet visibly moved, so do not lift or click yet.''
In Fold Towel Twice, it proposes a small test lift to ``verify that the
hem follows before carrying it forward for the first fold.'' Both trials
fail, but their notes articulate prerequisites and attempts to monitor
execution. These illustrative traces motivate studying GPT-6 as a
high-level planner. They do not measure planning accuracy across the
catalog or independently verify the physical states described by the model.
Here zero-shot refers to the absence of HandsOn demonstrations and
finetuning; the task instruction, robot contract, and static guidance
remain available.

\paragraph{High-dimensional dexterity calls for a richer agent harness.}
The current harness exposes only \texttt{move\_eef}. Alongside choosing
subgoals, GPT-6 must specify each palm pose and up to 40 finger-joint
targets, infer object geometry and contact from images and robot/execution
feedback, and recover from rejected or ineffective motions. It has no
grasp primitive, obstacle-aware planner, explicit object-pose estimate, or
tactile feedback. The inspected traces include repeated kinematic
rejections and attempts to establish or recover grasps. This suggests a
plausible gap between task-level strategy and its grounding in reliable
high-dimensional control. A promising design is to pair the multimodal
planner with estimated object/contact state and reusable, feedback-driven
manipulation skills. The present experiment motivates that design; it
does not establish how much of the failure rate is caused by the harness
or how much a richer interface would improve success.

\paragraph{Scope of the comparison.}
The Core-50 score is below every evaluated learned-policy reference, but
measures GPT-6 together with this minimal harness. The RGB-plus-state
interface, static task guidance, variable execution length, and call
budgets differ from the trained-policy conditions. Isolating planner
capability and the benefits of richer tools requires matched evaluations
that vary the interface and feedback while controlling the task population
and budgets. Appendix~\ref{app:gpt6-protocol} records the implemented
interface, settings, scoring qualifications, and uncertainty.

\FloatBarrier
\Needspace{6\baselineskip}
\subsection{Detailed Analysis}
\label{sec:detailed-analysis}

\subsubsection{Does Demonstration Scale Explain the Remaining Gap?}
\label{sec:ablations}
We finetune $\pi_{0.5}$ separately on \emph{Place Ball in Bowl} (Task~24) and
\emph{Squeeze Scissors} (Task~43) on nested subsets of 10, 50, 100, 200, and all
438 and 439 training trajectories, each run from the same pretrained
initialization for 20,000 updates at batch size 32, with one training seed
and the same 100 evaluation seeds, goals, and prompts per task
(Appendix~\ref{app:data-scaling}).

\begin{figure}[!htbp]
  \centering
  \includegraphics[width=\linewidth]{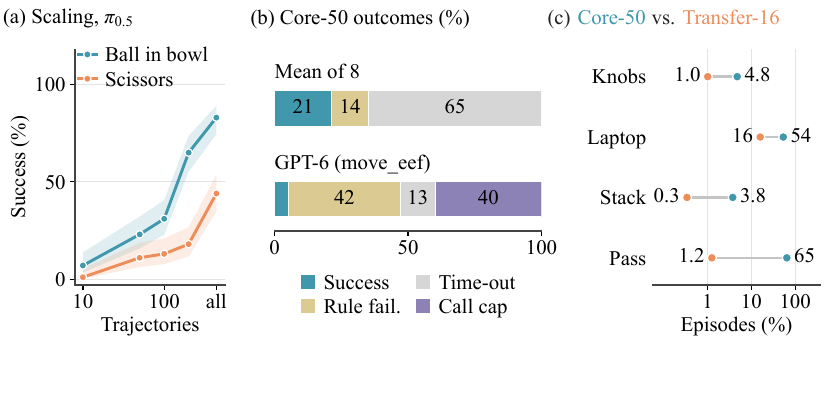}
  \caption{(a) Single-task data scaling for $\pi_{0.5}$, 20,000 updates per run
  (all: 438 and 439 trajectories); each point is one checkpoint on 100
  closed-loop trials, with 95\% Wilson bands over rollouts, not over training
  seeds or subsets. (b) How Core-50 episodes end: the task-macro mean of
  eight trained checkpoints and GPT-6 (\texttt{move\_eef}, ten trials/task).
  Rule failures and physical time-outs are distinct from call-budget stops;
  detailed measured categories are in Figure~\ref{fig:outcomes}.
  (c) Matched hand transitions in Core-50 sources and Transfer-16, pooled over
  the eight checkpoints (800 source and 296--400 transfer episodes per task).
  Panels (a) and (c) have no corresponding GPT-6 experiment.}
  \label{fig:data-scaling}\label{fig:results-row}
\end{figure}

Success rises at every tested size (Figure~\ref{fig:data-scaling}a): Task~24 from
7\% to 83\% and Task~43 from 1\% to 44\%, and going from 200 trajectories to the
full pool still adds 18 and 26 points. The functional scissors task remains
much harder. These task-local curves do not establish a cross-task scaling
law or isolate an architectural cause, and normalization uses the full
training pool at every size.

We also examine whether the Core-50 skill-group gaps persist after adjusting
for demonstration count and expert duration.
The hand-skill groups differ in how many training trajectories their tasks
have and how long the expert takes (Table~\ref{tab:skill-confound}), so group
differences could reflect data volume or episode length rather than the skill.
Across the 50 tasks, mean success over the eight checkpoints has Spearman
$\rho=0.27$ with the number of training trajectories and $\rho=-0.71$ with the
expert P75 duration. Regressing task success on the group indicators alone
explains $R^2=0.62$ of its variance; the two covariates alone explain 0.37.
With both covariates added ($R^2=0.73$), duration remains negatively associated
with success, whereas the interval for demonstration count includes zero.
The finger-actuation, in-hand, tool, and precision groups remain 20--25 points
below grasp--transport--release. The
in-hand and tool groups have few demonstrations, but precision fits have a
similar median demonstration count to the higher-performing
grasp--transport--release group. By quota, the seven 100-trajectory tasks
average 6.1\%, the 24 300-trajectory tasks 23.4\%, and the 19 500-trajectory
tasks 24.4\%; within the 500 quota, the four precision fits average 2.8\% and
the eight grasp--transport--release tasks 37.8\%.

\begin{table}[!htbp]
  \centering\footnotesize\setlength{\tabcolsep}{4pt}
  \caption{Task success (mean of the eight multi-task checkpoints, \%)
  regressed on hand-skill group over the 50 Core-50 tasks, with
  grasp--transport--release (GTR) as the reference; the GTR row gives the
  intercept of the raw model. Covariates are natural logs of the task's
  training trajectories and expert P75 duration. Brackets: 95\% intervals from
  4,000 task bootstrap draws stratified by group. Demos and P75: group medians
  over tasks.}
  \label{tab:skill-confound}
  \begin{tabularx}{\linewidth}{@{}>{\raggedright\arraybackslash}Xrr>{\raggedright\arraybackslash}X>{\raggedright\arraybackslash}X@{}}
    \toprule
    \TabHead\TabL{tabHead}\TabStrut Group & Demos & P75 (s) & Raw & \TabR{tabHead}With covariates\\\TabHeadEnd
    \midrule
    GTR (reference) & 392 & 8.2 & 34.9 [26.9, 43.1] & ---\\
    Nonprehensile & 362 & 10.5 & $+16.4$ [$-19.3$, 52.2] & $+18.7$ [$-16.1$, 52.9]\\
    Articulated part & 259 & 6.7 & $+8.0$ [$-4.6$, 21.9] & $+7.4$ [$-3.0$, 19.8]\\
    Finger actuation & 272 & 12.9 & $-20.9$ [$-31.2$, $-10.8$] & $-19.9$ [$-27.8$, $-12.2$]\\
    Pour or fill & 208 & 17.7 & $-21.5$ [$-34.6$, $-6.7$] & $-13.8$ [$-24.8$, $-3.3$]\\
    In-hand reorientation & 89 & 12.3 & $-23.5$ [$-35.2$, $-11.2$] & $-22.1$ [$-31.6$, $-11.7$]\\
    Tool on object & 159 & 23.5 & $-32.0$ [$-40.7$, $-23.5$] & $-21.3$ [$-29.9$, $-11.8$]\\
    Precision fit & 413 & 14.7 & $-32.7$ [$-41.0$, $-24.4$] & $-24.6$ [$-33.1$, $-16.6$]\\
    \midrule
    log training trajectories & & & --- & $-0.8$ [$-5.1$, 5.4]\\
    log P75 duration & & & --- & $-11.9$ [$-16.2$, $-7.6$]\\
    $R^2$ & & & 0.62 & 0.73\\
    \bottomrule
  \end{tabularx}
\end{table}

\FloatBarrier
\subsubsection{Do Policies Read the Instruction?}
\label{sec:language-probe}
To test whether a checkpoint reads the object name, we query it offline on 500
demonstration episodes of Place Ball in Bowl (training and validation
trajectories) at frames 0 and 15, 1,000 inputs in total. For each input we
sample twice with the true instruction and once with the instruction naming
the other ball, and compare the palm position commanded 30 steps (1~s) ahead.
The probe uses checkpoints of the evaluated training runs, within 6,000
updates of the evaluated ones. For $\pi_{0.5}$, DM0.5, HY-Embodied, and
LingBot-VLA, respectively, swapping the object name shifts the predicted
palm endpoint by 14.8, 17.0, 7.3, and 8.5~cm, versus 7.0, 8.2, 8.8, and
8.1~cm between two correct-instruction samples. In the same model order,
the correct-instruction endpoint is closer to the demonstration than the
swapped one in 92\%, 88\%, 61\%, and 71\% of inputs.

For a target-level test we use the $\pi_{0.5}$ true- and swapped-instruction
endpoints as references for the two balls and keep the 314 first frames where
they are at least 8~cm apart: DM0.5 heads for the named ball in 93\% of them
and, with the swapped instruction, still for the originally named ball in
10\%; HY-Embodied does so in 58 and 57\%, and LingBot-VLA in 74 and 71\% (an
earlier LingBot-VLA checkpoint: 55 and 55\%). A logistic classifier on the
frame-0 layout reaches 53\% cross-validated accuracy, near the 52\% majority
baseline. Together with the weak response to instruction swaps, these results
suggest limited instruction conditioning in HY-Embodied and LingBot-VLA;
scene-dependent memorization is one possible explanation. The serving and
offline paths produce bitwise-identical actions, ruling out discrepancies
between these two execution paths.

\FloatBarrier
\subsubsection{Where Do Episodes Fail?}
\label{sec:main-failures}
\begin{figure}[!htbp]
  \centering
  \includegraphics[width=.915\linewidth]{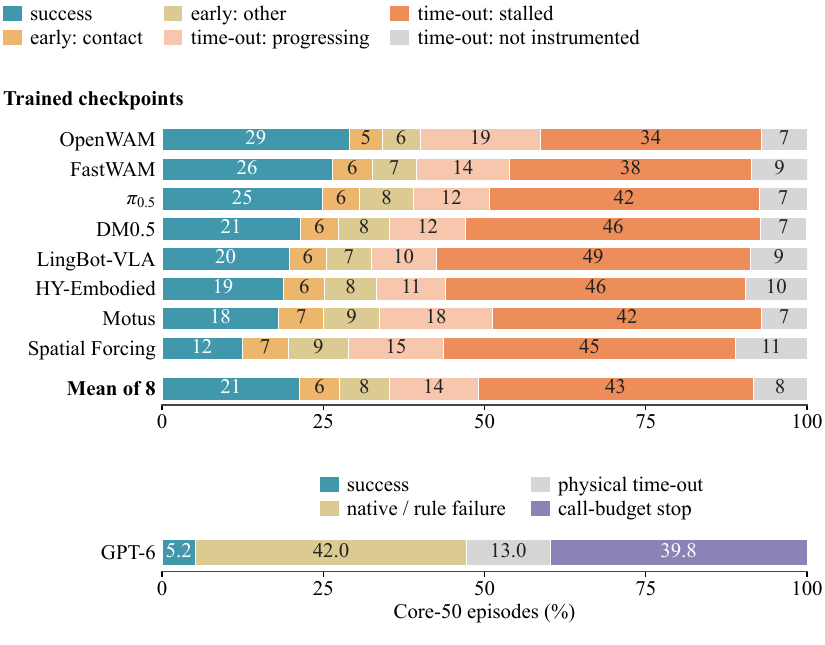}
  \caption{How Core-50 episodes end (task-macro share over 50 tasks).
  Top: the eight trained checkpoints and their mean. Early endings are native
  terminations before the time limit, including forbidden contact, premature
  events, or drops. A time-out is stalled if no new checker state follows
  the expert P75 duration, still progressing otherwise, or not instrumented
  if fewer than two progress states are logged. Bottom: GPT-6
  (\texttt{move\_eef}), 500 trials: 26 successes, 210 native/rule failures,
  65 physical time-outs, and 199 call-budget stops. Its progress-based
  time-out subtypes were not evaluated.}
  \label{fig:outcomes}
\end{figure}

Averaged over the eight trained multi-task checkpoints, 21.4\% of Core-50
episodes succeed, 13.9\% end
early, and 64.8\% time out, mostly as stalls (Figures~\ref{fig:results-row}b and~\ref{fig:outcomes}):
65--83\% of instrumented time-outs record no new checker state after
$1\times$ p75. The hands keep moving in a stall: in the last third of
time-outs the commanded mean finger angle still reverses 3--4 times per 10~s
while its range shrinks from 0.35--0.39 to 0.20--0.28~rad. Acquisition is not
the main loss on Core-50. From native checker state we locate most remaining
losses at three transitions of multi-finger bimanual control, pooling 800
episodes per task over the eight checkpoints
(Figure~\ref{fig:main-analysis}b--d).

\paragraph{Forming a grasp can fire the function.}
The fingers that hold a tool often rest on its button, trigger, or handles, so
closing the grasp can itself trigger an irreversible event. Move and Click
Mouse ends as soon as a button is pressed before the mouse has moved 8~cm onto
its pad. This ends 443 of 800 episodes, 404 of them before the mouse was ever
ready and 297 before it had moved at all (Figure~\ref{fig:main-analysis}b); of
the 362 episodes that reach the ready state without an early click, 196
succeed; the checkpoints differ mainly in the early clicks (16 for OpenWAM, 78
for Spatial Forcing). In Photograph Logo the first shutter press precedes
aiming in 774 of 778 episodes, and 163 of the 185 episodes that ever aim with
the right hand on the shutter succeed. In Water Plant the trigger is a
separate finger joint and the grasp is clean, so the loss moves to that
finger: of 514 episodes that aim the held sprayer, 238 never spray, 178 of
them after moving the trigger at least halfway. In Squeeze Scissors the same
fingers must hold and cut: failing episodes close the blades 1,368 times but
score 182 cuts, because at 57\% of the closings the scissors are not lifted
and at 26\% a handle contact has been lost. Where the functional part lies
under the holding contacts, grasping fires it early; where it is a separate
finger, that finger is not driven to threshold at the right moment; where
holding and actuating share fingers, the policy actuates only when it stops
holding.

\paragraph{Each side reaches its target, but not at the same time.}
Set Dual Knobs requires both knobs within $5^\circ$ of their targets at once,
nearly still, held, and then released. Of 800 episodes, the left and right
knob reach their target in 411 and 416, both at some point in 219, both at
once in 38, and 5 succeed (Figure~\ref{fig:main-analysis}c). When the first
knob arrives, the other is still a median $64^\circ$ away; the first knob then
leaves its $7^\circ$ exit band in 570 of 608 episodes after a median 0.33~s,
in 505 cases by overshooting and in 566 with the hand still in contact. The
hands turn each knob through its target instead of holding it there while the
other hand finishes. Timing alone is not the obstacle: in Pour Seasonings the
two streams overlap in 306 of the 398 episodes where both hit their targets.

\paragraph{Reaching a goal and keeping it are different skills.}
In Open Laptop, Spatial Forcing and OpenWAM bring the display to the requested
angle about equally often (73 and 71 of 100 episodes), but it is still there
at the end in 23 and 67 (Figure~\ref{fig:main-analysis}d). Of the 154 pooled
episodes that lose the angle, 148 end over-opened (median $31.5^\circ$ past),
all with the hand on the display: it is pushed through the target, not
dropped. The Cover Pitcher lid stays inserted in 47 of the 86 episodes that
insert it, and Stack Blocks keeps four layers in 30 of 164, with the hand
already released at 70 of the 134 losses. In Pass Block, the receiver takes
over alone in 188 of 367 episodes where both hands hold the block; without
takeover the block drops in 175 of 179, a median 0.30~s after the giving hand
lets go. Policies must stop pushing at the target and release only when the
other hand holds.

\paragraph{Tools are bypassed, and failures are not shared across checkpoints.}
Line Up Blocks, Scoop Marbles, and Flip Sponge end on direct hand contact with
the manipulated object in 81, 44, and 28\% of episodes (2 successes in
2,400): policies pursue the object with the hand rather than through the tool.
On the 41 fixed-goal tasks at least one checkpoint solves 62.4\% of scenes,
close to the 64.0\% expected under independence (median pairwise $\kappa=0.03$):
the failed scenes differ by checkpoint, so no shared set of hard scenes
explains the failures, although each checkpoint may have its own.

\paragraph{Transfer-16 fails at the same transitions, and new objects fail earlier.}
Every matched transition fails more often than in its Core-50 source
(Figure~\ref{fig:results-row}c). In Spray Shoe grasps form in 232 of 344
episodes, but at most 3 of 43 per checkpoint complete a trigger cycle. On a
solved source, a new object moves failure to acquisition: the left hand picks
up the glue stick in 5 of 400 Pass Glue Stick episodes, against 40--88\% per
checkpoint in Pass Block.

\paragraph{Compositions add a fourth failure: steps are repeated or started early.}
After the laptop step (146 episodes), no episode brings the mouse onto its
pad, and 111 push the display out of range again. The shutter is pressed
before the fruit is in the bucket in 311 of 399 Place Fruit and Photograph
Logo episodes, water runs in 107 of 400 Scrub Bottle Then Fill Cup episodes
although none finishes scrubbing, and the box lid moves before the pitcher lid
in 340 of 400 Cover Pitcher and Box episodes. Nothing marks a finished step,
so completion must be inferred from the scene.

\FloatBarrier
\subsubsection{Task Grouping and Reliability}
\label{sec:reliability}
\begin{figure}[!htbp]
  \centering
  \includegraphics[width=\linewidth]{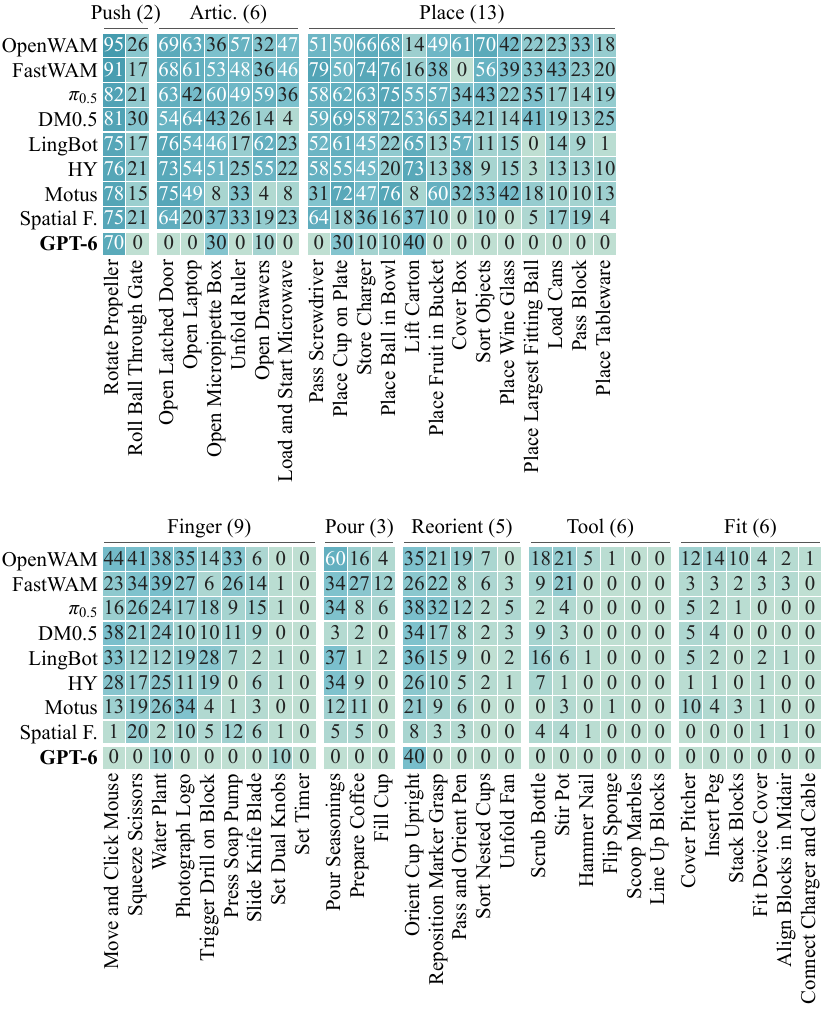}
  \caption{Core-50 success (\%) for eight trained multi-task checkpoints
  and GPT-6 (\texttt{move\_eef}), grouped by hand skill
  (Table~\ref{tab:skill-groups}). Task order is fixed by the trained-checkpoint
  mean. Trained cells have 100 episodes (99 for HY-Embodied on Place Wine
  Glass); GPT-6 cells have ten, so its 10-point increments reflect coarser
  sampling. DP, task IDs, and the same GPT-6 results are in
  Table~\ref{tab:seen-per-task}.}
  \label{fig:task-heatmap}
\end{figure}

Table~\ref{tab:skill-groups} lists the grouping of Core-50 tasks by the hand
skill that decides success, used in Figure~\ref{fig:main-analysis}a and
throughout Section~\ref{sec:results}. Each task is assigned to exactly one group by its
success condition; tasks that also require reaching and grasping are assigned
to the later skill. The grouping is an analysis lens and is independent of the
interaction and structure labels of the catalog.
Figure~\ref{fig:task-heatmap} shows per-task success for the eight trained
checkpoints and GPT-6 in this grouping.

\par\addvspace{\intextsep}
\noindent\begin{minipage}{\linewidth}
  \centering\small\setlength{\tabcolsep}{4pt}
  \captionsetup{hypcap=false}
  \captionof{table}{Hand-skill groups of Core-50 tasks. Pooled: successes over
  all episodes of the eight trained multi-task checkpoints; Best: task mean
  after selecting the best trained checkpoint separately for each task. GPT-6 (\texttt{move\_eef}) is reported
  separately with ten trials/task. Blue/orange row highlights refer to the
  relatively solved/unsolved trained-policy groups.}
  \label{tab:skill-groups}
  \begin{tabularx}{\linewidth}{@{}lXrrr@{}}
    \toprule
    \TabHead\TabStrut Group (tasks) & Tasks & Pooled \% & Best \% & GPT-6 \%\\\TabHeadEnd
    \midrule
    \rowcolor{tabSolved}Nonprehensile (2) & Rotate Propeller, Roll Ball Through Gate & 51.3 & 62.5 & 35.0\\
    \rowcolor{tabSolved}Articulated part (6) & Open Latched Door, Open Laptop, Open Micropipette Box, Open Drawers, Unfold Ruler, Load and Start Microwave & 42.9 & 61.0 & 6.7\\
    \rowcolor{tabSolved}Grasp--transport--release (13) & Place Cup on Plate, Place Ball in Bowl, Place Fruit in Bucket, Place Largest Fitting Ball, Place Tableware, Place Wine Glass, Sort Objects, Load Cans, Pass Block, Pass Screwdriver, Store Charger, Cover Box, Lift Carton & 34.9 & 58.0 & 6.9\\
    Finger actuation (9) & Squeeze Scissors, Water Plant, Trigger Drill on Block, Move and Click Mouse, Press Soap Pump, Set Timer, Set Dual Knobs, Photograph Logo, Slide Knife Blade & 14.0 & 26.2 & 2.2\\
    Pour or fill (3) & Pour Seasonings, Fill Cup, Prepare Coffee & 13.4 & 33.0 & 0.0\\
    In-hand reorientation (5) & Orient Cup Upright, Reposition Marker Grasp, Pass and Orient Pen, Sort Nested Cups, Unfold Fan & 11.4 & 20.2 & 8.0\\
    \rowcolor{tabLow}Tool on object (6) & Hammer Nail, Scoop Marbles, Line Up Blocks, Flip Sponge, Scrub Bottle, Stir Pot & 2.9 & 7.5 & 0.0\\
    \rowcolor{tabLow}Precision fit (6) & Insert Peg, Fit Device Cover, Align Blocks in Midair, Connect Charger and Cable, Stack Blocks, Cover Pitcher & 2.2 & 7.3 & 0.0\\
    \bottomrule
  \end{tabularx}
\end{minipage}\par\addvspace{\intextsep}

\Needspace{3\baselineskip}
For the eight trained multi-task checkpoints, per-task signal-to-noise,
the spread of success across checkpoints divided by
its binomial standard error, is below 1 on 6 tasks and at least 3 on 14.
Twelve tasks selected by this ratio on even-indexed episodes reproduce 96\% of
the 28 pairwise orders measured on odd-indexed episodes (88\% for random
twelve-task subsets), and 24 tasks reproduce all of them. Episode-only 95\%
half-widths of the 50-task macro score are about 1.9, 1.3, and 0.9 points for
25, 50, and 100 episodes per task. Selecting a checkpoint per task on half of
the episodes and scoring it on the other half gives 32.9\% versus 29.1\% for
the best single checkpoint.

An earlier evaluation campaign of the same $\pi_{0.5}$ and OpenWAM
checkpoints used the fixed canonical prompt and a single goal per task; on the
48 shared tasks it gives 24.8 versus 23.4\% ($\pi_{0.5}$) and 30.5 versus
28.5\% (OpenWAM), with per-task Spearman $\rho$ of 0.98 and 0.97 and per-task
differences whose standard deviation (4.3 and 6.3 points) is close to the
binomial expectation (5.1 and 5.5). Within the Core-50 snapshot, success is
not measurably affected by which of the four table materials, three colour
palettes, or two roughness levels is drawn (largest within-task effect 0.8
points, permutation $p\ge0.17$), and training paraphrases score within 2.2
points of the canonical sentence for seven checkpoints (OpenWAM: canonical
$+4.0$ points, $z=2.9$). All appearance factors and paraphrases are inside the
training distribution, so these are invariance checks, not generalization
results.

\FloatBarrier
\section{Discussion and Conclusion}
HandsOn provides a common experimental basis for studying the scope and limits of manipulation competence. By treating embodiment, environment, and task structure as configurable experimental variables, it enables systematic investigation of how policy behavior changes across conditions and where generalization breaks down.

Our trained-policy evaluation spans more than 40,000 Core-50 episodes and includes eight multi-task checkpoints, revealing substantial differences in performance across task categories. Pooled across those eight checkpoints, success rates reach 35–51\% for pushing, articulation, and pick-and-place, but remain at 2–3\% for tool use and precision fitting. These disparities persist after accounting for demonstration count and expert demonstration duration. All tasks in both suites have been successfully demonstrated by human operators, making even the hardest cases concrete, achievable targets for policy improvement.

A prompted GPT-6 controller using \texttt{move\_eef} succeeds in 29 of 660 trials: 5.2\% on Core-50 and 1.9\% on Transfer-16. Its saved action notes provide qualitative signs of task-aware sequencing and execution monitoring without benchmark demonstrations or finetuning, suggesting potential as a high-level planner. Reliable execution remains difficult: the minimal harness requires the same model to translate subgoals into palm poses and up to 40 finger-joint targets with limited geometric and contact feedback. This motivates richer agent interfaces that connect high-level planning to state estimation and feedback-driven manipulation skills. The high dimensionality and contact sensitivity of dexterous tasks make this interface a substantive design problem. Matched harness ablations are needed to quantify the contribution of these components.

The diagnostic results point to a common requirement: policies must regulate the state of an interaction throughout execution. They need to distinguish contacts that stabilize an object from those that actuate it, condition release on whether the partner's hand has secured the object, and maintain achieved goals as the task progresses. Failures on composed tasks further highlight the importance of tracking completed steps and satisfying prerequisites before advancing. Together, these findings motivate learning methods that explicitly account for contact roles, inter-hand dependencies, and task progress.\looseness=-1

Future evaluations should measure both task completion and maintenance of success conditions, testing whether control improvements transfer across embodiments and to physical robots. Promising directions include evaluating VLM planners with richer agent harnesses, expanding embodiment and compositional-task coverage, strengthening paired simulation and real-world evaluation, and modeling contact state, inter-hand coordination, and task progress.

\section{Future Work}
\label{sec:future-work}

\paragraph{Tactile sensing.}
HandsOn policies receive vision and proprioception without tactile feedback, but the
wrist cameras are mounted on the hand and the held object occludes them:
once a grasp forms, the very region the policy must control leaves the
view. The failures we report sit exactly there. A grasp that closes over a
button hides the button as it presses it; a cut that loses a handle contact
happens on the far side of the object; a goal pushed past its target is
obscured by the hand doing the pushing. Tactile sensing would observe these
contact events directly instead of inferring them from a partially hidden
view. Adding it requires taxels placed identically on the simulated and
physical hand, a contact model verified rather than assumed, and tasks that
still admit a visual solution, so that visual and tactile policies remain
comparable on the same predicates.

\paragraph{Quantified physical evaluation.}
The real-robot rollouts in Appendix~\ref{sec:sim_real_complementarity}
show that simulation-trained checkpoints act sensibly on the physical
platform, but they are recorded evidence, not a measurement. A physical
success rate needs the discipline the simulated suite already has: paired
simulation and real instances, an executable predicate on both, a fixed
trial count, and reported failure modes rather than selected successes. The
protocol is specified in Appendix~\ref{sec:direct-transfer}; what remains
is the trial budget to run it alongside simulation-plus-real and real-only
training under an explicit data budget.

\paragraph{Long-horizon household and production tasks.}
HandsOn scores individual skills, and even the compositional Transfer-16
tasks chain at most a few operations. Household and production work is
longer and less forgiving: tens of operations over a changing world state,
objects found rather than placed, and success measured in throughput and
rework rather than a binary predicate. Our failure analysis indicates what
such a suite must measure beyond success, namely whether a policy tracks
which steps are complete, notices that an earlier step went wrong, and can
recover without restarting the episode. That requires long-horizon
demonstrations, a stage model that admits recovery rather than enforcing an
ordered prefix, and evaluation that reports progress and interventions.

\paragraph{Evaluating VLMs across richer agent harnesses.}
The GPT-6 experiment motivates evaluating VLMs with harnesses that add
estimated object poses, contact feedback, reusable manipulation skills,
and hierarchical control to direct \texttt{move\_eef} commands. Comparing
harnesses for a fixed model, and models within a shared harness, would help
distinguish planning capability from interface design. Evaluations should
match tasks, resets, physical-time and query budgets, disclose training of
lower-level skills, and measure success, subgoal completion, recovery, and
inference cost. This would establish when VLMs can serve as effective
high-level planners for dexterous manipulation.

\bibliography{ref}
\bibliographystyle{unsrt}

\clearpage
\appendix

\section{Task Inventory and Taxonomy}
\label{app:task-inventory}

Table~\ref{tab:full-task-inventory} lists all 66 tasks in HandsOn:
Core-50 provides 50 tasks for learning and evaluation; Transfer-16
provides 16 additional tasks exclusively for generalization evaluation.
Transfer-16 contains 6 compositional tasks (EVAL-Comb),
7 generalization tasks (EVAL-Gen), and 3 tasks in the EVAL-Flex family.
Evaluation-only tasks are excluded from the dataset for policy training.
Within each split, task identifiers follow alphabetical order of the English task names.

Source superscripts distinguish task-design adaptation
(J: DexJoCo; V: DexVerse) from asset reuse (JA and VA, respectively).
The sources and modifications are detailed in Appendix~\ref{app:task-provenance}.
Unmarked entries do not imply that every asset or component is original.

The three label columns are the axes of the taxonomy in
Figure~\ref{fig:taxonomy}. Interaction Type records the dominant
manipulation requirement. Asset Type includes all applicable types of
task-relevant assets, including designated targets and receptacles,
according to their current simulation models and active degrees of freedom.
The robot, generic support surfaces, and background assets are excluded.
Task Structure describes the organization of semantic goals and operations;
approaching, grasping, and releasing do not by themselves create separate
stages.


\begingroup
\normalsize
\setlength{\tabcolsep}{3pt}
\renewcommand{\arraystretch}{1}
\setlength{\LTleft}{\fill}
\setlength{\LTright}{\fill}
\setlength{\LTcapwidth}{\textwidth}
\setlength{\LTpre}{\baselineskip}
\setlength{\LTpost}{\baselineskip}

\begin{longtable}{
  >{\raggedright\arraybackslash}p{0.16\textwidth}
  >{\raggedright\arraybackslash}p{0.395\textwidth}
  >{\raggedright\arraybackslash}p{0.13\textwidth}
  >{\raggedright\arraybackslash}
    p{\dimexpr0.18\textwidth-10\tabcolsep\relax}
  >{\raggedright\arraybackslash}p{0.135\textwidth}
}
\caption{Complete task inventory and three-axis taxonomy of HandsOn;
shaded rows are the evaluation-only Transfer-16 tasks.
Abbreviations are defined below the table.}
\label{tab:full-task-inventory}\\
\toprule
\TabHead\TabStrutTop\textbf{Task ID} & \textbf{Task Name} &
\textbf{Interaction}\newline\textbf{Type} &
\textbf{Task}\newline\textbf{Structure} &
\textbf{Asset}\newline\textbf{Type}\TabStrutBot\\\TabHeadEnd
\midrule
\endfirsthead

\caption[]{Complete task inventory and three-axis taxonomy of HandsOn (continued).}\\
\toprule
\TabHead\TabStrutTop\textbf{Task ID} & \textbf{Task Name} &
\textbf{Interaction}\newline\textbf{Type} &
\textbf{Task}\newline\textbf{Structure} &
\textbf{Asset}\newline\textbf{Type}\TabStrutBot\\\TabHeadEnd
\midrule
\endhead

\midrule
\multicolumn{5}{r@{}}{\normalsize Continued on next page}\\
\endfoot

\TabGapBot\bottomrule
\endlastfoot

01 & \mbox{Align Blocks in Midair} & CR & SS & RB \\
02 & \mbox{Connect Charger and Cable} & CR & LH & RB \\
03 & \mbox{Cover Box} & CR & SS & RB \\
04 & \mbox{Cover Pitcher} & CR & SS & RB \\
05 & \mbox{Fill Cup} & F & LH & RB, AR \\
06 & \mbox{Fit Device Cover} & CR & SS & RB \\
07 & \mbox{Flip Sponge} & F & SS & RB \\
08\textsuperscript{J} & \mbox{Hammer Nail} & F & SS & RB, AP \\
09\textsuperscript{JA} & \mbox{Insert Peg} & CR & SS & RB \\
10\textsuperscript{V} & \mbox{Lift Carton} & CR & SS & RB \\
11 & \mbox{Line Up Blocks} & F & RP & RB \\
12\textsuperscript{J} & \mbox{Load and Start Microwave} & F & LH & RB, AR, AP \\
13\textsuperscript{VA} & \mbox{Load Cans} & P & SS & RB \\
14\textsuperscript{J} & \mbox{Move and Click Mouse} & F & LH & RB, AP \\
15 & \mbox{Open Drawers} & P & SS & AP \\
16\textsuperscript{V} & \mbox{Open Laptop} & P & SS & AR \\
17 & \mbox{Open Latched Door} & F & LH & AR \\
18 & \mbox{Open Micropipette Box} & P & SS & AR \\
19 & \mbox{Orient Cup Upright} & P & SS & RB \\
20 & \mbox{Pass and Orient Pen} & CR & LH & RB \\
21 & \mbox{Pass Block} & P & LH & RB \\
22 & \mbox{Pass Screwdriver} & P & LH & RB \\
23\textsuperscript{J} & \mbox{Photograph Logo} & F & SS & RB, AP \\
24 & \mbox{Place Ball in Bowl} & P & SS & RB \\
25 & \mbox{Place Cup on Plate} & P & SS & RB \\
26 & \mbox{Place Fruit in Bucket} & P & SS & RB \\
27 & \mbox{Place Largest Fitting Ball} & P & SS & RB \\
28 & \mbox{Place Tableware} & P & SS & RB \\
29 & \mbox{Place Wine Glass} & P & SS & RB \\
30 & \mbox{Pour Seasonings} & F & SS & RB \\
31 & \mbox{Prepare Coffee} & F & LH & RB, AR, AP \\
32 & \mbox{Press Soap Pump} & F & SS & RB, AP \\
33 & \mbox{Reposition Marker Grasp} & CR & SS & RB \\
34 & \mbox{Roll Ball Through Gate} & P & SS & RB \\
35 & \mbox{Rotate Propeller} & CR & RP & AR \\
36 & \mbox{Scoop Marbles} & F & SS & RB \\
37 & \mbox{Scrub Bottle} & CR & RP & RB \\
38 & \mbox{Set Dual Knobs} & F & SS & AR \\
39 & \mbox{Set Timer} & F & RP & AP \\
40\textsuperscript{V} & \mbox{Slide Knife Blade} & CR & SS & AP \\
41 & \mbox{Sort Nested Cups} & CR & LH & RB \\
42 & \mbox{Sort Objects} & P & LH & RB \\
43 & \mbox{Squeeze Scissors} & F & RP & AR \\
44 & \mbox{Stack Blocks} & P & RP & RB \\
45 & \mbox{Stir Pot} & F & RP & RB \\
46\textsuperscript{VA} & \mbox{Store Charger} & P & LH & RB, AP \\
47 & \mbox{Trigger Drill on Block} & F & SS & RB, AP \\
48 & \mbox{Unfold Fan} & CR & SS & AR \\
49 & \mbox{Unfold Ruler} & CR & SS & AR \\
50\textsuperscript{J} & \mbox{Water Plant} & F & SS & RB, AR \\
\midrule\TabGapTop
\rowcolor{tabAside}\TabStrutTop EVAL-Comb01 & \mbox{Cover Pitcher and Box} & CR & RP & RB \\
\rowcolor{tabAside}EVAL-Comb02 & \mbox{Open Laptop and Click Mouse} & F & LH & RB, AR, AP \\
\rowcolor{tabAside}EVAL-Comb03 & \mbox{Pass Block and Close Box} & CR & LH & RB \\
\rowcolor{tabAside}EVAL-Comb04 & \mbox{Place Fruit and Photograph Logo} & F & LH & RB, AP \\
\rowcolor{tabAside}EVAL-Comb05 & \mbox{Scrub Bottle Then Fill Cup} & F & LH & RB, AR \\
\rowcolor{tabAside}EVAL-Comb06 & \mbox{Season and Stir} & F & LH & RB\TabStrutBot\\\TabGapBot
\midrule\TabGapTop
\rowcolor{tabAside}\TabStrutTop EVAL-Gen01 & \mbox{Cover Lunchbox} & CR & SS & RB \\
\rowcolor{tabAside}EVAL-Gen02 & \mbox{Open Trash Bin} & P & SS & AR \\
\rowcolor{tabAside}EVAL-Gen03 & \mbox{Orient Bottle Upright} & P & SS & RB \\
\rowcolor{tabAside}EVAL-Gen04 & \mbox{Pass Glue Stick} & P & LH & RB \\
\rowcolor{tabAside}EVAL-Gen05 & \mbox{Set Mixer Knobs} & F & RP & AR \\
\rowcolor{tabAside}EVAL-Gen06 & \mbox{Spray Shoe} & F & SS & RB, AR \\
\rowcolor{tabAside}EVAL-Gen07 & \mbox{Stack Boxes} & P & RP & RB\TabStrutBot\\\TabGapBot
\midrule\TabGapTop
\rowcolor{tabAside}\TabStrutTop EVAL-Flex01 & \mbox{Compress Sponge into Mug} & CR & SS & RB, DF \\
\rowcolor{tabAside}EVAL-Flex02 & \mbox{Fold Towel Twice} & CR & RP & DF \\
\rowcolor{tabAside}EVAL-Flex03 & \mbox{Pack Pillow} & CR & SS & RB, DF\TabStrutBot\\
\end{longtable}

\noindent\begin{minipage}{\linewidth}
\textit{Abbreviations.}
\textbf{Interaction Type:} P = Primitive; F = Functional;
CR = Contact-rich.
\textbf{Task Structure:} SS = Single-stage;
RP = Repetitive; LH = Long-horizon (multi-stage).
\textbf{Asset Type:} RB = Rigid body;
AP = Articulated (prismatic); AR = Articulated (revolute);
DF = Deformable.
\end{minipage}
\par\smallskip

\noindent\textit{Classification notes.}
Primitive covers object transport or reorientation whose success is decided
by reaching a target configuration, including pushing, pulling, or rolling an object
and moving an articulated part along its free degree of freedom (e.g.,
opening a drawer or a hinged lid). Functional requires the purposeful use of
a tool, appliance, or control interface (buttons, knobs, levers, faucets,
triggers) to produce an effect. Contact-rich denotes tasks dominated by fine
contact coordination, such as regrasping, precise mating, sustained
contact-force regulation, coordinated bimanual articulation, or controlled
deformation. RB describes each non-articulated rigid asset and does not
limit a task to one object. Rigid links of articulated assets are not
counted separately. Deformable labels follow the current simulation model;
visual liquid effects do not add a DF label.
In Task 08, the nail has a prismatic joint relative to the wood block,
and the hammer is a separate rigid object.
EVAL-Flex03 combines a deformable foam pillow with a rigid carton.
SS denotes a single semantic goal. RP denotes repeated applications of
the same kind of manipulation operation, whereas LH requires distinct, dependent
subgoals. Routine setup and terminal activation, release, or waiting do
not by themselves turn a repetitive task into LH. Task 39 (Set Timer) is
classified as RP because its main manipulation consists of setting the
minute and second values through button operations; starting the timer,
withdrawing the hands, and observing a countdown tick are terminal
completion conditions.
\par
\endgroup

\subsection{Task Provenance and Adaptation}
\label{app:task-provenance}

HandsOn incorporates selected task designs and assets from
DexJoCo~\cite{wang2026dexjoco} and DexVerse~\cite{yao2026dexverse}.
Table~\ref{tab:task-provenance} separates task-design lineage from the reuse of
particular assets. The J and V markers identify documented adaptations;
JA and VA identify asset provenance without asserting that the whole task was
transferred. Shared infrastructure and generic similarities, such as grasping
or stacking, are insufficient to establish task-specific lineage.

\begin{table}[!htbp]
  \centering\small
  \caption{Acknowledged task-design and asset sources. IDs follow
  Table~\ref{tab:full-task-inventory}. All listed tasks use our platform's
  execution and observation pipeline; source attribution does not imply
  identical reset distributions or success predicates.}
  \label{tab:task-provenance}
  \begin{tabularx}{\linewidth}{@{}p{0.08\linewidth}p{0.28\linewidth}X@{}}
    \toprule
    \TabHead\TabStrut\TabL{tabHead}ID & Upstream task or asset & \TabR{tabHead}Adaptation or reuse in HandsOn\\\TabHeadEnd
    \midrule
    08 J & DexJoCo: Hammer nail & Adds a supporting hand holding the wood block and a task-specific nail-depth condition.\\
    12 J & DexJoCo: Bimanual microwave & Adapts the loading, door-closing, and activation sequence to the collection platform.\\
    14 J & DexJoCo: Click mouse & Requires moving the mouse onto its pad before a permitted click.\\
    23 J & DexJoCo: Bimanual photograph & Retains camera aiming and shutter activation with platform-specific contact and execution logic.\\
    50 J & DexJoCo: Water plant & Adds grasp, trigger-contact, and aim-sensitive spray conditions.\\
    \midrule
    10 V & DexVerse: BimanualLiftCarton & Adapts bilateral lifting to lower-side support and a three-second airborne hold.\\
    16 V & DexVerse: OpenLaptop & Adapts the opening task to the ROKAE--Wuji platform; the released task uses a replacement laptop asset and a 60-degree target.\\
    40 V & DexVerse: SlideUtilityKnife & Adapts blade sliding to coordinated hand roles and target-conditioned motion; the collection asset was replaced.\\
    \midrule
    09 JA & IndustReal peg/socket via DexJoCo & Reuses the peg/socket asset family with a separately implemented single-arm insertion task.\\
    13 VA & YCB soup can via DexVerse & Reuses the can asset in a new synchronized cross-loading task with a project-authored partitioned bin.\\
    46 VA & Drawer and charger via DexVerse & Reuses the drawer mechanism and charger; adds a replacement cabinet appearance and a hold--store--release sequence.\\
    \bottomrule
  \end{tabularx}
\end{table}

\paragraph{Compositions and Transfer-16.}
Source credit also follows reused components into new evaluations.
EVAL-Comb02 combines the adapted laptop and mouse skills, EVAL-Comb04
reuses the photograph skill, and EVAL-Gen06 transfers the spray interaction
from Water Plant. These are compositions or task variants of the credited
components, rather than entire tasks taken from an upstream benchmark.
The evaluation-only designation specifies exclusion from policy-training
demonstrations; it does not imply that all constituent skills or assets are new.

\paragraph{Attribution boundaries.}
Task names alone do not establish reuse. For example, Prepare Coffee and
Squeeze Scissors have related counterparts in DexVerse but different
documented workflows or asset lineages; we do not label them as direct ports
on that basis. The source markers identify documented task-design
adaptations and retained asset reuse.

\FloatBarrier
\section{Additional Data, Platform, and Protocol Records}
\label{app:data-protocol}

\subsection{Physical System and Alignment Record}
\label{app:physical-system}
Figure~\ref{fig:protocol}a compares simulated and physical views of the
toy-aircraft setup from the setup, wrist, and~main~cameras.

The ROKAE AR5-08 has seven degrees of freedom, a rated payload of 5 kg,
a 683 mm reach, and manufacturer-specified repeatability of
$\pm0.1$ mm~\cite{rokae_ar5}. The Wuji Hand has 20 actively actuated
joints, four per finger, with position control over USB 2.0
\cite{wuji_hand_v1}. Its specified 1 kHz low-level frequency is distinct
from policy inference and host command rates.

The correspondence record covers hand/robot geometry and limits, asset
scale, camera calibration, state/action coordinate frames, joint/side
ordering, timing, initial conditions, and success.
Calibration data are disclosed independently of target-task demonstrations.

\subsection{Transfer-16 Duration Map}
\label{app:time-budgets}
Table~\ref{tab:eval-budgets} reports the 30 Hz protocol. The source-frame
column sums each mapped atomic task's P75 before multiplying by three.
Task validity is established by the verification in
Section~\ref{sec:task-design}; a repaired task requires fresh
trials under its qualified implementation.
\begin{table}[h]
\centering\small\setlength{\tabcolsep}{3pt}
\caption{Transfer-16 physical-time budgets. Flex budgets (shaded) are native
exceptions and are not described as P75 estimates.}
\label{tab:eval-budgets}
\begin{tabularx}{\linewidth}{@{}lXrrr@{}}
\toprule
\TabHead\TabStrut\TabL{tabHead}Task ID & Supporting Core-50 tasks & P75 frames & Seconds & \TabR{tabHead}Actions\\\TabHeadEnd
\midrule
Comb01 & Cover Pitcher + Cover Box & $382+210$ & 59.2 & 1776\\
Comb02 & Open Laptop + Move and Click Mouse & $196+125$ & 32.1 & 963\\
Comb03 & Pass Block + Cover Box & $195+210$ & 40.5 & 1215\\
Comb04 & Place Fruit in Bucket + Photograph Logo & $110+158$ & 26.8 & 804\\
Comb05 & Scrub Bottle + Fill Cup & $259+530$ & 78.9 & 2367\\
Comb06 & Pour Seasonings + Stir Pot & $236+547$ & 78.3 & 2349\\
Gen01 & Cover Box & 210 & 21.0 & 630\\
Gen02 & Open Micropipette Box & 125 & 12.5 & 375\\
Gen03 & Orient Cup Upright & 74 & 7.4 & 222\\
Gen04 & Pass Block & 195 & 19.5 & 585\\
Gen05 & Set Dual Knobs & 431 & 43.1 & 1293\\
Gen06 & Water Plant & 139 & 13.9 & 417\\
Gen07 & Stack Blocks & 387 & 38.7 & 1161\\
\midrule\TabGapTop
\rowcolor{tabAside}\TabL{tabAside}\TabStrutTop Flex01 & Native exception: Compress Sponge into Mug & -- & 240 & \TabR{tabAside}7200\\
\rowcolor{tabAside}\TabL{tabAside}Flex02 & Native exception: Fold Towel Twice & -- & 180 & \TabR{tabAside}5400\\
\rowcolor{tabAside}\TabL{tabAside}Flex03 & Native exception: Pack Pillow\TabStrutBot & -- & 180 & \TabR{tabAside}5400\\\TabGapBot
\bottomrule
\end{tabularx}
\end{table}
Table~\ref{tab:full-task-inventory} lists the paper task IDs and names.\looseness=-1

\subsection{Task Partitions and Generalization Design}
\label{app:generalization-details}
A task is defined by its environment, initial-state distribution,
observations, actions, success predicate, and physical time budget:
\begin{equation}
  \mathcal{T}_i=(\mathcal{E}_i,p_i(s_0),\mathcal{O}_i,
    \mathcal{A}_i,G_i,H_i).
  \label{eq:task-definition}
\end{equation}
We distinguish training trajectories, held-out trajectories from training
tasks, fresh closed-loop trials on those tasks, and evaluation-only tasks.
Trajectory validation measures action prediction; fresh closed-loop trials
measure execution under the task's native training-source distribution.
Transfer-16 tasks contribute no demonstrations to benchmark
training or task-specific finetuning. Their absence from this dataset does
not establish their absence from a foundation model's external pretraining.

The evaluation catalog contains six EVAL-Comb, seven EVAL-Gen, and three
EVAL-Flex tasks. These implementation groups support complementary
generalization questions rather than a set of perfectly orthogonal factors.
Each task must specify the supporting training skills and assets, the
held-out change, and its executable goal. Multiple changes are identified
when they co-occur.

\paragraph{Asset and relation transfer.}
This dimension tests the use of skills with new objects, asset pairings,
or goal relations. An unseen combination of familiar assets is distinguished
from a new instance or category. For example, covering a new container and
orienting a different object test transfer of related functional behavior,
but whether an asset is actually familiar is determined by source and
asset identities, not by its task name.

\paragraph{Skill transfer and composition.}
These tasks reuse skills in new sequences or contexts. \emph{Open Laptop and
Click Mouse} combines opening an articulated object with interface operation;
\emph{Scrub Bottle Then Fill Cup} connects tool interaction to a subsequent
goal. Evaluation records where an execution stops and whether a successful
stage leaves the state needed by the next stage. A previously unseen atomic
operation is identified separately from a new composition of trained skills.

\paragraph{Transfer to deformable manipulation.}
\emph{Compress Sponge into Mug}, \emph{Fold Towel Twice}, and
\emph{Pack Pillow} require handling deformation without deformable-task
demonstrations in the benchmark training set. These are transfer challenges
across object, goal, and physical behavior. Isolating compliance as a causal
factor would additionally require matched geometry and goals; the present
suite does not assume that all differences are due to material properties.

\begin{table}[t]
  \centering\small
  \caption{Training support and evaluation changes. Relations shown here
  describe task semantics; the source/asset mapping specifies actual exposure.
  Training support refers to the learned methods; GPT-6 receives no HandsOn training.}
  \label{tab:generalization-design}
  \begin{tabularx}{\linewidth}{@{}p{0.18\linewidth}XX@{}}
    \toprule
    \TabHead\TabStrut\TabL{tabHead}Dimension & Training support & \TabR{tabHead}Evaluation question\\\TabHeadEnd
    \midrule
    Asset/relation & Object interaction, covering, orienting, passing
      & Can a skill operate on a new target or asset relation?\\
    Skill composition & Opening a laptop; moving and clicking a mouse;
      scrubbing; filling & Can trained behaviors be connected in a new
      sequence with valid stage transitions?\\
    Deformable tasks & Rigid/articulated object manipulation
      & Can the policy adapt to compression, folding, and packing?\\
    \bottomrule
  \end{tabularx}
\end{table}

\subsection{Visual and Language Robustness Protocols}
\label{sec:language-eval}
Training randomization and evaluation under domain shift serve different
roles. Fresh samples from the training distribution measure in-distribution
performance; held-out combinations or expanded ranges of visual factors
measure robustness under specified shifts. These tests retain the task goal
and identify the factors changed. They are reported separately from the
evaluation-only tasks, which change asset relations, skills, or physical
interactions. Combining both changes is an additional challenge condition,
not a replacement for the matched-distribution baseline.

The eight held-out paraphrases per active goal define a language-only test:
the environment, target values, reset seeds, and success predicate stay
fixed while the instruction changes. Paired trials compare canonical,
training-paraphrase, and held-out-paraphrase success. We report the success
gap and paired outcome agreement, so consistently failing all wordings
cannot masquerade as successful language robustness. This test is separate
from Transfer-16 physical tasks and from visual domain shifts.\looseness=-1

A complementary goal-sensitivity test changes the requested object,
relation, or numerical setting only within a task's supported configurations,
with its executable goal updated accordingly. For example, the timer must
match the requested minutes and seconds; moving a knob must reach the
requested angle and direction. Such tests assess whether the policy uses
the instruction, rather than repeating a familiar motion. Unseen numerical
values require an explicitly held-out goal split. Text-invariant policies
are not scored for language robustness; their structured-goal performance
is reported separately. These additional tests require dedicated trials
and are not inferred from the canonical-prompt evaluations.

\subsection{Matched Physical-Transfer Protocol}
\label{sec:direct-transfer}
\label{app:physical-transfer}
A paired task comprises simulation and real instances with a correspondence
record:
\begin{equation}
  \mathcal{P}_i=(\mathcal{T}^{\mathrm{sim}}_i,
    \mathcal{T}^{\mathrm{real}}_i,\mathcal{C}_i).
  \label{eq:pair-definition}
\end{equation}
The record specifies robot and camera configuration, state/action frames,
control timing, initialization conditions, and success definitions. Direct
transfer deploys the same checkpoint without target-task real demonstration
training or real-data finetuning. Calibration measurements and any real
images used to configure the system are disclosed separately from policy
training. External pretraining of a VLA or WAM retains its own data identity.

Only tasks with physical counterparts enter the paired analysis, which
reports their coverage and per-task simulation and real trial counts.
The signed success-rate difference for model $m$ and task $i$ is defined as\looseness=-1
\begin{equation}
  \Delta_{i,m}=S^{\mathrm{sim}}_{i,m}-S^{\mathrm{real}}_{i,m}.
  \label{eq:transfer-gap}
\end{equation}
{\widowpenalty=10000
Same-task real transfer and Transfer-16 generalization are independent
claims. The recorded real-robot rollouts of this platform are reported
qualitatively (Appendix~\ref{sec:sim_real_complementarity}) and do not
constitute \mbox{a scored physical evaluation.}\par}

\subsection{Additional Task-Capability Examples}

Articulated collection assets occur in 21 tasks: 12 involve revolute joints
and 11 prismatic joints, with two containing both. Goals specify motion
direction, travel, angle, or unlocking prerequisites. Set Timer uses
requested minutes and seconds; Set Dual Knobs uses left/right angle and
direction targets; and Unfold Fan targets 60, 90, or 120 degrees. Fixed
three-cycle scissors and propeller requirements do not by themselves
establish unseen-count generalization. Place Tableware combines identity,
position, and orientation; Place Largest Fitting Ball and Sort Nested Cups
add size comparisons. Prepare Coffee and Load and Start Microwave require
intermediate states that enable later operations. These capability labels
overlap the catalog's 17 primitive, 18 functional, and 15 contact-rich tasks
and its 31 single-stage, seven repetitive, and 12 long-horizon tasks.

\FloatBarrier
\Needspace{8\baselineskip}
\section{GPT-6 \texorpdfstring{\texttt{move\_eef}}{move\_eef} Evaluation Protocol}
\label{app:gpt6-protocol}

\subsection{Model, Task Population, and Observations}
We request \texttt{gpt-6-astra} with reasoning effort \texttt{medium}
through the Responses API, a maximum output of 4,096 tokens per request,
required function calling, and parallel calls disabled. The only exposed
function is \texttt{move\_eef}; the controller returns one call per turn. No
task demonstrations, benchmark finetuning, or learned low-level policy
are used. This describes benchmark-provided exposure; it makes no claim
about the model's external pretraining. Requested model names, returned
model identifiers, provider provenance, and token usage are retained in
the request records rather than inferred from a saved client label.

The fixed batch contains ten slots for each Core-50 and Transfer-16 task,
660 in total. The choice of ten episodes per task reflects the available
API-cost and wall-clock evaluation-time budgets while retaining coverage
of all 66 tasks. This sample-count allocation is separate from the
per-trial model-call cap and simulated physical-time limit. It yields
coarser per-task estimates than the larger trained-policy evaluations;
the uncertainty analysis below accounts for the smaller sample count.
A slot retains its task, goal, seed/accepted-reset record,
prompt, and source identity across attempts. Chinese Chess is excluded.
The declared reset-admission rules and any repaired reset identities remain
in the record; this batch is not assumed to be a seed-paired subsample of
every trained policy. Pilot trajectories do not enter this population.

Each observation supplies the main, left-wrist, and right-wrist RGB views
as native-resolution PNGs with high-detail image input, plus measured
world-frame palm poses, all 20 finger joints per hand, elapsed physical
time, and the remaining physical budget. The robot contract supplies active
arms, joint order, and joint-position bounds. The prompt contains the
official task instruction verbatim and a static task card: a general
completion reminder, simple goal clarification for the three basic
pick-and-place tasks, or the declared carton scoring exception. This
additional text is part of the GPT-6 condition. Object poses, hidden
rewards, and live task-checker state are not supplied to the model.

\subsection{The \texorpdfstring{\texttt{move\_eef}}{move\_eef} Palm-and-Finger Interface}
\label{app:gpt6-motion}
For either active hand, \texttt{move\_eef} accepts world position
\texttt{xyz} in metres, absolute extrinsic-xyz Euler orientation
\texttt{rpy\_deg} in degrees, and 20 absolute \texttt{joints\_rad}
targets. Joints are ordered thumb, index, middle, ring, and little finger,
four per digit. An omitted palm target holds the measured pose; omitted
finger targets retain their last commanded setpoints, preserving a
commanded grip even when contact prevents arrival. Finger targets are
clipped to native joint-position bounds. Inactive arms cannot be commanded.
There is no scalar gripper command or task-specific grasp primitive.

A read-only kinematic planner constructs arm waypoints, paced at at most
0.03 rad per planning step and any stricter native limit. Finger targets
follow quintic interpolation from their previous commanded values, paced
by the native bounds and at most 0.18 of each joint's range per step.
Arm and finger paths share a schedule. These are command-generation
conventions, not guarantees of measured velocity, collision clearance, or
target arrival. The planner has no obstacle model. The adapter passes each
52-dimensional two-hand action (palm position, rotation vector, and
20 joint targets per side) through the benchmark decoder and task-native
executor, retaining native controller and physical actuation behavior.

Every executed action advances physical time by $1/30$ s; the recorded
physics timestep times the substep count must equal this duration.
A \texttt{move\_eef} schedule is limited to 300 actions (10 physical seconds), with
longer plans rejected before execution. A no-op consumes one action.
Malformed or infeasible \texttt{move\_eef} calls receive feedback without advancing
physics and still consume their model calls. Native termination can
interrupt a schedule. A completed schedule means that commands were
executed, not that the target was reached or a grasp succeeded; the next
observation supplies measured state and palm-target error. There is no
\texttt{give\_up} tool and no extension of the task's physical-time budget.
API latency does not advance simulation or imply real-time execution.

\subsection{Context and Interaction Budgets}
\label{app:gpt6-budget}
The controller retains images from the latest two observation turns,
including all three views at each turn. Its compact-context configuration
summarizes executed steps, physical duration, and execution status, removes
duplicate execution feedback, and rounds model-facing numerical state to four
decimal places. Full-precision simulator state, actions, responses, and
execution diagnostics are retained for verification. Earlier image turns
are removed; the recorded text history is retained.

The final controller configuration allows 100 counted model calls per
trial. Invalid or infeasible \texttt{move\_eef} calls consume this budget without
advancing physics; an additional upstream request also consumes a call.
Cached successful replies and replay copies are not counted again.
The reported batch retains the recorded settings of earlier completed
runs, some of which use full text feedback or exceed 100 calls. Thus the
660-trial result describes one fixed evaluation population under its
recorded budgets, not a controlled compact-context or uniform-100-call
ablation. These settings are included in the machine-readable record.

A trial succeeds only when the native task predicate is satisfied. An
unsuccessful call-budget stop counts in the denominator even when the
native environment has not terminated. Such outcomes require an
independent check of task/seed identity, loaded source, 30~Hz execution,
finite recorded state and actions, saved calls, and completed video files;
their acceptance does not certify native terminal completion. Physical
limits follow Section~\ref{sec:evaluation} and
Appendix~\ref{app:time-budgets}, including the three Flex exceptions.

\subsection{Scoring, Verification, and Comparability}
\label{app:gpt6-resume}
The scored population has exactly one outcome for each of the 660 fixed
task/goal/seed slots. Infrastructure interruptions alone are not policy
failures, and no unresolved slot remains in the final report. Result and
video hashes, source and seed records, controller invariance, and physical
clocks were checked on the server for all scored outcomes and 1,980
three-view videos. This check concerns the implemented protocol and file
integrity; it is not a certificate of full training-source alignment.

Two declared qualifications apply to the corresponding results. Open
Latched Door retains the known mismatch between its frozen instruction
(left lever above $45^\circ$, right door) and admitted native source
(right lever at least $25^\circ$, left door). Lift Carton observes contact
force but waives the 12~N magnitude threshold while retaining bilateral
support, lift, hold, and drop conditions; its four successes certify the
lift/hold task, not a 12~N force limit. Line Up Blocks records forbidden
direct hand contact as a task-rule failure, including when the native
return flags have not yet asserted termination.

The controller receives measured robot state and static task guidance in
addition to images. Its trial count, accepted resets, task information,
query timing, and call budget differ from those of the trained policies;
its ten trials per task are not assumed to be seed-paired with every
checkpoint. The comparison is consequently descriptive. It does not
identify an architectural effect or equate external pretraining, task
information, or inference compute across model families.

\subsection{Aggregation and Uncertainty}
Success is the unweighted mean of per-task success rates within each
suite or task family. Every task has ten trials, so the macro rate equals
pooled successes divided by trials. The overall 29/660 is provided for
coverage accounting; the Core-50 and Transfer-16 scores remain separate
because their task populations answer different questions.

Table~\ref{tab:gpt6-results} gives 95\% percentile intervals from 20,000
within-task bootstrap resamples: for each task, redraw ten binary outcomes
from its observed success rate, retain the task catalog, and recompute the
macro rate. The intervals quantify conditional finite-rollout variation;
they exclude task-selection, provider/model-version, and protocol-setting
uncertainty. Table~\ref{tab:seen-intervals} additionally reports a
2,000-resample task-and-episode interval for Core-50, redrawing tasks and
then ten outcomes per selected task from its observed rate. Ten trials
give only coarse per-task resolution, and an
observed 0/10 does not imply zero underlying success probability.

\clearpage
\subsection{Per-Task GPT-6 Results}
\label{app:gpt6-tasks}
Each task contributes ten scored trials. The identifiers and names below follow Table~\ref{tab:full-task-inventory}; runtime directory aliases are resolved through the machine-readable task map. Budget stops contribute to the denominator and are unsuccessful outcomes. A measured 0/10 is not evidence that the true success probability is zero. The Lift Carton and Open Latched Door qualifications in Appendix~\ref{app:gpt6-resume} apply.
\begingroup\small\setlength{\tabcolsep}{5pt}
\begin{longtable}{@{}lp{.43\textwidth}rrr@{}}
\caption{GPT-6 with \texttt{move\_eef}: per-task results. All rows have ten trials; Cap gives the number stopped by the applicable call budget. Positive success entries are shaded.}\label{tab:gpt6-per-task}\\
\toprule
ID & Task & Successes & SR (\%) & Cap\\
\midrule
\endfirsthead
\caption[]{GPT-6 with \texttt{move\_eef}: per-task results (continued).}\\
\toprule
ID & Task & Successes & SR (\%) & Cap\\
\midrule
\endhead
\midrule\multicolumn{5}{r}{Continued on next page}\\
\endfoot
\bottomrule
\endlastfoot
\multicolumn{5}{@{}l}{\textbf{Core-50}}\\
\midrule
01 & Align Blocks in Midair & 0 / 10 & 0 & 5\\
02 & Connect Charger and Cable & 0 / 10 & 0 & 10\\
03 & Cover Box & 0 / 10 & 0 & 0\\
04 & Cover Pitcher & 0 / 10 & 0 & 6\\
05 & Fill Cup & 0 / 10 & 0 & 10\\
06 & Fit Device Cover & 0 / 10 & 0 & 10\\
07 & Flip Sponge & 0 / 10 & 0 & 7\\
08 & Hammer Nail & 0 / 10 & 0 & 10\\
09 & Insert Peg & 0 / 10 & 0 & 10\\
10 & Lift Carton & 4 / 10 & \cellcolor{tabSolved}40 & 1\\
11 & Line Up Blocks & 0 / 10 & 0 & 0\\
12 & Load and Start Microwave & 0 / 10 & 0 & 9\\
13 & Load Cans & 0 / 10 & 0 & 1\\
14 & Move and Click Mouse & 0 / 10 & 0 & 0\\
15 & Open Drawers & 1 / 10 & \cellcolor{tabSolved}10 & 0\\
16 & Open Laptop & 0 / 10 & 0 & 0\\
17 & Open Latched Door & 0 / 10 & 0 & 0\\
18 & Open Micropipette Box & 3 / 10 & \cellcolor{tabSolved}30 & 0\\
19 & Orient Cup Upright & 4 / 10 & \cellcolor{tabSolved}40 & 0\\
20 & Pass and Orient Pen & 0 / 10 & 0 & 9\\
21 & Pass Block & 0 / 10 & 0 & 0\\
22 & Pass Screwdriver & 0 / 10 & 0 & 0\\
23 & Photograph Logo & 0 / 10 & 0 & 0\\
24 & Place Ball in Bowl & 1 / 10 & \cellcolor{tabSolved}10 & 0\\
25 & Place Cup on Plate & 3 / 10 & \cellcolor{tabSolved}30 & 0\\
26 & Place Fruit in Bucket & 0 / 10 & 0 & 0\\
27 & Place Largest Fitting Ball & 0 / 10 & 0 & 1\\
28 & Place Tableware & 0 / 10 & 0 & 9\\
29 & Place Wine Glass & 0 / 10 & 0 & 0\\
30 & Pour Seasonings & 0 / 10 & 0 & 0\\
31 & Prepare Coffee & 0 / 10 & 0 & 10\\
32 & Press Soap Pump & 0 / 10 & 0 & 7\\
33 & Reposition Marker Grasp & 0 / 10 & 0 & 7\\
34 & Roll Ball Through Gate & 0 / 10 & 0 & 0\\
35 & Rotate Propeller & 7 / 10 & \cellcolor{tabSolved}70 & 0\\
36 & Scoop Marbles & 0 / 10 & 0 & 8\\
37 & Scrub Bottle & 0 / 10 & 0 & 1\\
38 & Set Dual Knobs & 1 / 10 & \cellcolor{tabSolved}10 & 5\\
39 & Set Timer & 0 / 10 & 0 & 10\\
40 & Slide Knife Blade & 0 / 10 & 0 & 9\\
41 & Sort Nested Cups & 0 / 10 & 0 & 4\\
42 & Sort Objects & 0 / 10 & 0 & 10\\
43 & Squeeze Scissors & 0 / 10 & 0 & 0\\
44 & Stack Blocks & 0 / 10 & 0 & 2\\
45 & Stir Pot & 0 / 10 & 0 & 10\\
46 & Store Charger & 1 / 10 & \cellcolor{tabSolved}10 & 0\\
47 & Trigger Drill on Block & 0 / 10 & 0 & 9\\
48 & Unfold Fan & 0 / 10 & 0 & 1\\
49 & Unfold Ruler & 0 / 10 & 0 & 8\\
50 & Water Plant & 1 / 10 & \cellcolor{tabSolved}10 & 0\\
\midrule
\multicolumn{5}{@{}l}{\textbf{Transfer-16}}\\
\midrule
EVAL-Comb01 & Cover Pitcher and Box & 0 / 10 & 0 & 10\\
EVAL-Comb02 & Open Laptop and Click Mouse & 0 / 10 & 0 & 0\\
EVAL-Comb03 & Pass Block and Close Box & 0 / 10 & 0 & 6\\
EVAL-Comb04 & Place Fruit and Photograph Logo & 0 / 10 & 0 & 0\\
EVAL-Comb05 & Scrub Bottle Then Fill Cup & 0 / 10 & 0 & 10\\
EVAL-Comb06 & Season and Stir & 0 / 10 & 0 & 5\\
EVAL-Gen01 & Cover Lunchbox & 0 / 10 & 0 & 0\\
EVAL-Gen02 & Open Trash Bin & 2 / 10 & \cellcolor{tabSolved}20 & 0\\
EVAL-Gen03 & Orient Bottle Upright & 1 / 10 & \cellcolor{tabSolved}10 & 0\\
EVAL-Gen04 & Pass Glue Stick & 0 / 10 & 0 & 0\\
EVAL-Gen05 & Set Mixer Knobs & 0 / 10 & 0 & 4\\
EVAL-Gen06 & Spray Shoe & 0 / 10 & 0 & 0\\
EVAL-Gen07 & Stack Boxes & 0 / 10 & 0 & 2\\
EVAL-Flex01 & Compress Sponge into Mug & 0 / 10 & 0 & 5\\
EVAL-Flex02 & Fold Towel Twice & 0 / 10 & 0 & 9\\
EVAL-Flex03 & Pack Pillow & 0 / 10 & 0 & 10\\
\midrule
\multicolumn{2}{l}{All tasks} & 29 / 660 & 4.4 & 260\\
\end{longtable}
\endgroup

\section{Core-50 Results and Coverage}
\label{app:seen-results}
The trained-policy rows in Table~\ref{tab:seen-results} are computed from 45,000 episodes on all 50
Core-50 tasks. The eight multi-task checkpoints contribute 100 episodes per
task (39,999 valid; HY-Embodied has 99 valid episodes on Place Wine Glass),
and single-task DP contributes 5,000 episodes from task-specific checkpoints.
Overall success is the unweighted mean of the 50 task success rates;
interaction-type and hand-skill means use the same rule within each group.
Chinese Chess remains outside the catalog. GPT-6 (\texttt{move\_eef}) has
ten trials per task and appears separately in the uncertainty and per-task
tables below; its outcomes are not pooled into trained-policy aggregates.
Table~\ref{tab:gpt6-per-task} also lists its call-budget stops and Transfer-16
results.

\paragraph{Scope and validity.}
All multi-task evaluations passed the recorded runtime validation and
match the selected checkpoints; the DP checks are described below. The evaluator samples each reset from the admitted source
templates of its task, so the scores describe this implemented reset
distribution. Native reset rejection is handled before policy scoring by
selecting the first accepted candidate while retaining the assigned goal and
language; rejected resets and infrastructure interruptions are not policy
failures. On 41 tasks, the eight multi-task checkpoints share the reset seed,
variant, goal, appearance, and prompt at nearly every episode index; on the
nine goal-conditioned tasks, the goal and prompt are sampled per checkpoint.

\paragraph{Replacement and scoring records.}
Exact original-slot associations include the recovered minority-goal
Fit Device Cover episodes and the recovered DM0.5 Place Wine Glass episode;
no slot is counted twice. Each checkpoint contributes 100 Lift Carton
episodes evaluated using the following criteria: both hands support opposite
lower sides of the carton, the entire carton clears the table, and it remains
continuously airborne for three seconds.

\paragraph{Single-task DP reference.}
DP uses a distinct checkpoint at 30,000 updates for each task. Ten tasks use
completed retraining with corrected near-degenerate normalization; the other
tasks retain their previously selected task checkpoints. Lift Carton is
evaluated in a separate batch using the lift/hold criteria described above;
its result is included in the 50-task score but excluded from the 11-task
matched comparison. Original and retrained model identities
are preserved per task. The 38 earlier evaluations use the earlier native
goal sampler, whereas the newer runs use frozen training-frame-weighted
goal quotas, so DP is a descriptive reference rather than a
goal-distribution-matched comparison. Three rejected knife-task resets are omitted under the original
first-100-accepted-seeds rule. The DP audit checks checkpoint hashes, native
completion, native success, and recorded three-view video completion.

The accompanying machine-readable record stores the task map, checkpoint
identities, source plans, supplemental-slot associations, and the hash of
the episode-level snapshot.

\paragraph{Uncertainty.}
Table~\ref{tab:seen-intervals} reports two 95\% percentile bootstrap
intervals. The episode-only interval (20,000 resamples) redraws each task's
outcomes from its observed success rate with the 50 tasks fixed; it
quantifies finite rollout sampling under an independent-episode assumption.
The task-and-episode interval (2,000 resamples) also resamples the 50 tasks
with replacement,
so it describes a new draw of tasks from the same catalog. Neither interval
includes training-run or checkpoint-selection uncertainty.\looseness=-1
\begin{table}[ht]
\centering\small
\caption{Core-50 task-macro success on all 50 tasks with 95\% bootstrap
intervals over episodes only and over tasks and episodes. GPT-6 uses ten
trials/task; its task-and-episode interval uses the same 2,000-resample
procedure. These intervals describe sampling uncertainty, not equivalence
of model inputs or query budgets. Shaded: best trained checkpoint.}
\label{tab:seen-intervals}
\begin{tabularx}{\linewidth}{>{\raggedright\arraybackslash}Xrrrr}
\toprule
\TabHead\TabStrut Model & Successes / episodes & SR (\%) & Episodes CI (\%) & Tasks+episodes CI (\%)\\\TabHeadEnd
\midrule\TabGapTop
\TabStrutTop OpenWAM & 1,453 / 5,000 & \TabBest{29.1} & [28.0, 30.1] & [22.8, 36.0]\\
FastWAM & 1,319 / 5,000 & 26.4 & [25.4, 27.4] & [20.0, 33.8]\\
$\pi_{0.5}$ & 1,243 / 5,000 & 24.9 & [23.8, 25.9] & [18.6, 31.7]\\
DM0.5 & 1,072 / 5,000 & 21.4 & [20.5, 22.4] & [15.2, 28.1]\\
LingBot-VLA & 984 / 5,000 & 19.7 & [18.8, 20.6] & [13.6, 26.2]\\
HY-Embodied & 947 / 4,999 & 18.9 & [18.0, 19.8] & [13.2, 25.4]\\
Motus & 903 / 5,000 & 18.1 & [17.2, 18.9] & [12.3, 25.0]\\
Spatial Forcing & 620 / 5,000 & 12.4 & [11.6, 13.2] & [7.9, 18.0]\\
DP (single-task) & 483 / 5,000 & 9.7 & [9.0, 10.3] & [5.0, 14.9]\\
\midrule
GPT-6 (\texttt{move\_eef}) & 26 / 500 & 5.2 & [3.6, 6.8] & [1.6, 9.8]\\
\bottomrule
\end{tabularx}
\end{table}

\clearpage
\begingroup
\small\setlength{\tabcolsep}{2.0pt}
\renewcommand{\arraystretch}{1.06}
\begin{longtable}{>{\raggedleft\arraybackslash}p{12pt}p{0.30\textwidth}*{10}{>{\raggedleft\arraybackslash}p{\dimexpr(0.70\textwidth-24\tabcolsep-12pt)/10\relax}}}
\caption{Per-task success rates (\%) on the 50 Core-50 tasks. Each cell has
100 episodes for trained checkpoints, so those cells also give success
counts; GPT denotes GPT-6 (\texttt{move\_eef}), with ten trials/cell, and its
values are percentages, not counts. $^\ast$HY-Embodied
has 99 valid episodes on Place Wine Glass. Columns use the selected
checkpoints from Section~\ref{sec:experimental-setup}; OW, FW, DM, LB, HY, and SF
denote OpenWAM, FastWAM, DM0.5, LingBot-VLA, HY-Embodied, and Spatial Forcing. The
DP sampler caveat above applies to every DP cell. Shaded: best trained
checkpoint on each task, ties included. GPT-6 is a descriptive comparison
with different inputs and query budgets.}
\label{tab:seen-per-task}\\
\toprule
\TabHead\TabStrut ID & Task & OW & FW & $\pi_{0.5}$ & DM & LB & HY & Motus & SF & DP & GPT\\\TabHeadEnd
\midrule\TabGapTop
\endfirsthead
\caption[]{Per-task success rates (continued).}\\
\toprule
\TabHead\TabStrut ID & Task & OW & FW & $\pi_{0.5}$ & DM & LB & HY & Motus & SF & DP & GPT\\\TabHeadEnd
\midrule
\endhead
\midrule\multicolumn{12}{r}{Continued on next page}\\
\endfoot
\TabGapBot\bottomrule
\endlastfoot
\TabStrutTop 01 & Align Blocks in Midair & 2 & \cellcolor{tabBest}3 & 0 & 0 & 1 & 0 & 0 & 1 & 0 & 0\\
02 & Connect Charger and Cable & \cellcolor{tabBest}1 & 0 & 0 & 0 & 0 & 0 & 0 & 0 & 0 & 0\\
03 & Cover Box & \cellcolor{tabBest}61 & 0 & 34 & 34 & 57 & 38 & 32 & 0 & 0 & 0\\
04 & Cover Pitcher & \cellcolor{tabBest}12 & 3 & 5 & 5 & 5 & 1 & 10 & 0 & 2 & 0\\
05 & Fill Cup & 4 & \cellcolor{tabBest}12 & 6 & 0 & 2 & 0 & 0 & 0 & 0 & 0\\
06 & Fit Device Cover & \cellcolor{tabBest}4 & 3 & 0 & 0 & 2 & 1 & 1 & 1 & 0 & 0\\
07 & Flip Sponge & \cellcolor{tabBest}1 & 0 & 0 & 0 & 0 & 0 & \cellcolor{tabBest}1 & 0 & 0 & 0\\
08 & Hammer Nail & \cellcolor{tabBest}5 & 0 & 0 & 0 & 1 & 0 & 0 & 1 & 0 & 0\\
09 & Insert Peg & \cellcolor{tabBest}14 & 3 & 2 & 4 & 2 & 1 & 4 & 0 & 0 & 0\\
10 & Lift Carton & 14 & 16 & 55 & 53 & 65 & \cellcolor{tabBest}73 & 8 & 37 & 51 & 40\\
11 & Line Up Blocks & 0 & 0 & 0 & 0 & 0 & 0 & 0 & 0 & 0 & 0\\
12 & Load and Start Microwave & \cellcolor{tabBest}47 & 46 & 36 & 4 & 23 & 22 & 8 & 23 & 0 & 0\\
13 & Load Cans & 23 & \cellcolor{tabBest}43 & 17 & 19 & 14 & 13 & 10 & 17 & 0 & 0\\
14 & Move and Click Mouse & \cellcolor{tabBest}44 & 23 & 16 & 38 & 33 & 28 & 13 & 1 & 29 & 0\\
15 & Open Drawers & 32 & 36 & 59 & 14 & \cellcolor{tabBest}62 & 55 & 4 & 19 & 29 & 10\\
16 & Open Laptop & 63 & 61 & 42 & \cellcolor{tabBest}64 & 54 & 54 & 49 & 20 & 23 & 0\\
17 & Open Latched Door & 69 & 68 & 63 & 54 & \cellcolor{tabBest}76 & 73 & 75 & 64 & 53 & 0\\
18 & Open Micropipette Box & 36 & 53 & \cellcolor{tabBest}60 & 43 & 46 & 51 & 8 & 37 & 46 & 30\\
19 & Orient Cup Upright & 35 & 26 & \cellcolor{tabBest}38 & 34 & 36 & 26 & 21 & 8 & \cellcolor{tabBest}38 & 40\\
20 & Pass and Orient Pen & \cellcolor{tabBest}19 & 8 & 12 & 8 & 9 & 5 & 6 & 3 & 0 & 0\\
21 & Pass Block & \cellcolor{tabBest}33 & 23 & 14 & 13 & 9 & 13 & 10 & 19 & 2 & 0\\
22 & Pass Screwdriver & 51 & \cellcolor{tabBest}79 & 58 & 59 & 52 & 58 & 31 & 64 & 23 & 0\\
23 & Photograph Logo & \cellcolor{tabBest}35 & 27 & 17 & 10 & 19 & 11 & 34 & 10 & 4 & 0\\
24 & Place Ball in Bowl & 68 & \cellcolor{tabBest}76 & 75 & 72 & 22 & 20 & \cellcolor{tabBest}76 & 16 & 6 & 10\\
25 & Place Cup on Plate & 50 & 50 & 62 & 69 & 61 & 55 & \cellcolor{tabBest}72 & 18 & 25 & 30\\
26 & Place Fruit in Bucket & 49 & 38 & 57 & \cellcolor{tabBest}65 & 13 & 13 & 60 & 10 & 6 & 0\\
27 & Place Largest Fitting Ball & 22 & 33 & 35 & \cellcolor{tabBest}41 & 0 & 3 & 18 & 5 & 0 & 0\\
28 & Place Tableware & 18 & 20 & 19 & \cellcolor{tabBest}25 & 1 & 10 & 13 & 4 & 0 & 0\\
29 & Place Wine Glass$^\ast$ & \cellcolor{tabBest}42 & 39 & 22 & 14 & 15 & 15 & \cellcolor{tabBest}42 & 0 & 9 & 0\\
30 & Pour Seasonings & \cellcolor{tabBest}60 & 34 & 34 & 3 & 37 & 34 & 12 & 5 & 1 & 0\\
31 & Prepare Coffee & 16 & \cellcolor{tabBest}27 & 8 & 2 & 1 & 9 & 11 & 5 & 0 & 0\\
32 & Press Soap Pump & \cellcolor{tabBest}33 & 26 & 9 & 11 & 7 & 0 & 1 & 12 & 0 & 0\\
33 & Reposition Marker Grasp & 21 & 22 & \cellcolor{tabBest}32 & 17 & 15 & 10 & 9 & 3 & 5 & 0\\
34 & Roll Ball Through Gate & 26 & 17 & 21 & \cellcolor{tabBest}30 & 17 & 21 & 15 & 21 & 8 & 0\\
35 & Rotate Propeller & \cellcolor{tabBest}95 & 91 & 82 & 81 & 75 & 76 & 78 & 75 & 79 & 70\\
36 & Scoop Marbles & 0 & 0 & 0 & 0 & 0 & 0 & 0 & 0 & 0 & 0\\
37 & Scrub Bottle & 18 & 9 & 2 & 9 & 16 & 7 & 0 & 4 & \cellcolor{tabBest}22 & 0\\
38 & Set Dual Knobs & 0 & \cellcolor{tabBest}1 & \cellcolor{tabBest}1 & 0 & \cellcolor{tabBest}1 & \cellcolor{tabBest}1 & 0 & \cellcolor{tabBest}1 & 0 & 10\\
39 & Set Timer & 0 & 0 & 0 & 0 & 0 & 0 & 0 & 0 & 0 & 0\\
40 & Slide Knife Blade & 6 & 14 & \cellcolor{tabBest}15 & 9 & 2 & 6 & 3 & 6 & 0 & 0\\
41 & Sort Nested Cups & \cellcolor{tabBest}7 & 6 & 2 & 2 & 0 & 2 & 0 & 0 & 0 & 0\\
42 & Sort Objects & \cellcolor{tabBest}70 & 56 & 43 & 21 & 11 & 9 & 33 & 10 & 0 & 0\\
43 & Squeeze Scissors & \cellcolor{tabBest}41 & 34 & 26 & 21 & 12 & 17 & 19 & 20 & 2 & 0\\
44 & Stack Blocks & \cellcolor{tabBest}10 & 2 & 1 & 0 & 0 & 0 & 3 & 0 & 0 & 0\\
45 & Stir Pot & \cellcolor{tabBest}21 & \cellcolor{tabBest}21 & 4 & 3 & 6 & 1 & 3 & 4 & 0 & 0\\
46 & Store Charger & 66 & \cellcolor{tabBest}74 & 63 & 58 & 45 & 45 & 47 & 36 & 9 & 10\\
47 & Trigger Drill on Block & 14 & 6 & 18 & 10 & \cellcolor{tabBest}28 & 19 & 4 & 5 & 2 & 0\\
48 & Unfold Fan & 0 & 3 & \cellcolor{tabBest}5 & 3 & 2 & 1 & 0 & 0 & 0 & 0\\
49 & Unfold Ruler & \cellcolor{tabBest}57 & 48 & 49 & 26 & 17 & 25 & 33 & 33 & 2 & 0\\
50 & Water Plant & 38 & \cellcolor{tabBest}39 & 24 & 24 & 12 & 25 & 26 & 2 & 7 & 10\\
\midrule
\multicolumn{2}{l}{\TabStrutTop Task-macro mean} & 29.1 & 26.4 & 24.9 & 21.4 & 19.7 & 18.9 & 18.1 & 12.4 & 9.7 & 5.2\TabStrutBot\\
\end{longtable}
\endgroup

\section{Single-Task Demonstration Scaling}
\label{app:data-scaling}
\paragraph{Training controls.}
Tasks~24 and~43 use independent $\pi_{0.5}$ finetuning runs initialized from
the same generic pretrained base with a freshly initialized 52-dimensional
action interface. All runs use full finetuning, batch size 32, training
seed 0, and exactly 20,000 optimizer updates. The learning rate warms up
for 1,333 updates to $5\times10^{-5}$ and decays to $5\times10^{-6}$.
The 10/50/100/200-trajectory subsets are a single frozen nested draw;
sampling for Task~24 is stratified approximately in proportion to the three ball types.
The full tier contains every task-local training trajectory: 438 for Task~24
(192 tennis, 175 mini soccer, 71 golf) and 439 for Task~43. The 62 held-out
trajectories for Task~24 and 61 for Task~43 are excluded at every size. Images, augmentation,
state/action semantics, the prompt bank, and the 20/80
canonical/paraphrase training sampler are unchanged across sizes.

Task-wide normalization statistics are fixed across sizes and computed
from the full training pool, excluding validation. Thus small-data runs
have access to larger-pool summary statistics; this is not a strictly
subset-only data-access experiment. Fixed update counts also imply
different reuse rates across dataset sizes. The selected checkpoint is
always update 20,000, not the checkpoint with the highest test success.
Full-inference offline validation runs every 1,000 updates on fixed anchors;
the held-out trajectories for Task~24 contain golf balls only, so this validation
does not represent the complete ball distribution. The curves here report
fresh closed-loop success, not offline reconstruction error.\looseness=-1

\paragraph{Closed-loop evaluation.}
Each checkpoint has 100 accepted, runtime-valid episodes, giving 1,000
episodes in total with no missing slots. Sizes share the same planned
task-specific reset seeds, goals, and prompts. Native reset rejection may
advance an accepted seed; shared planned draws do not certify identical
complete physical states. Evaluation keeps the native EEF controller,
30 Hz action execution, the accepted rendering revision, and checkpoint
normalization. The physical budgets are $3\times$P75: 321 actions
(10.7 s) for Task~24 and 663 (22.1 s) for Task~43. Inference uses a 30-action
horizon and ten denoising steps. The prompt bank and goal distribution
are fixed across training sizes rather than refitted to each subset.
Task~24 evaluates 42 tennis, 40 mini-soccer, and 18 golf goals; Task~43 requires
lifting the scissors and completing three cuts in the air.

\begin{table}[htbp]
  \centering\small
  \caption{Single-task scaling results. Success counts are out of 100
  trials per cell; intervals are pointwise 95\% Wilson intervals in percent.
  Full-pool sizes are given for Tasks~24 and~43, \mbox{respectively. Shaded: best.}}
  \label{tab:data-scaling}
  \begin{tabularx}{\linewidth}{>{\raggedright\arraybackslash}X*{4}{>{\raggedleft\arraybackslash}X}}
    \toprule
    \TabHead\TabStrut Trajectories & Task 24 $s/n$ & 95\% CI & Task 43 $s/n$ & 95\% CI\\\TabHeadEnd
    \midrule
    10 & 7/100 & [3.4, 13.7] & 1/100 & [0.2, 5.4]\\
    50 & 23/100 & [15.8, 32.2] & 11/100 & [6.3, 18.6]\\
    100 & 31/100 & [22.8, 40.6] & 13/100 & [7.8, 21.0]\\
    200 & 65/100 & [55.3, 73.6] & 18/100 & [11.7, 26.7]\\
    Full (438 / 439)\TabStrutBot & \TabBest{83/100} & [74.5, 89.1] & \TabBest{44/100} & [34.7, 53.8]\\\TabGapBot
    \bottomrule
  \end{tabularx}
\end{table}

Figure~\ref{fig:data-scaling} and Table~\ref{tab:data-scaling} use raw
success fractions under these fixed evaluation draws. Reweighting the
per-goal success of Task~24 by the frozen training-frame prior instead gives
6.99\%, 22.89\%, 30.88\%, 64.67\%, 82.80\%, in increasing data-size order;
Task~43 has one goal and is unchanged. The intervals describe approximate
rollout-level binomial uncertainty conditional on each checkpoint; they
do not estimate training-seed or subset-selection variance and are not
simultaneous confidence bands. There is only one training seed and one
subset draw at each size.

\paragraph{Result scope and provenance.}
The runtime checks pass. As for Core-50 (Appendix~\ref{app:seen-results}),
the evaluator reuses admitted source templates, so these curves describe
the implemented closed-loop reset distribution rather than every
training-source cohort. Machine-readable counts, result hashes, checkpoint
steps, goal strata, and evidence-manifest hashes accompany the manuscript
in the ancillary file \texttt{anc/scaling\_results\_20260926.json}.

\section{Qualitative Real-Robot Rollouts}
\label{sec:sim_real_complementarity}
The videos released with this work show real-robot rollouts of checkpoints
adapted using this dataset's simulation demonstrations, without target-task
real-robot demonstrations
(Section~\ref{sec:real-platform}); they are available at
\url{https://github.com/TsinghuaC3I/HandsOn}. Two task families are shown: single-hand
reach-and-place interactions and bimanual tasks that require both hands to
hold their targets at once.\looseness=-1

We include these recordings as qualitative evidence that the simulated
demonstrations transfer to a physical dexterous hand: the policies establish
contact with the intended object, carry it, and preserve the object pose
through the hand transitions that the simulated episodes also rely on. The
recordings are not a scored evaluation. They cover a small number of
episodes on one physical platform, they are not paired with a controlled
simulation-plus-real or real-only training condition under a matched data
budget, and no success rate is reported. We therefore do not use them to
quantify physical transfer, and we list a budget-controlled version of this
comparison as future work (Section~\ref{sec:future-work}).



\clearpage
\section{Task Profiles}
\label{app:task-profiles}

Table~\ref{tab:task-profiles} follows the task order in
Table~\ref{tab:full-task-inventory}.
Each row presents the task identifier and name, a screenshot,
and the task instruction; the interaction, task-structure, and asset labels of
each task are listed in Table~\ref{tab:full-task-inventory}.
Tasks 01--50 are used for both training and evaluation;
the 16 tasks with EVAL-prefixed identifiers form Transfer-16.
Source superscripts match Table~\ref{tab:full-task-inventory};
Appendix~\ref{app:task-provenance} explains task adaptations and asset reuse.
For training tasks, we show one canonical prompt from the training set;
for Transfer-16 tasks, we show the configured task prompt.
When a task has multiple target-specific configurations, one instance is shown.

Within the Task Instruction column, \textbf{bold text} marks the randomly initialized task parameters, such as target objects, spatial relations, orientations, angles, durations, and action choices.
Concrete values show one example configuration.
Visual randomization that does not alter the instruction is not marked.
Meaning-preserving prompt variants are a separate language dimension
(Section~\ref{sec:language-bank}); the instructions here do not enumerate the 32 training and eight
held-out paraphrases associated with each goal-specific canonical instruction.

\begingroup
\normalsize

\newcommand{\DexProfileCenter}[1]{%
  \raisebox{\dimexpr(\depth-\height)/2\relax}{#1}%
}

\newcommand{\DexProfileImage}[1]{%
  \IfFileExists{figures/tasks/#1.png}{%
    \includegraphics[
      width=0.30\textwidth,
      height=0.16875\textwidth,
      keepaspectratio
    ]{figures/tasks/#1.png}%
  }{%
    \begingroup
    \setlength{\fboxsep}{0pt}%
    \setlength{\fboxrule}{0.4pt}%
    \fcolorbox{black!20}{black!3}{%
      \parbox[c][\dimexpr0.16875\textwidth-2\fboxrule\relax][c]
        {\dimexpr0.30\textwidth-2\fboxrule\relax}{%
        \centering\normalfont\normalsize
        Screenshot\par placeholder}}%
    \endgroup
  }%
}

\newcommand{\DexProfileRow}[3]{%
  \DexProfileCenter{%
    \parbox{0.25\textwidth}{%
      \raggedright\textbf{#1}\par #2}}%
  &
  \DexProfileCenter{%
    \makebox[0.30\textwidth][c]{\DexProfileImage{#1}}}%
  &
  \DexProfileCenter{%
    \parbox{\dimexpr0.45\textwidth-12pt\relax}{%
      \raggedright #3}}%
  \tabularnewline\noalign{\vskip 6pt}
}

\renewcommand{\arraystretch}{1}
\setlength{\LTleft}{\fill}
\setlength{\LTright}{\fill}
\setlength{\LTcapwidth}{\textwidth}
\setlength{\LTpre}{\baselineskip}
\setlength{\LTpost}{\baselineskip}

\begin{longtable}{@{}l@{\hspace{6pt}}l@{\hspace{6pt}}l@{}}
\caption{Task profiles of HandsOn.}
\label{tab:task-profiles}\\
\toprule
\TabHead\TabStrut\TabL{tabHead}%
\makebox[0.25\textwidth][c]{\textbf{Task}} &
\makebox[0.30\textwidth][c]{\textbf{Screenshot}} &
\TabR{tabHead}%
\makebox[\dimexpr0.45\textwidth-12pt\relax][c]{\textbf{Task Instruction}}\\\TabHeadEnd
\midrule
\endfirsthead

\caption[]{Task profiles of HandsOn (continued).}\\
\toprule
\TabHead\TabStrut\TabL{tabHead}%
\makebox[0.25\textwidth][c]{\textbf{Task}} &
\makebox[0.30\textwidth][c]{\textbf{Screenshot}} &
\TabR{tabHead}%
\makebox[\dimexpr0.45\textwidth-12pt\relax][c]{\textbf{Task Instruction}}\\\TabHeadEnd
\midrule
\endhead

\multicolumn{3}{r@{}}{Continued on next page}\\
\endfoot

\bottomrule
\endlastfoot

\DexProfileRow{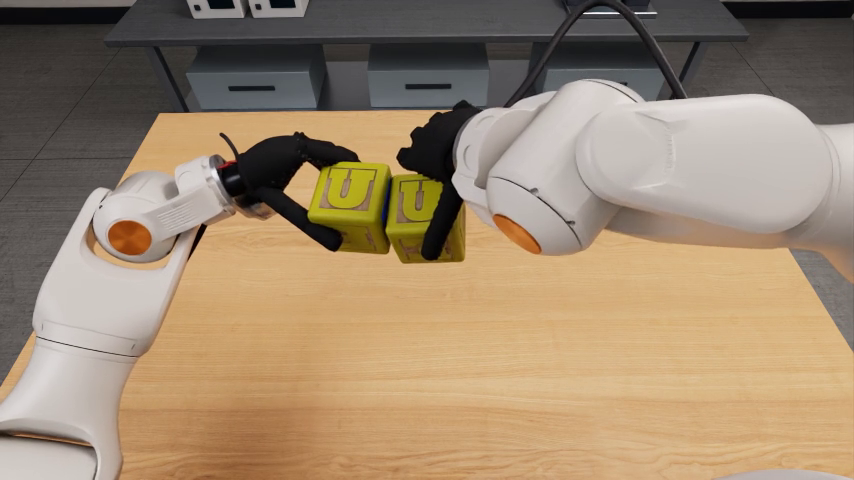}{Align Blocks in Midair}
  {Lift both assigned blocks with their assigned hands, then mate their marked faces in midair and hold the physical contact stable for 0.4 seconds.}

\DexProfileRow{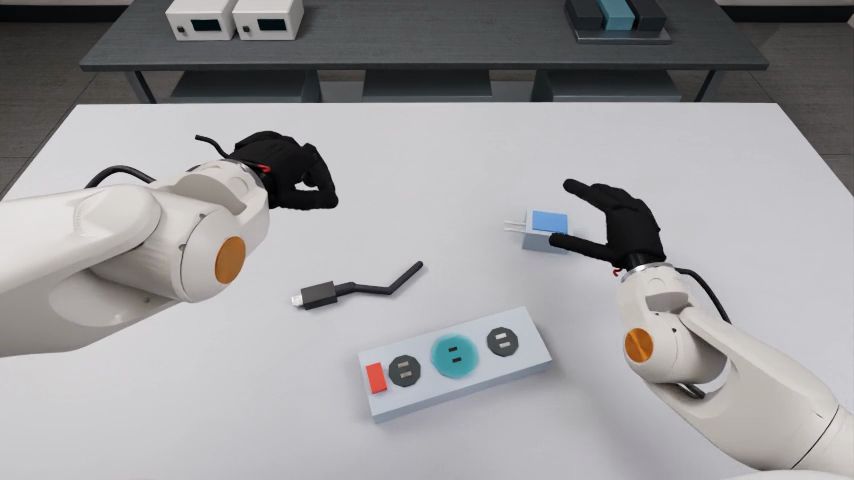}{Connect Charger and Cable}
  {Insert the charger plug into the power strip first, then insert the charging cable connector into the charger head.}

\DexProfileRow{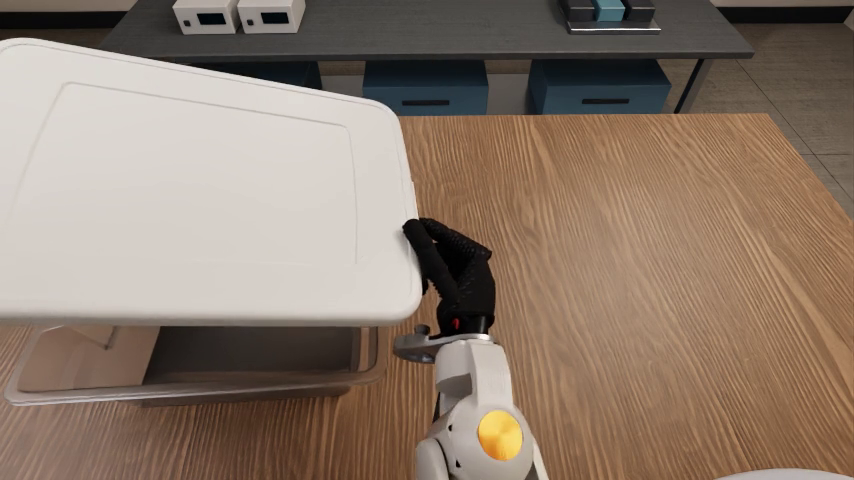}{Cover Box}
  {Align the box lid and place it onto the clear box.}

\DexProfileRow{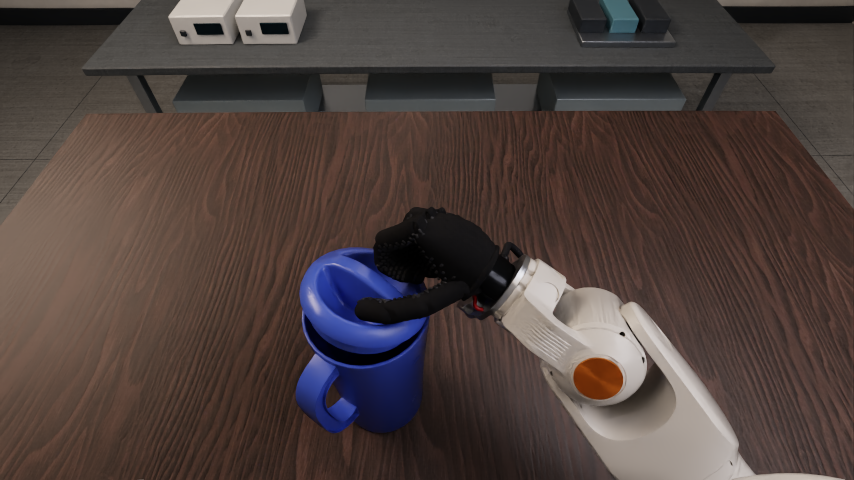}{Cover Pitcher}
  {Align the pitcher lid and place it onto the pitcher body.}

\DexProfileRow{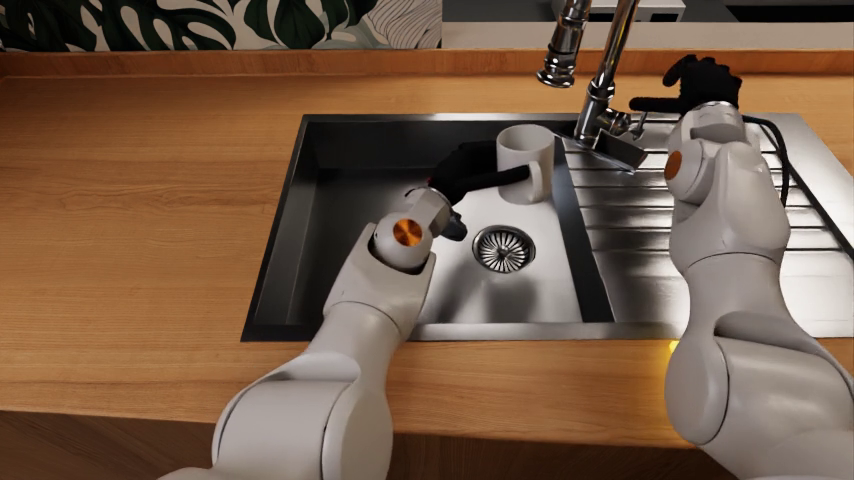}{Fill Cup}
  {Hold the mug upright under the faucet with the left hand. Open the right lever outward past 70 percent, fill for half a second, place the mug on the marked pad, and close the faucet.}

\DexProfileRow{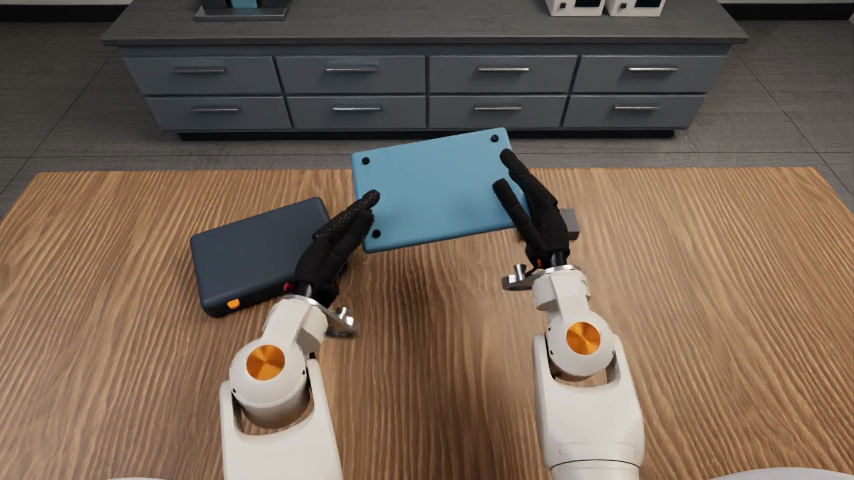}{Fit Device Cover}
  {Use both hands on opposite sides of the device cover to carry it together over the device body, then align it with its matching interface and seat it.}

\DexProfileRow{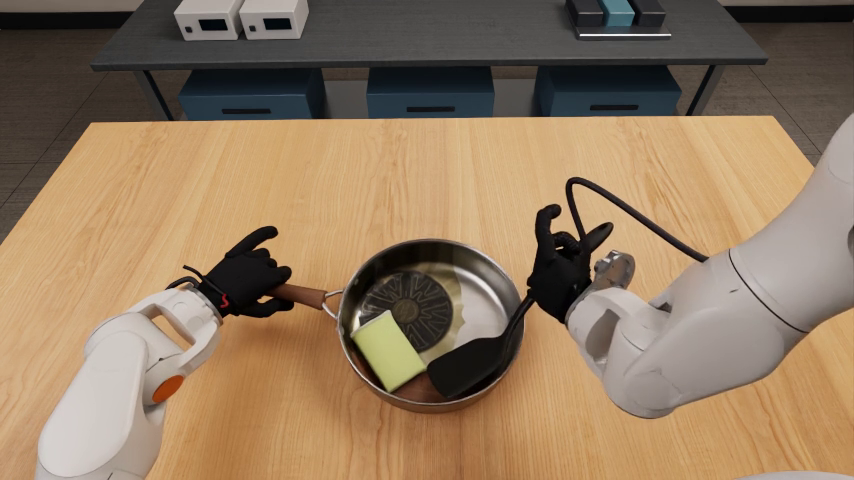}{Flip Sponge}
  {Hold the skillet level with the left hand. Use the right-hand spatula to lift and flip the sponge once, leaving its opposite face stable inside the skillet. Do not touch the sponge directly.}

\DexProfileRow{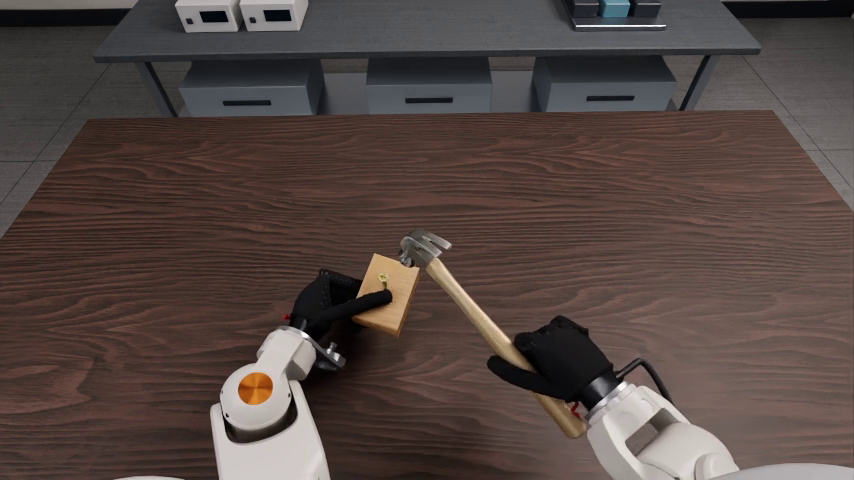}{Hammer Nail\textsuperscript{J}}
  {Hold the wood block with the left hand, keeping its nail inserted. Pick up the hammer with the right hand and strike the nail so it sinks into the block.}

\DexProfileRow{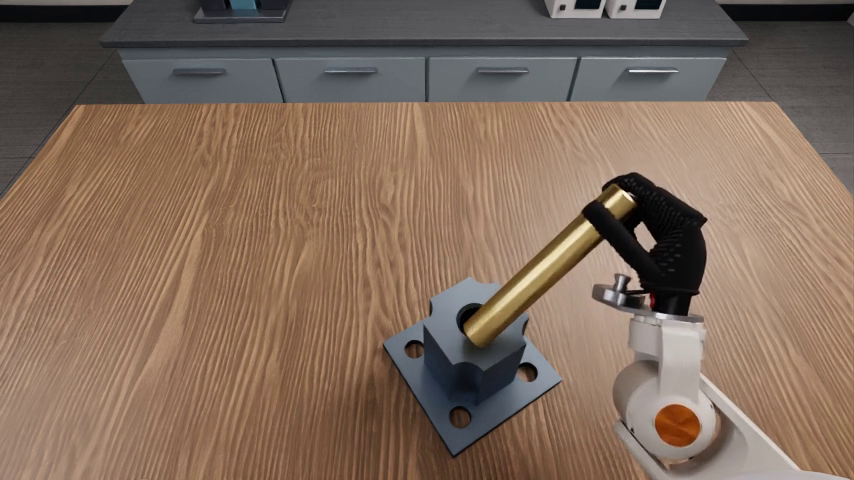}{Insert Peg\textsuperscript{JA}}
  {Insert the round peg into the socket tray hole.}

\DexProfileRow{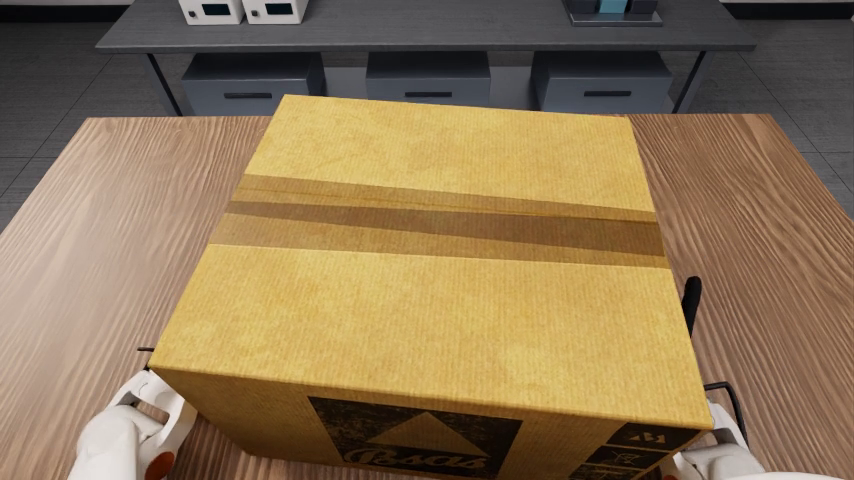}{Lift Carton\textsuperscript{V}}
  {Support opposite lower carton sides with both hands, lift the entire carton clear of the table, and keep it continuously airborne for 3 seconds.}

\DexProfileRow{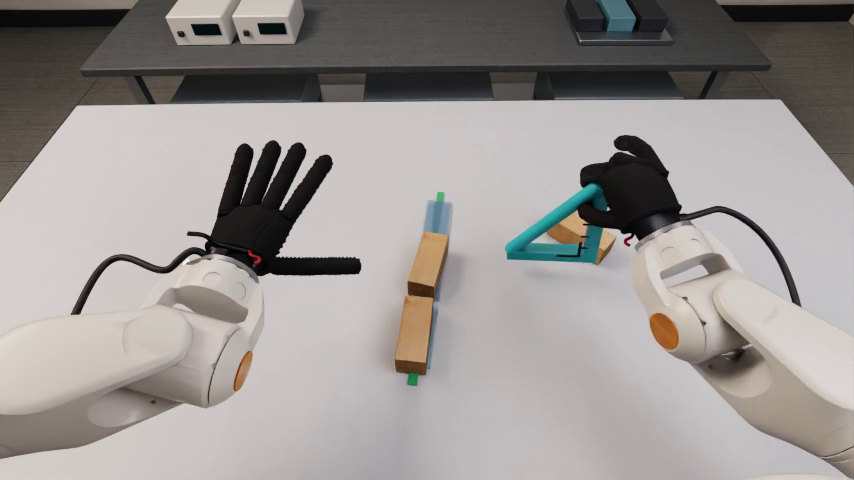}{Line Up Blocks}
  {Use only the set square tool to push three scattered wood blocks onto the baseline and align them side by side.}

\DexProfileRow{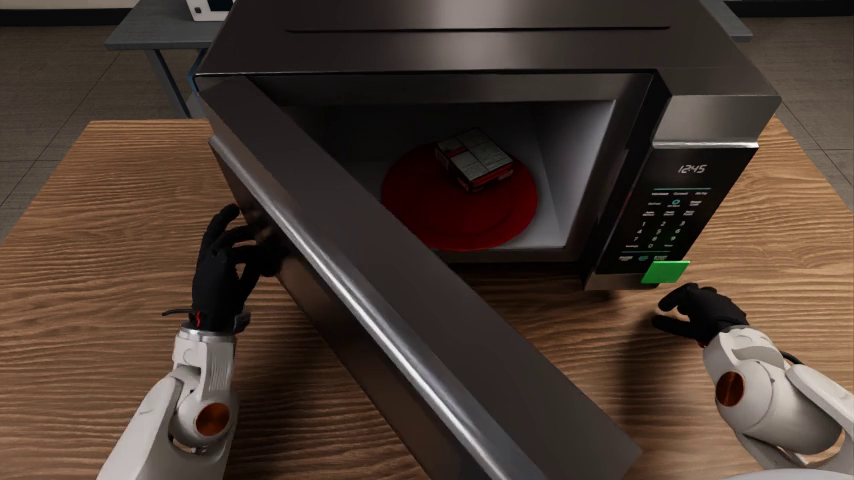}{Load and Start Microwave\textsuperscript{J}}
  {Open the microwave door, carry the plate with its food box into the microwave, close the door, then press the green start button.}

\DexProfileRow{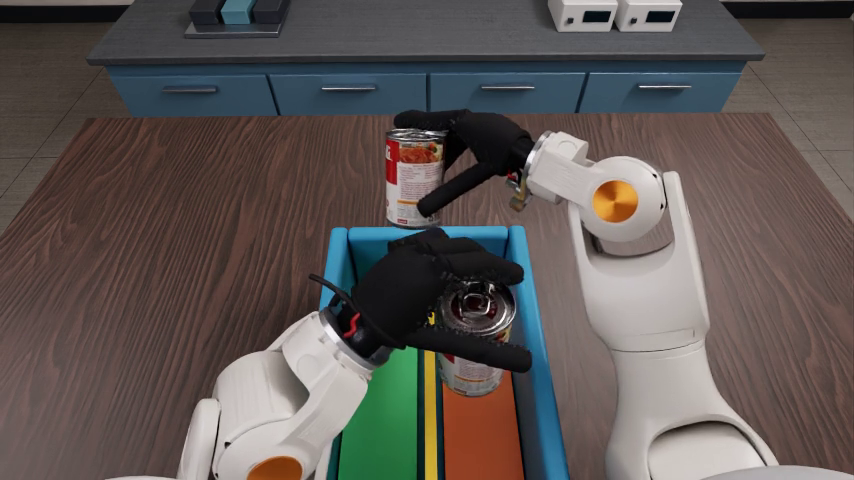}{Load Cans\textsuperscript{VA}}
  {Pick up both soup cans with their assigned hands and cross-load them into the opposite bin partitions. Release them within the synchronization window and leave them stable.}

\DexProfileRow{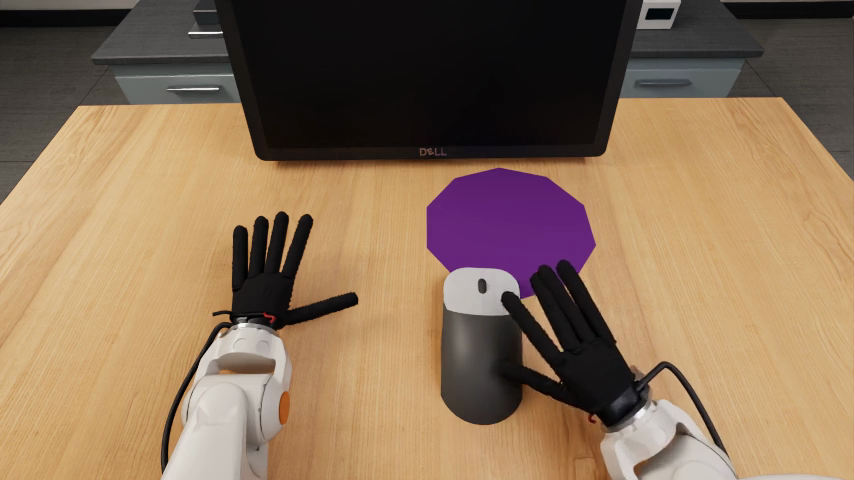}{Move and Click Mouse\textsuperscript{J}}
  {Move the mouse onto the mouse pad without pressing either button, then click the left mouse button.}

\DexProfileRow{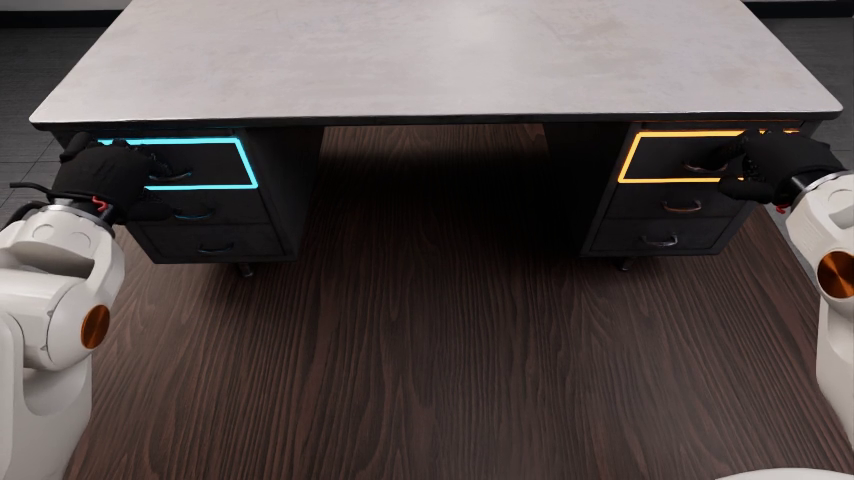}{Open Drawers}
  {Pull each \textbf{highlighted drawer} with its assigned hand past 60 percent travel, crossing both thresholds within 1.2 seconds.}

\DexProfileRow{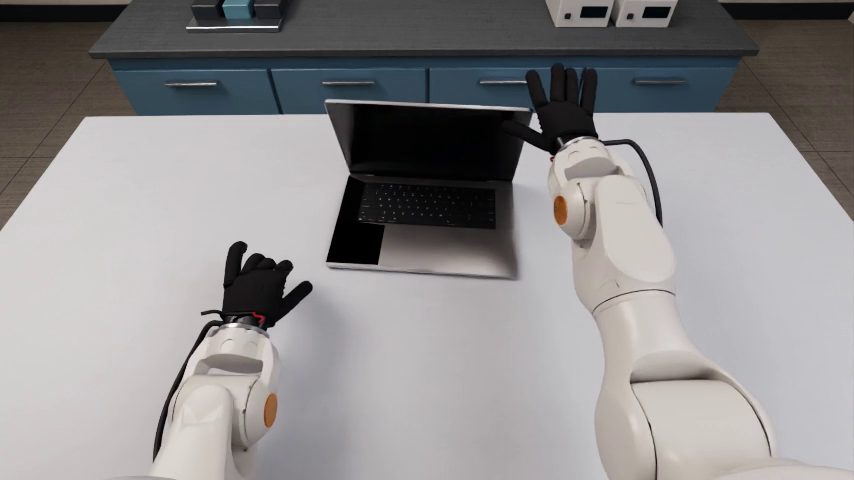}{Open Laptop\textsuperscript{V}}
  {Lift the laptop display from its front edge with the right hand and open it to 60 degrees.}

\DexProfileRow{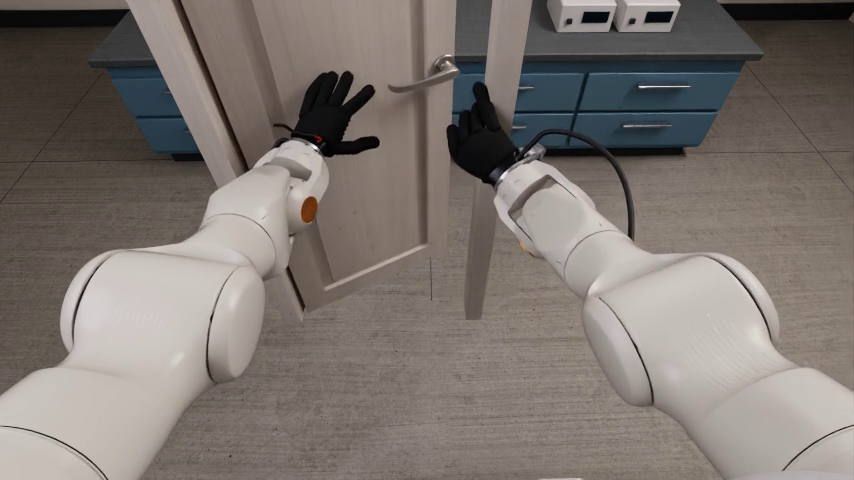}{Open Latched Door}
  {Depress the near lever past 45 degrees with the left hand for five frames. Push the unlatched door to 45 degrees with the right hand, avoiding the table, frame, and far handle.}

\DexProfileRow{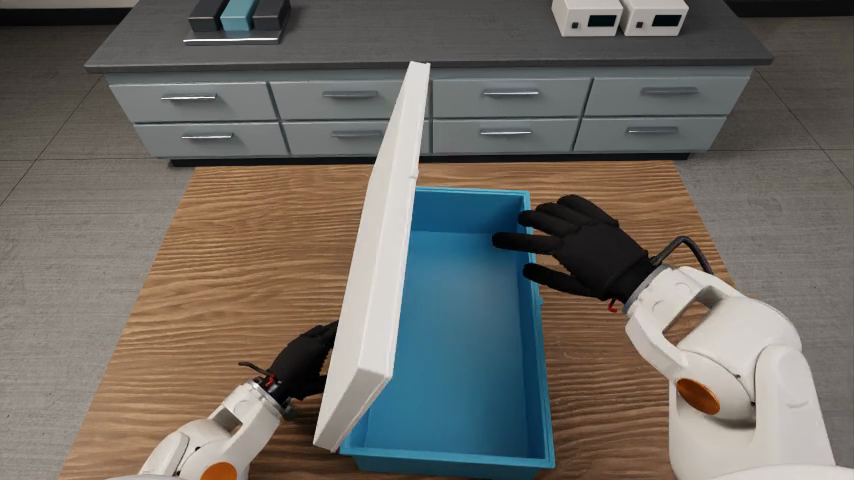}{Open Micropipette Box}
  {Open the hinged lid of the micropipette box until it is fully open.}

\DexProfileRow{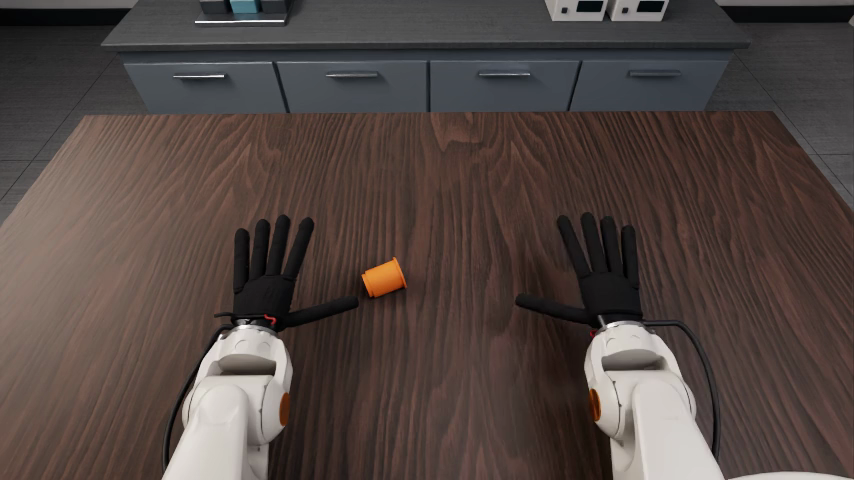}{Orient Cup Upright}
  {Pick up the cup and place it upright on the table.}

\DexProfileRow{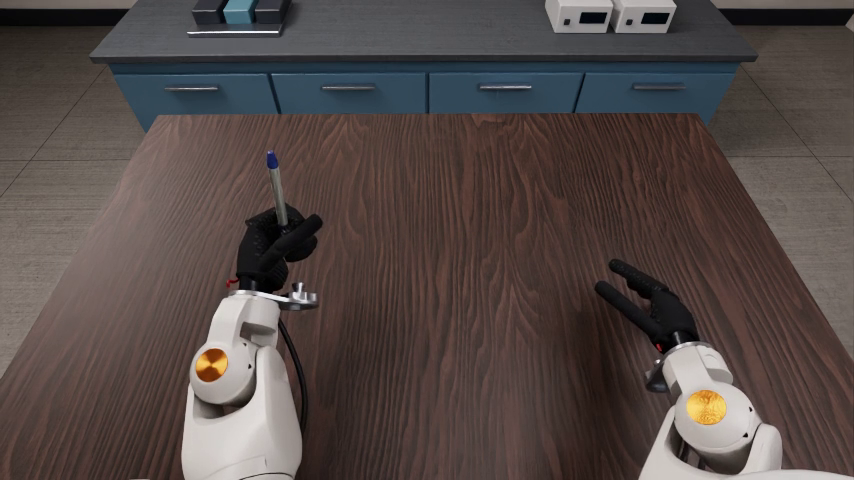}{Pass and Orient Pen}
  {Pass the pen from right-hand-only contact through both-hand contact to left-hand-only contact without dropping it. Hold it tip-up within 2 degrees of vertical.}

\DexProfileRow{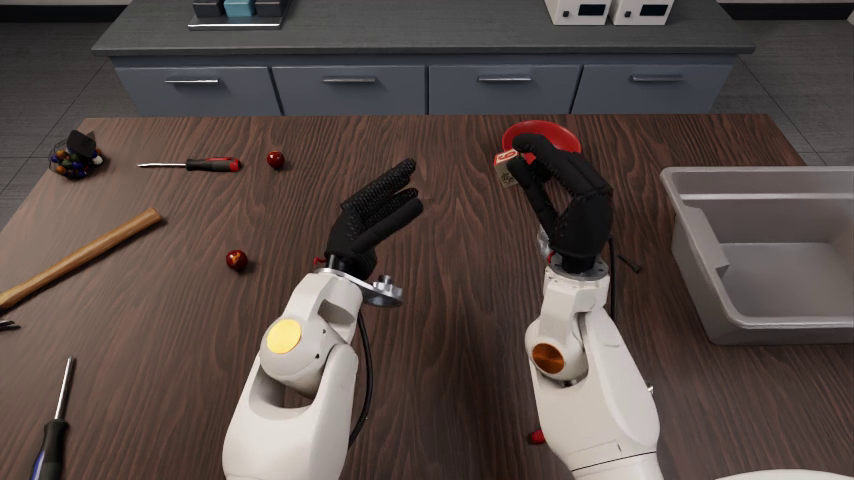}{Pass Block}
  {Pick up the left block with the left hand, transfer it to the right hand in the middle, then place it stably inside the right box.}

\DexProfileRow{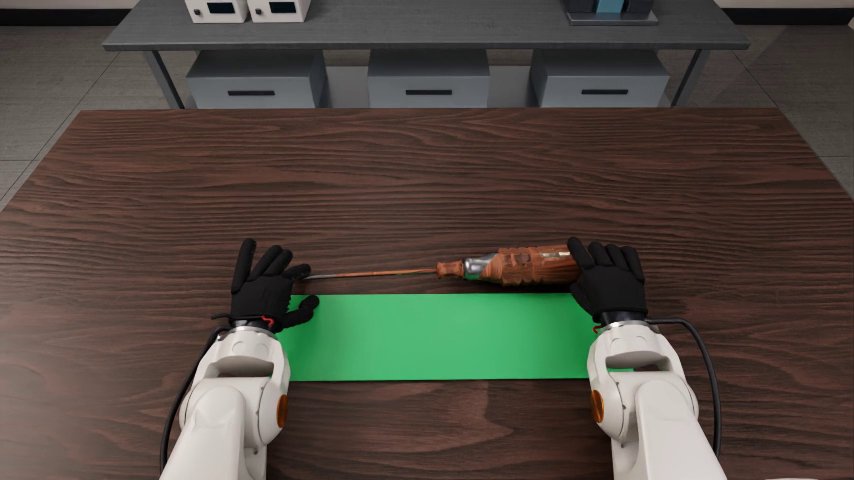}{Pass Screwdriver}
  {Touch the screwdriver with the right hand first, then touch it with the left hand while it is airborne, and let it land in the horizontal green target zone.}

\DexProfileRow{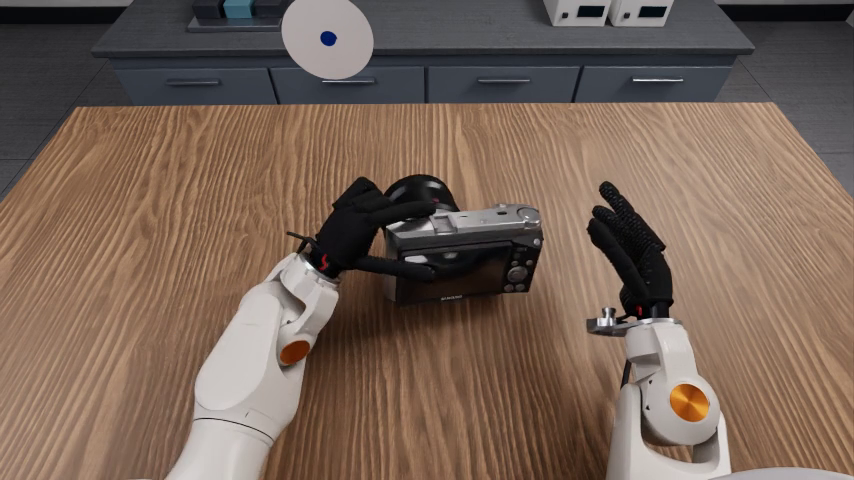}{Photograph Logo\textsuperscript{J}}
  {Pick up the camera, aim it at the logo, and press the shutter button.}

\DexProfileRow{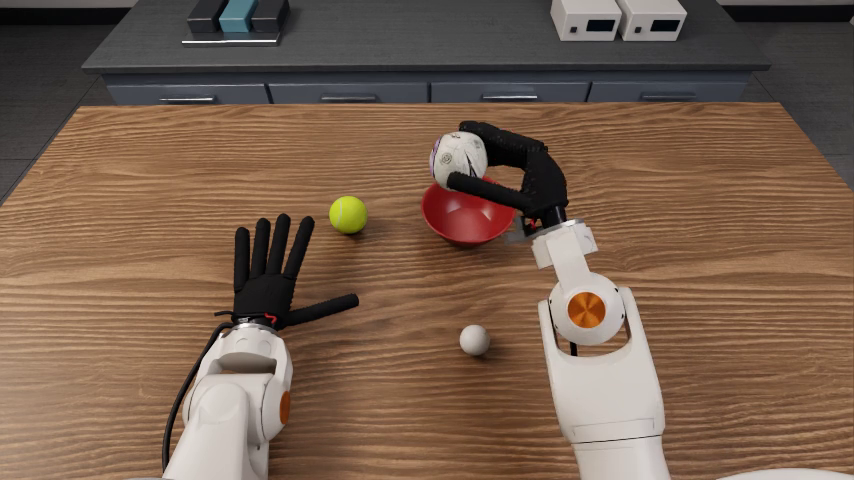}{Place Ball in Bowl}
  {Pick the \textbf{tennis ball} and place it into the bowl.}

\DexProfileRow{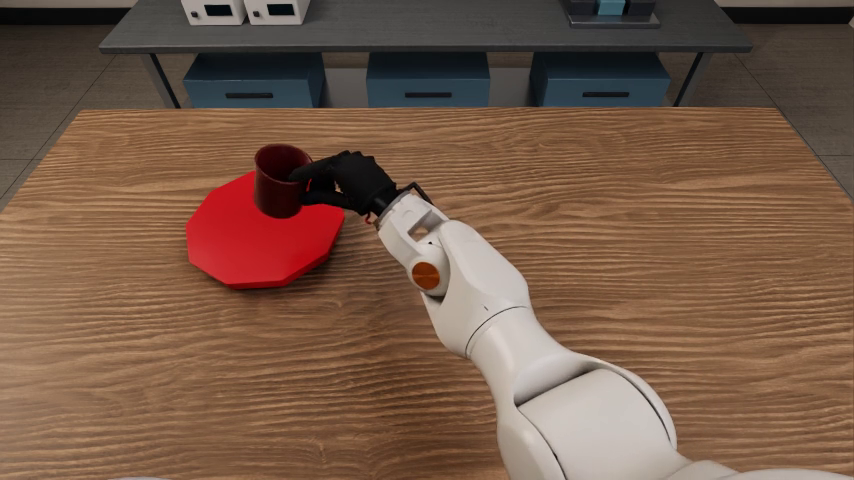}{Place Cup on Plate}
  {Pick up the mug and place it upright in the center of the plate.}

\DexProfileRow{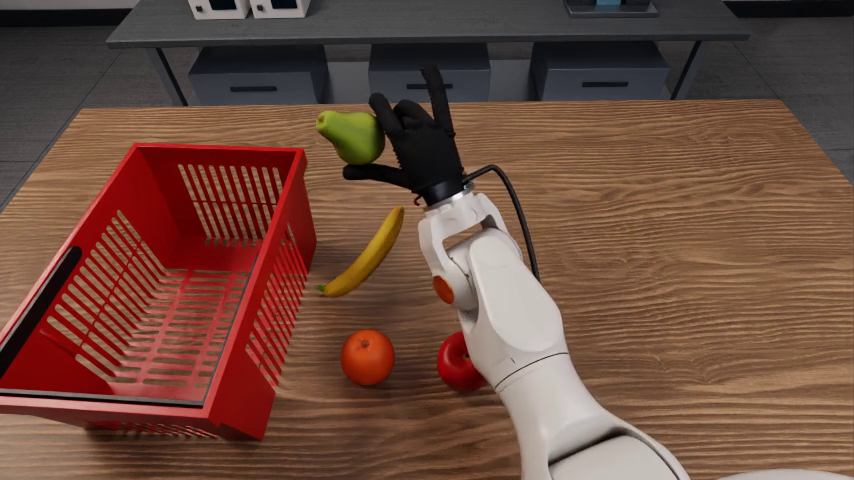}{Place Fruit in Bucket}
  {Pick the \textbf{orange} and place it into the bucket.}

\DexProfileRow{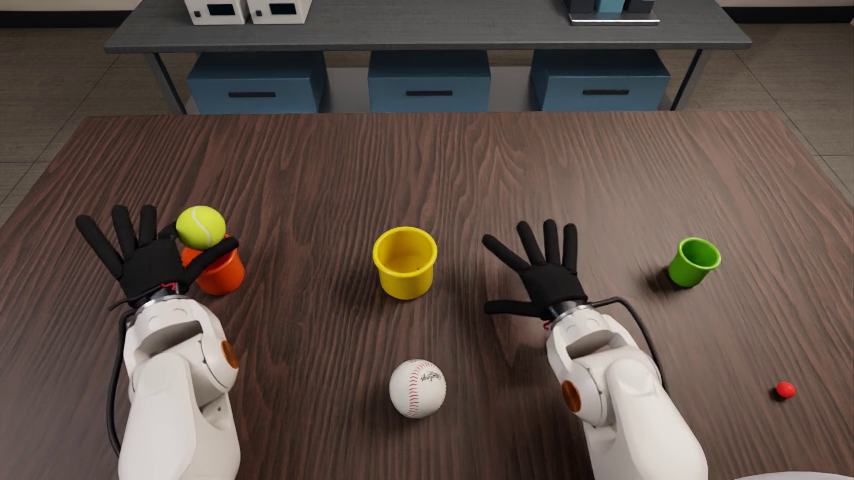}{Place Largest Fitting Ball}
  {Choose the largest ball that fits the \textbf{large yellow J cup}. Place it in that cup, release both hands, and leave ball and cup stable.}

\DexProfileRow{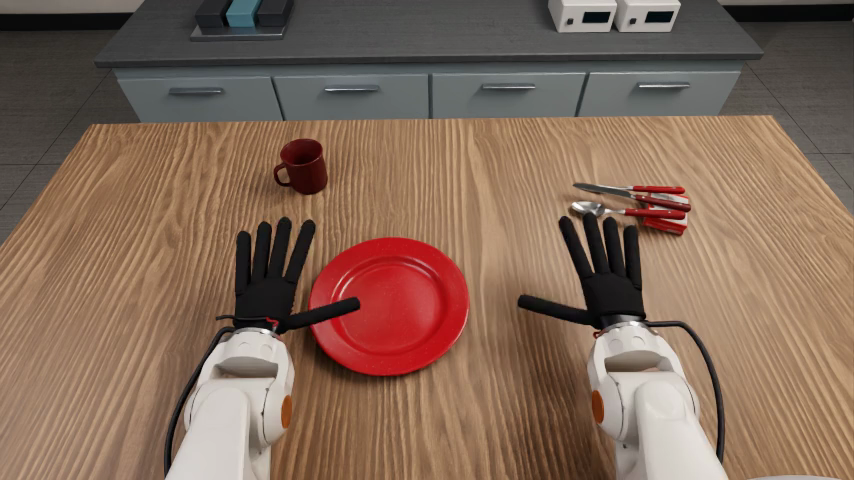}{Place Tableware}
  {Pick up the \textbf{mug} and place it \textbf{above and to the right of the plate}, \textbf{upright with its opening facing up}. Move only the \textbf{mug} and leave the plate and distractors in place.}

\DexProfileRow{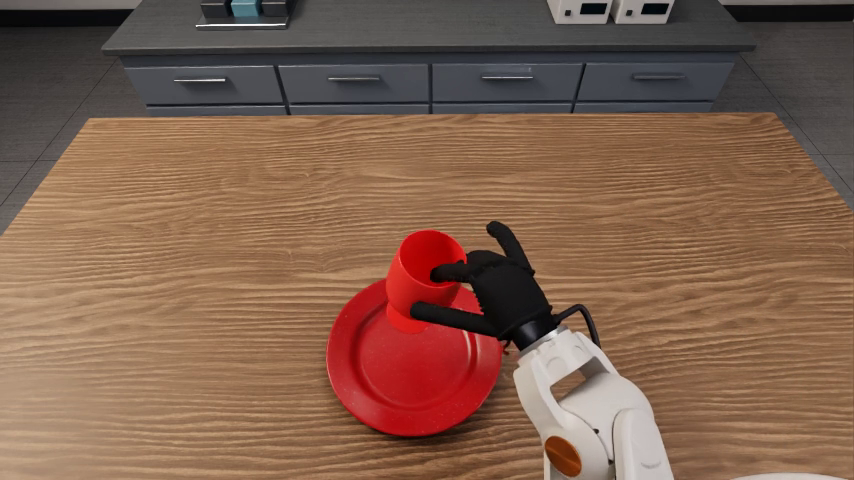}{Place Wine Glass}
  {Place the wine glass upright at the plate center, release it, and leave it stable.}

\DexProfileRow{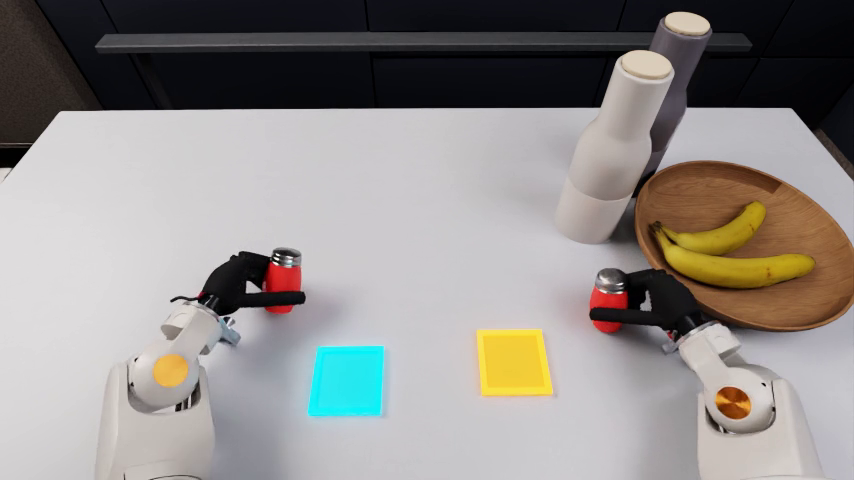}{Pour Seasonings}
  {Use the assigned hands to pour salt into the cyan target and pepper into the orange target, with overlapping pouring. Keep each jar's total pouring time below four seconds.}

\DexProfileRow{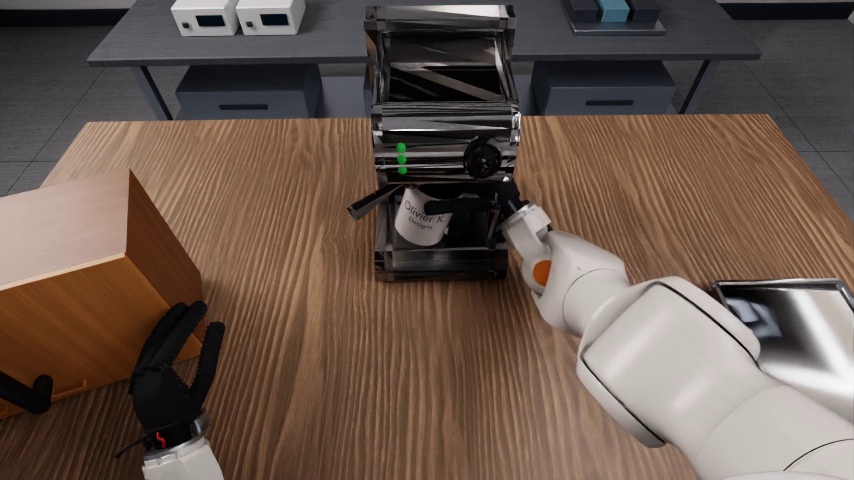}{Prepare Coffee}
  {Take the mug from the cabinet and close both doors. Place it under the coffee outlet, press the green button, and brew for two seconds. Release the filled mug upright on the serving tray.}

\DexProfileRow{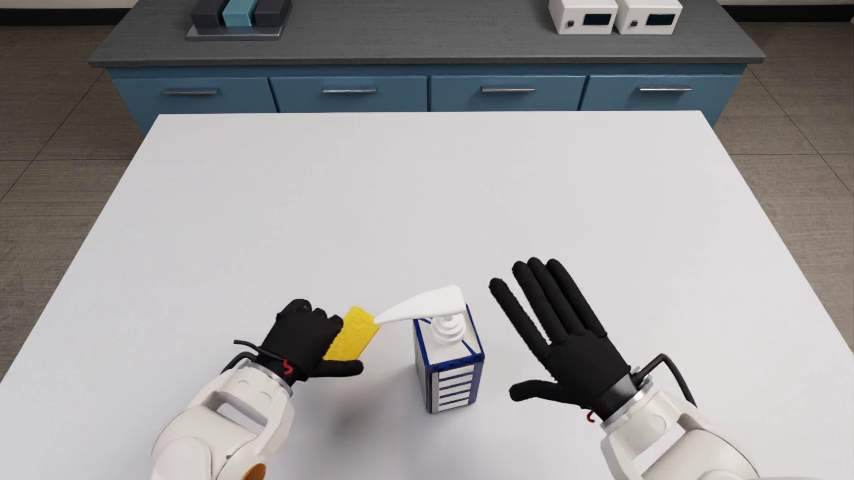}{Press Soap Pump}
  {Stabilize the sponge with the left hand, align the soap nozzle above it with the right hand, and physically depress the dispenser.}

\DexProfileRow{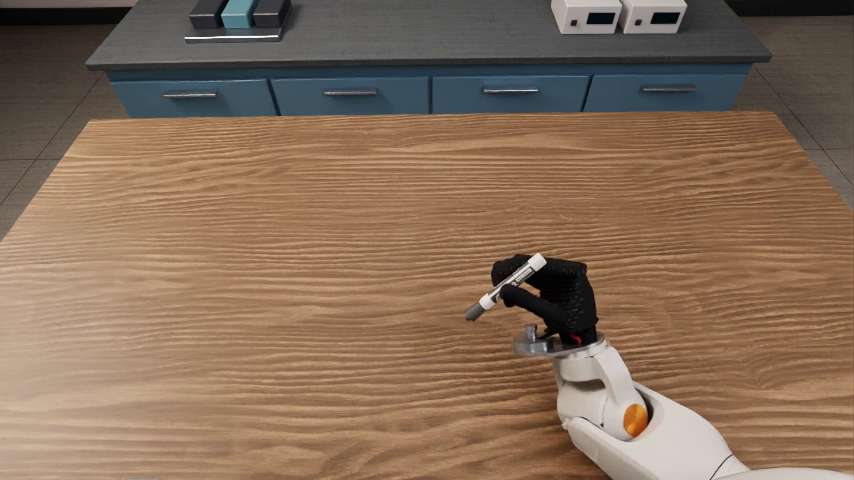}{Reposition Marker Grasp}
  {Lift the capped marker with the right hand. Move both thumb and index contact points at least 2.5 cm toward its writing tip while keeping it lifted and grasped.}

\DexProfileRow{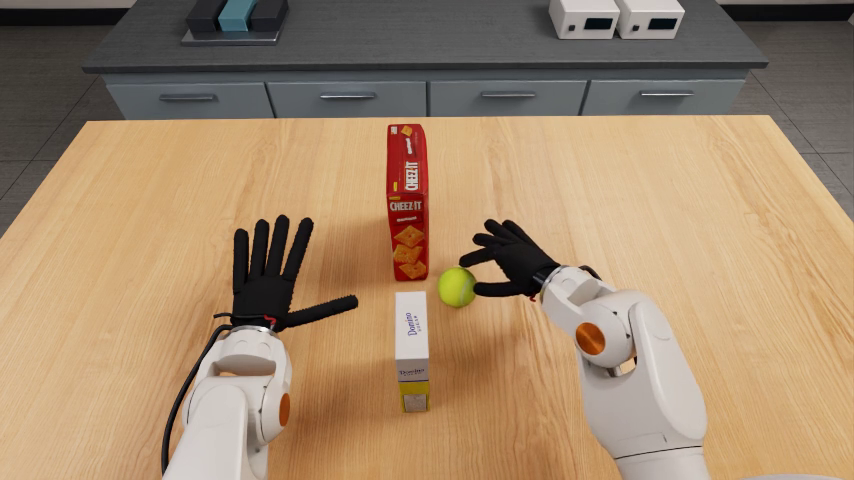}{Roll Ball Through Gate}
  {Roll or flick the tennis ball from its starting side through the opening between the cracker box and the sugar box. Keep it on the table without picking it up or touching either box.}

\DexProfileRow{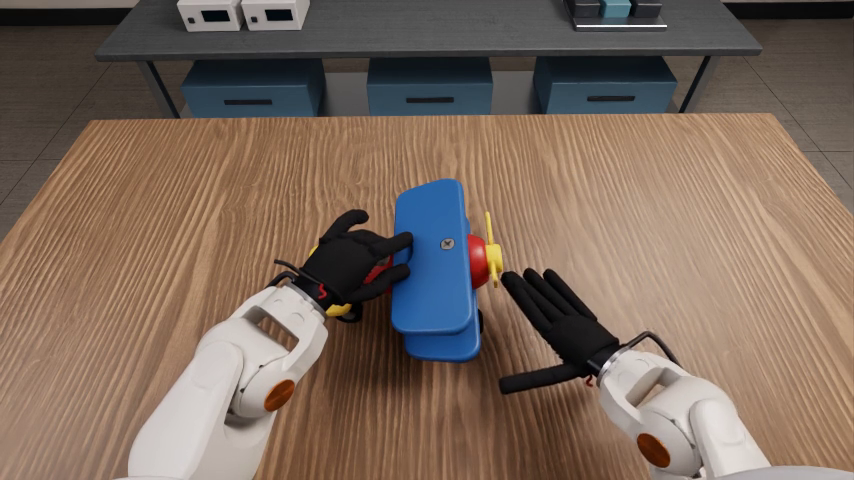}{Rotate Propeller}
  {Spin the airplane propeller through three full turns of accumulated travel, in either direction.}

\DexProfileRow{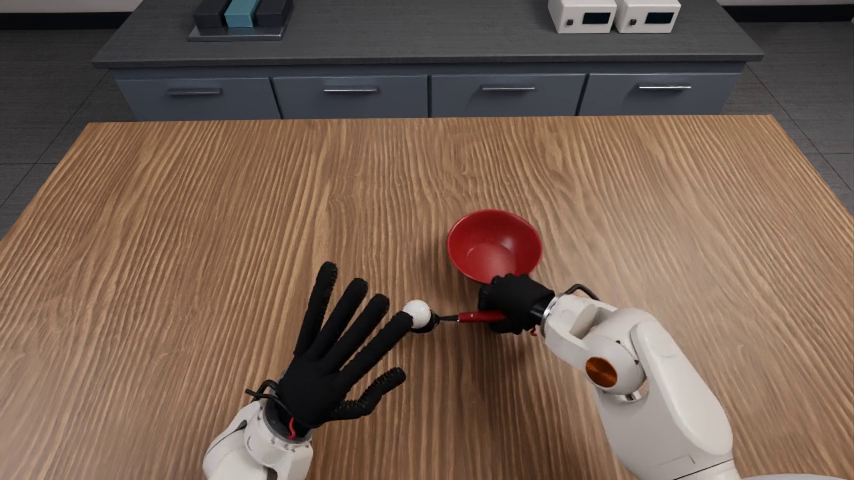}{Scoop Marbles}
  {Keep the spoon level or slightly nose-down, scoop the marble, carry it over the bowl, and pour it into the bowl.}

\DexProfileRow{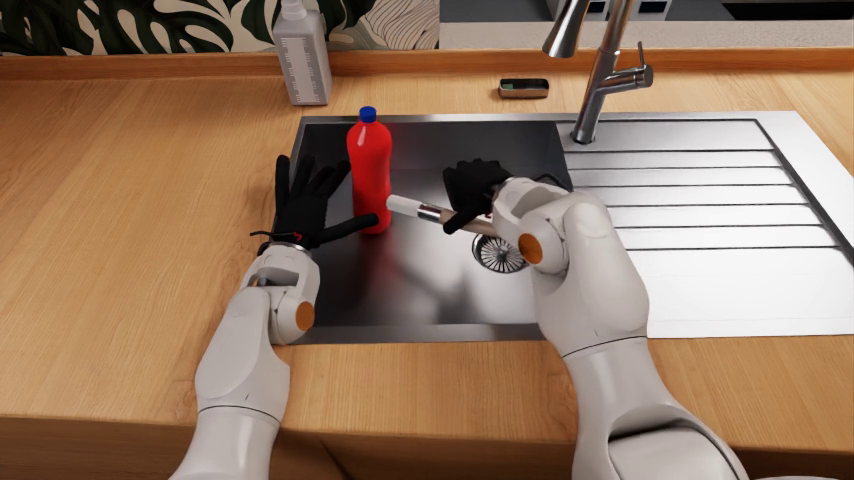}{Scrub Bottle}
  {Hold the water bottle with the left hand. Brush its lower, middle, then upper bands with the right hand. Release both objects, leaving the bottle upright and stable.}

\DexProfileRow{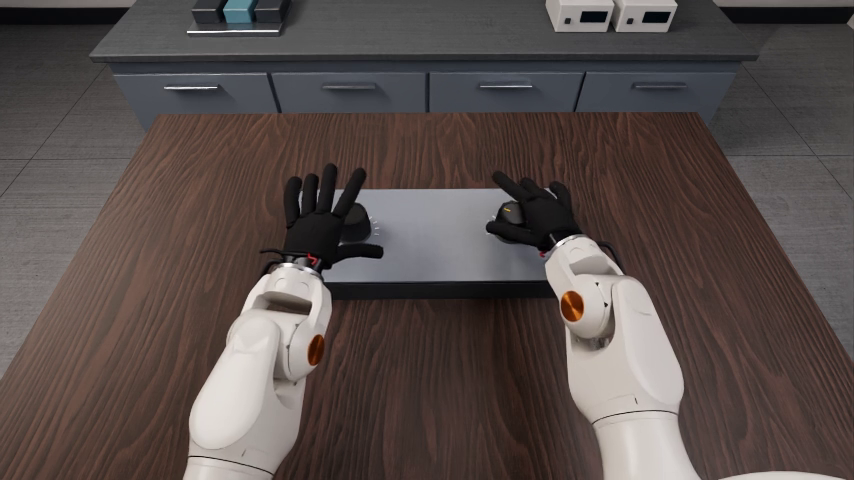}{Set Dual Knobs}
  {Turn both knobs together using their assigned hands, set the left knob to \textbf{60 degrees} counter-clockwise and the right knob to \textbf{60 degrees} clockwise, release both knobs, and keep the settings stable.}

\DexProfileRow{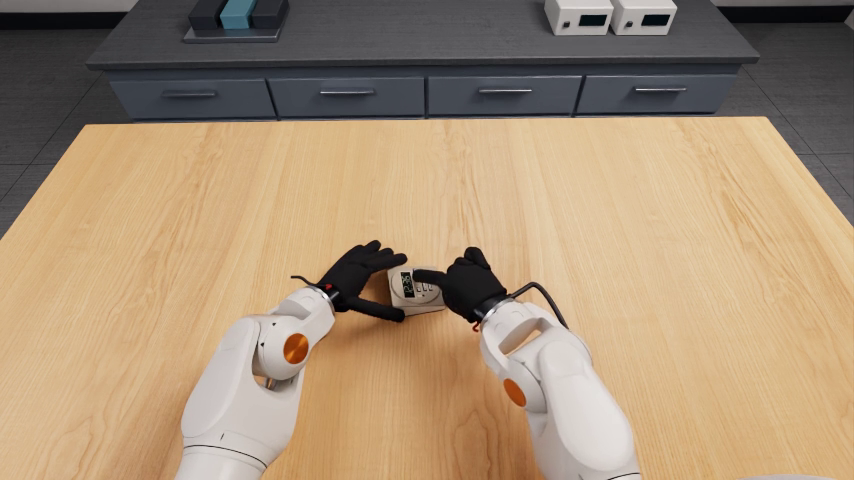}{Set Timer}
  {Hold the timer with the left hand. Enter \textbf{1 minute} and \textbf{1 second} with the right, press START last, then withdraw both hands for one countdown tick. Clear incorrect entries first.}

\DexProfileRow{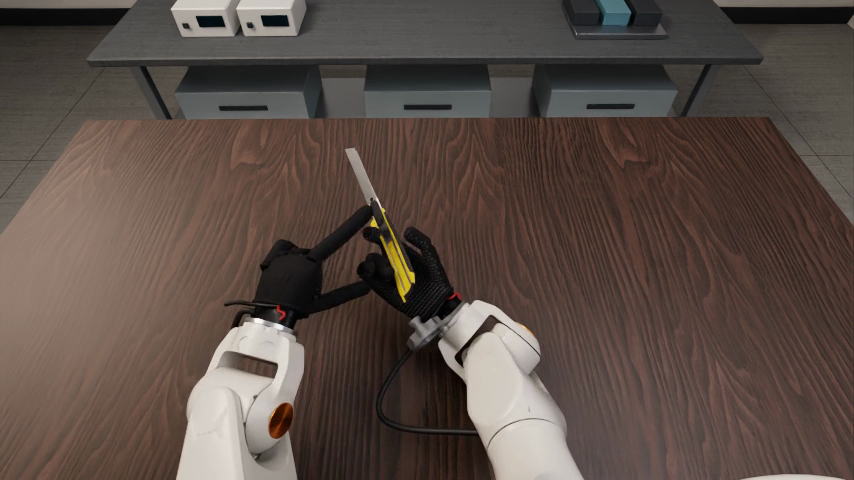}{Slide Knife Blade\textsuperscript{V}}
  {Pick up the utility knife with the right hand, then use the left fingertips on its slider to \textbf{extend} the blade fully.\newline
  (The action goal is random: \textbf{extend} may be replaced with \textbf{retract}.)}

\DexProfileRow{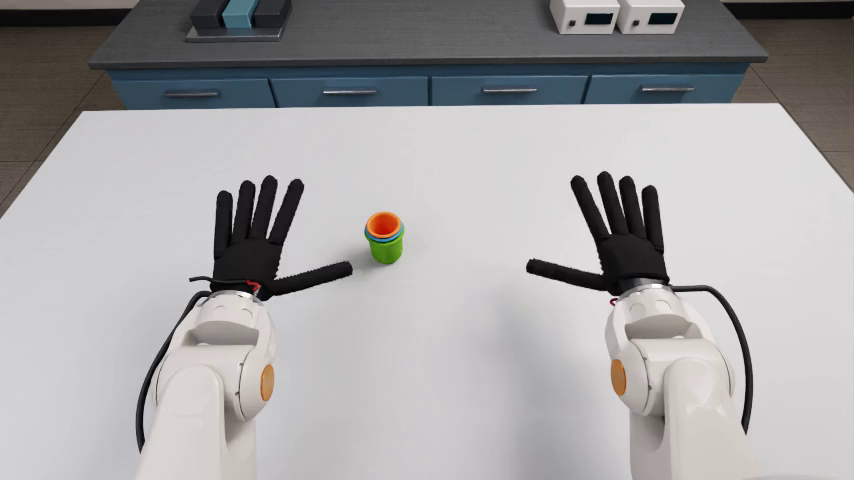}{Sort Nested Cups}
  {Separate the three nested YCB cups, then place them upright in a straight row from left to right in the main camera: large, medium, small. Release all cups.}

\DexProfileRow{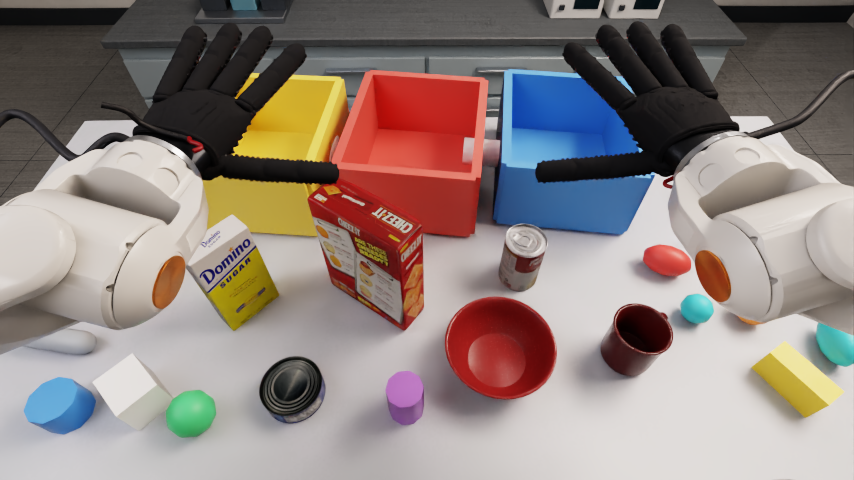}{Sort Objects}
  {Find six targets among the clutter and sort boxes, cans, and dishes into their marked baskets. Transfer the far-right target through both hands before placing it.}

\DexProfileRow{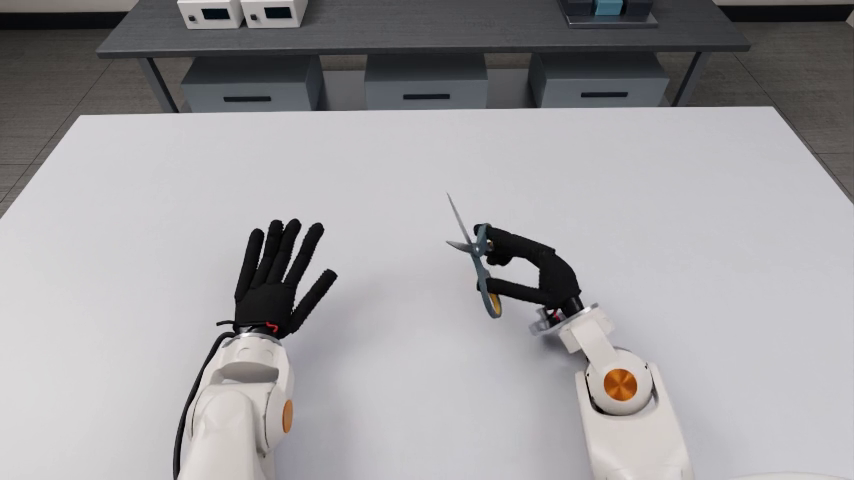}{Squeeze Scissors}
  {Lift the scissors clear of the table and make three complete cuts in the air. Try one hand first; the other hand may reposition or assist.}

\DexProfileRow{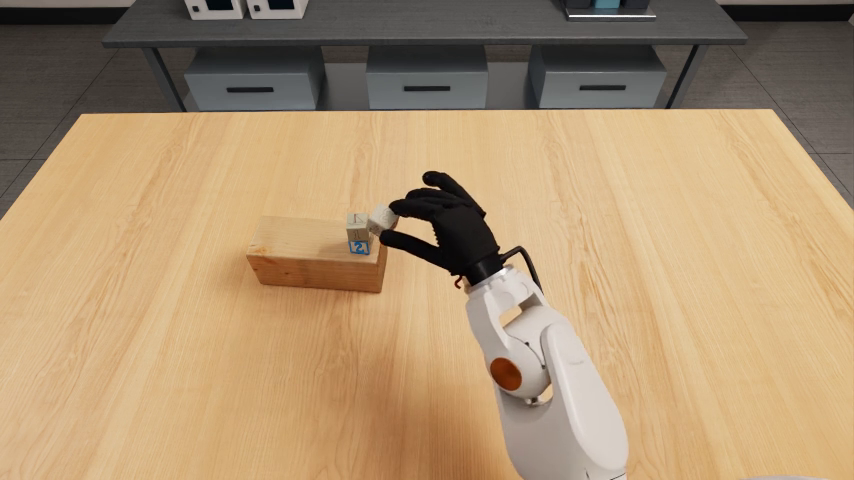}{Stack Blocks}
  {Arrange the four blocks into four distinct height layers with the large wood block on the bottom layer, then keep the hand from touching the blocks for 0.5 seconds.}

\DexProfileRow{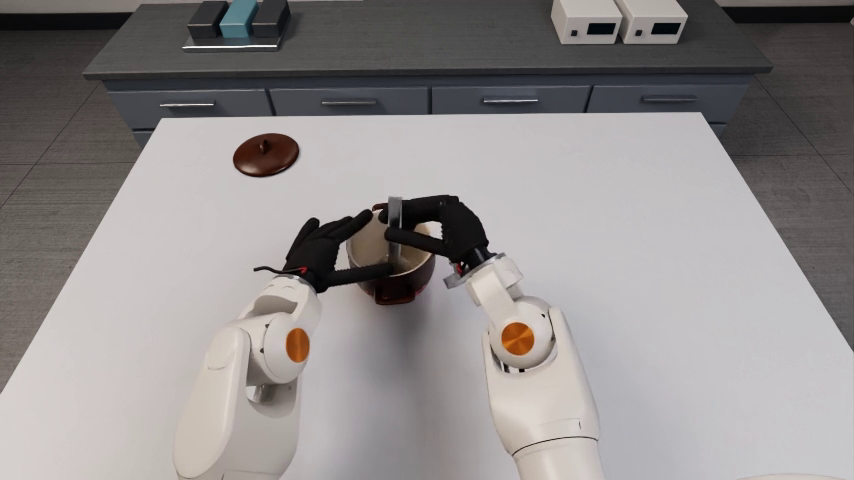}{Stir Pot}
  {Hold the open pot with the left hand and stir with the right-hand spoon through the six ordered checkpoints and over 650 degrees.}

\DexProfileRow{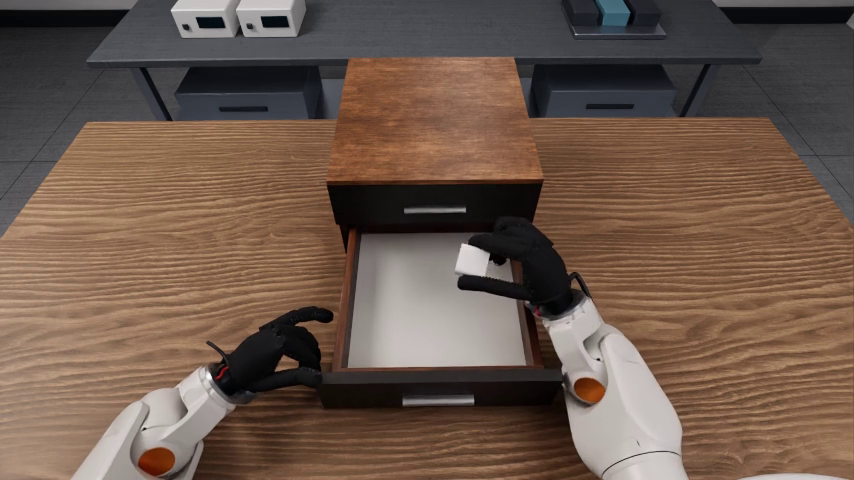}{Store Charger\textsuperscript{VA}}
  {Hold the drawer open with the left hand and place the charger fully inside with the right. Release the charger, then the drawer, and hold stable for half a second.}

\DexProfileRow{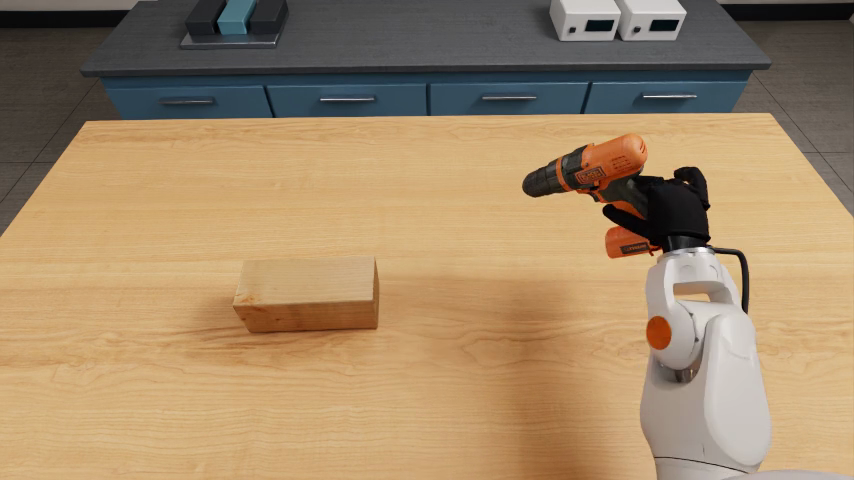}{Trigger Drill on Block}
  {Use the right hand to bring the drill into contact with the wood block, then press the drill's trigger and hold it for half a second while keeping the contact.}

\DexProfileRow{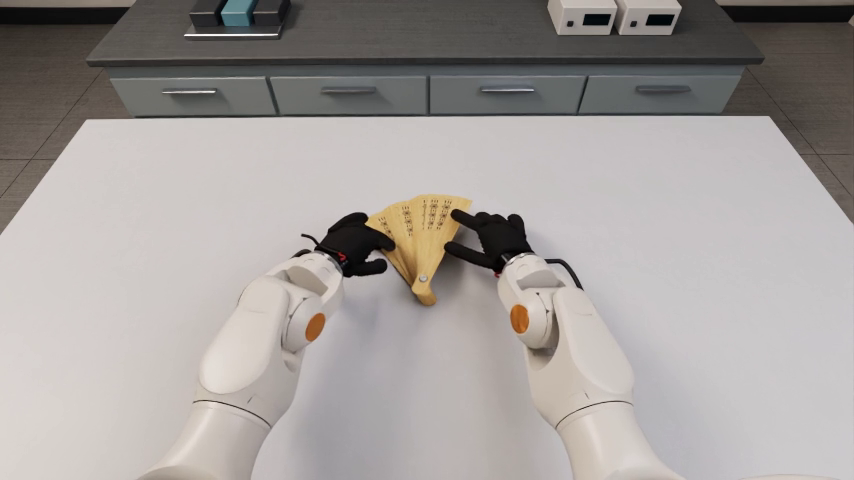}{Unfold Fan}
  {Use both hands to grasp the two outermost fan ribs and unfold the fan symmetrically to \textbf{60 degrees}, then hold it steadily on the table.}

\DexProfileRow{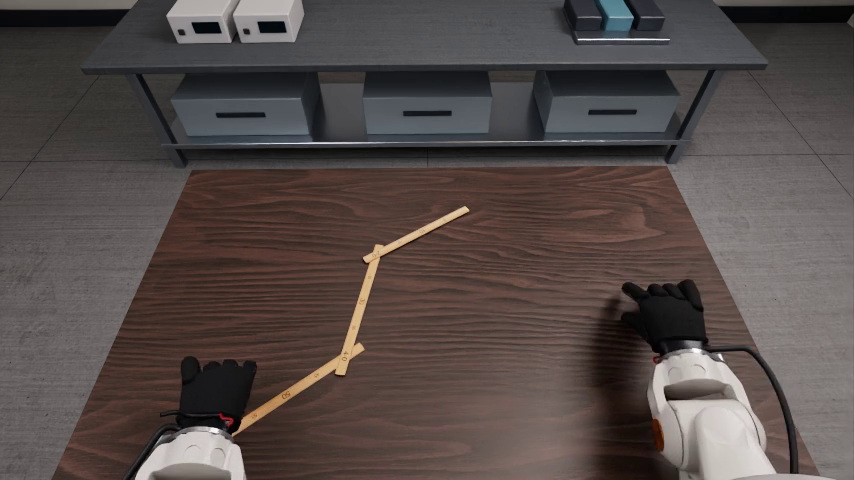}{Unfold Ruler}
  {Pinch the two outer ruler segments with their assigned hands and unfold them together until the ruler is straight.}

\DexProfileRow{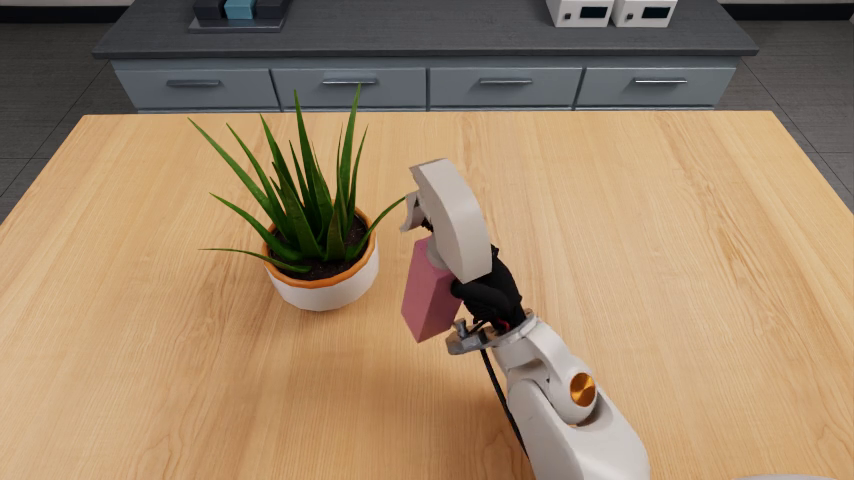}{Water Plant\textsuperscript{J}}
  {Grasp the spray bottle and aim it at the plant before pressing, then press the sprayer so that the water lands on the plant or its soil.}


\DexProfileRow{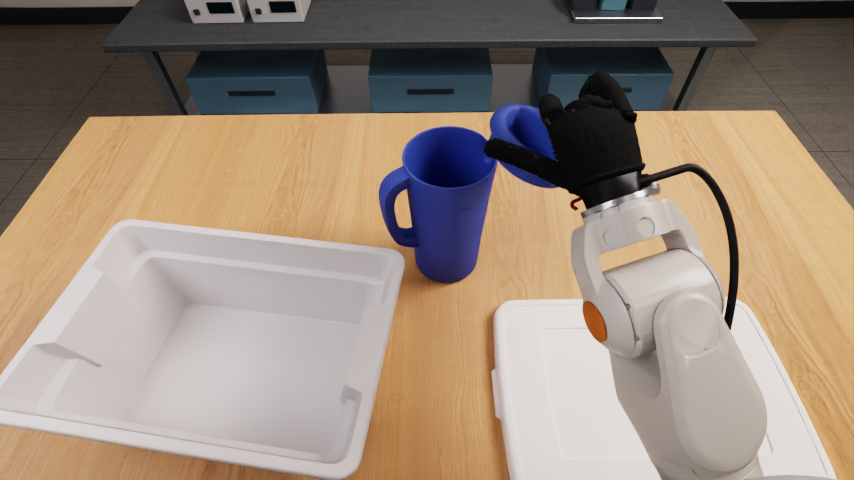}{Cover Pitcher and Box}
  {First align and seat the pitcher lid, then release it and let it settle. Next align and seat the clear-box lid, release it and let it settle. Both lids must remain correctly seated on their own containers.}

\DexProfileRow{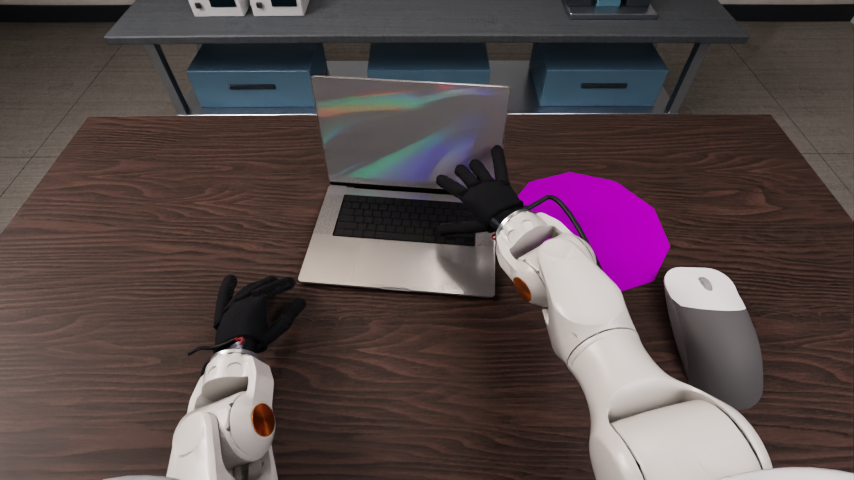}{Open Laptop and Click Mouse}
  {Open the initially closed laptop to the \textbf{60 degrees} with either hand and let it settle; you may release the screen. Move the mouse onto its pad without pressing either button, then click the left button once. Keep the laptop at the \textbf{target angle} and the mouse on the pad while the scene settles.}

\DexProfileRow{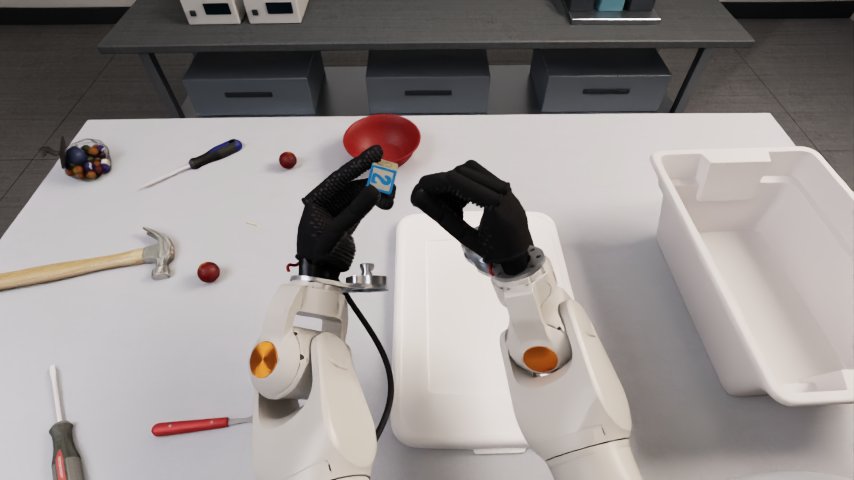}{Pass Block and Close Box}
  {Pick up the left block with the left hand, hand it to the right hand in the middle, and release it stably inside the clear box. Then use the right hand to pick up the lid, align and close the box, and release.}

\DexProfileRow{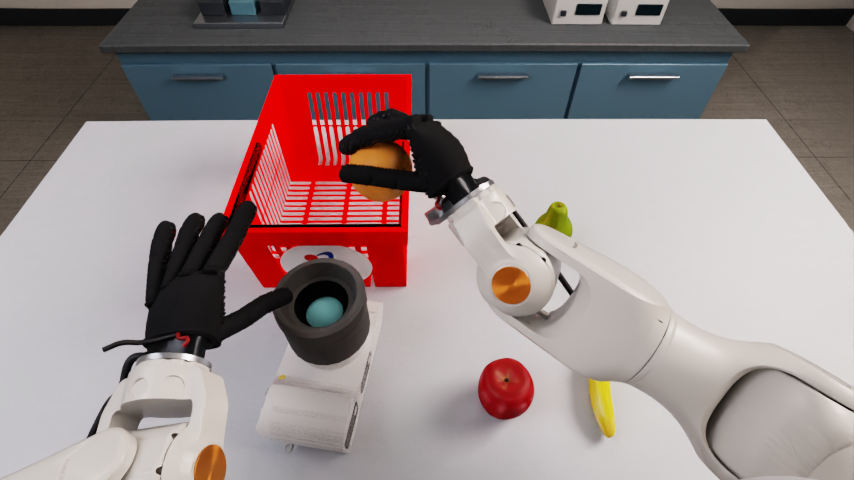}{Place Fruit and Photograph Logo}
  {Pick the \textbf{orange}, place it in the bucket and release it. Then pick up the camera, aim at the logo on the bucket and press the shutter.}

\DexProfileRow{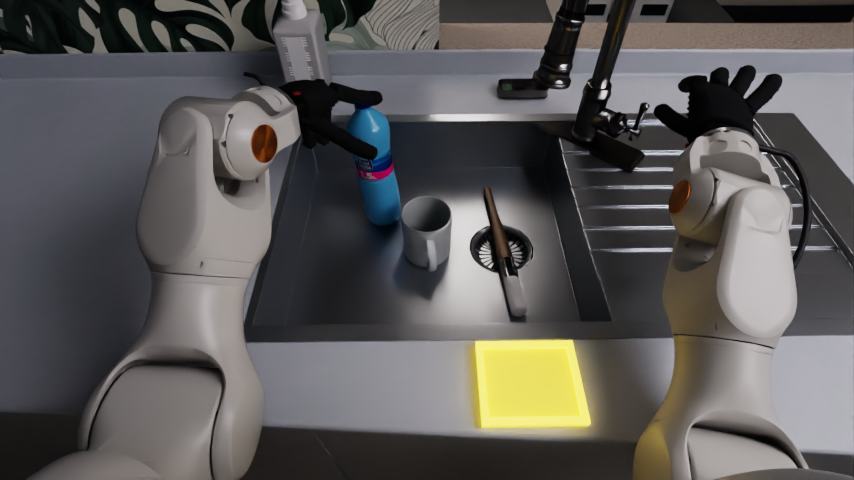}{Scrub Bottle Then Fill Cup}
  {Stabilize the bottle with the left hand and grasp the cleaning brush by its handle with the right hand. Scrub the bottle bands in order, either lower-middle-upper or upper-middle-lower, then release both objects and leave the bottle upright and stable. Next, hold the mug under the faucet with the left hand and open the faucet with the right hand. Keep the stream entering the upright mug for 0.5 seconds, then place the mug on the marked counter pad outside the sink and close the faucet. Leave the scrubbed bottle upright.}

\DexProfileRow{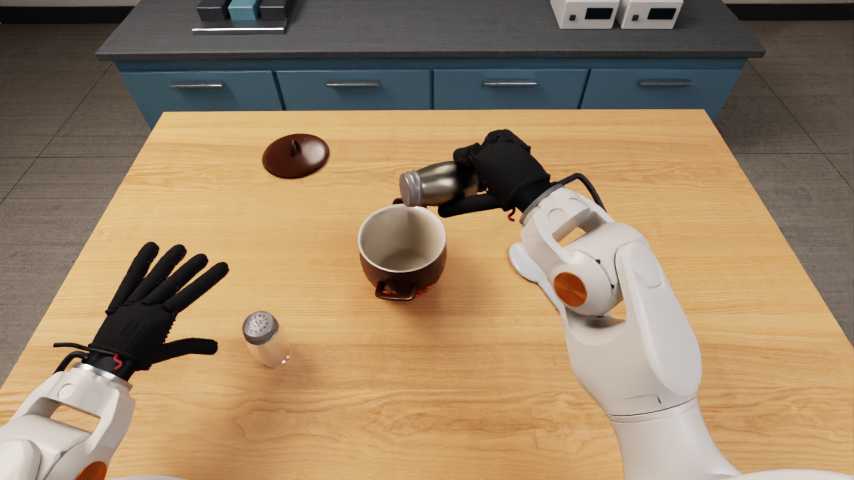}{Season and Stir}
  {Use the left hand for salt and the right hand for pepper. Pour both into the same open pot together, then put down and release the jars. Hold the pot with the left hand and stir with the right-hand spoon through all six angular sectors and more than 650 degrees, keeping the pot stable.}


\DexProfileRow{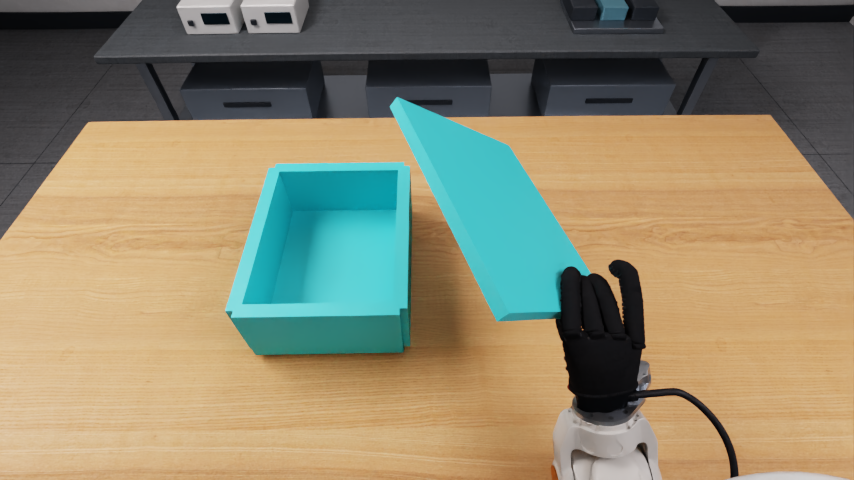}{Cover Lunchbox}
  {Align the lunchbox lid and place it onto the lunchbox.}

\DexProfileRow{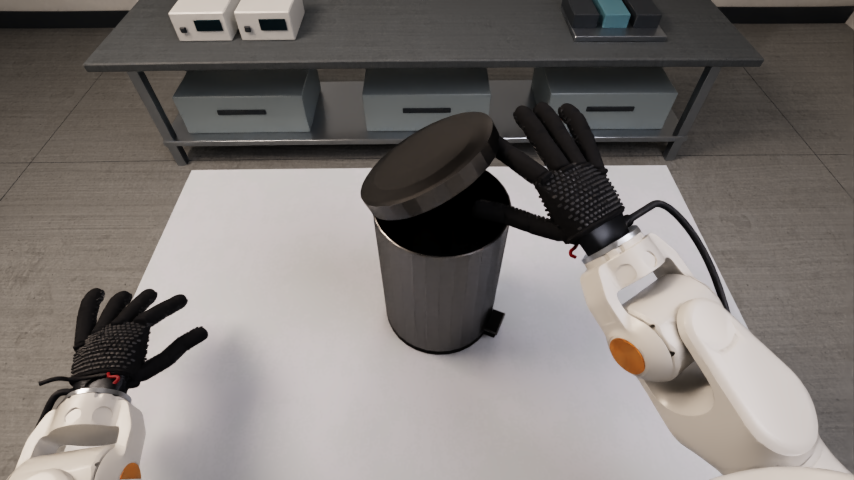}{Open Trash Bin}
  {The trash bin is fixed after reset at a randomized tabletop pose and yaw. Open its hinged lid from closed to at least 75 degrees. Success is immediate when the physical lid reaches that opening. Either hand may be used; no assigned contact, release, or stable hold is required.}

\DexProfileRow{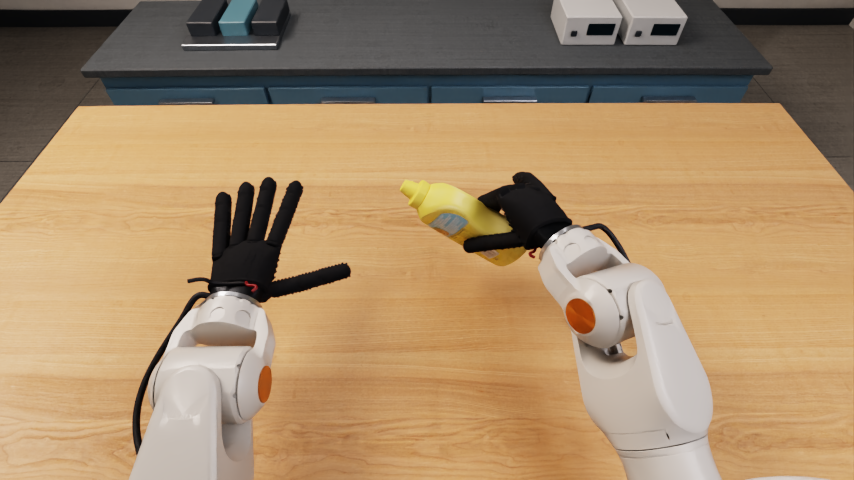}{Orient Bottle Upright}
  {Pick up the yellow mustard bottle and place it upright on the table with its bottom facing down. Release the bottle and keep it upright for 0.5 seconds.}

\DexProfileRow{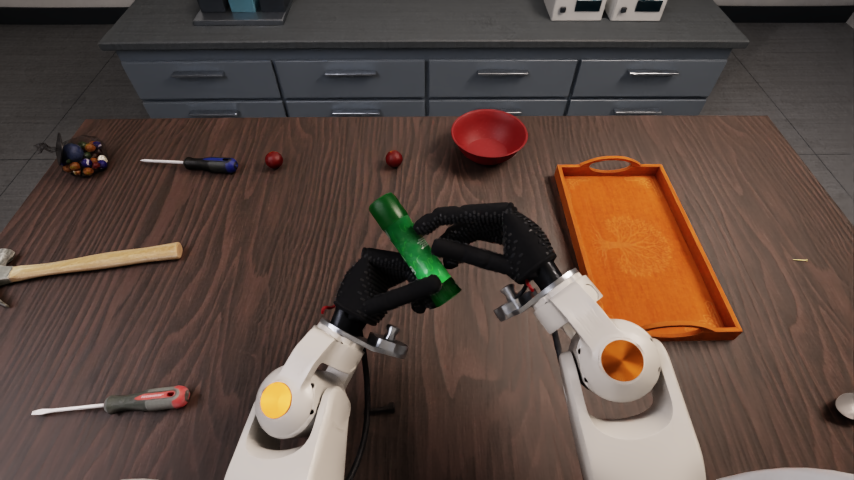}{Pass Glue Stick}
  {Pick up the glue stick with the left hand, transfer it to the right hand in the middle, then place it stably inside the right tray and release it.}

\DexProfileRow{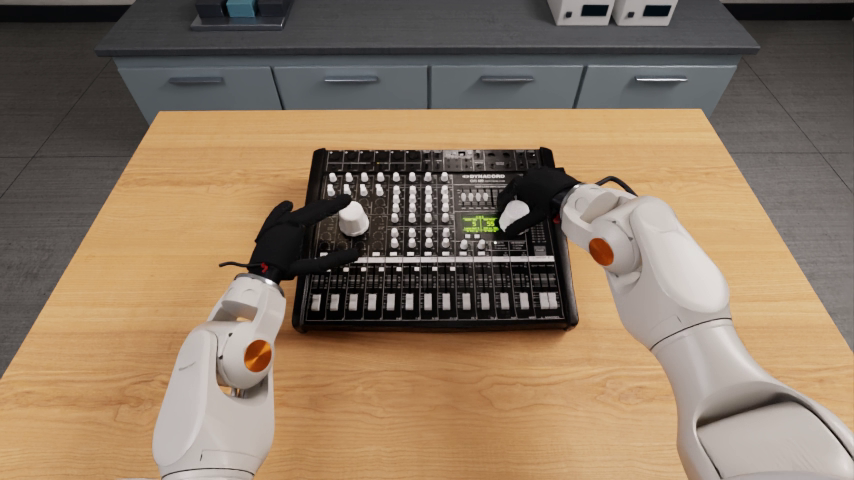}{Set Mixer Knobs}
  {Use your left hand to turn the left knob counterclockwise and your right hand to turn the right knob clockwise, in either order. Set the left knob to \textbf{60 degrees} and the right knob to \textbf{-60 degrees}, release both knobs, and keep the settings stable.}

\DexProfileRow{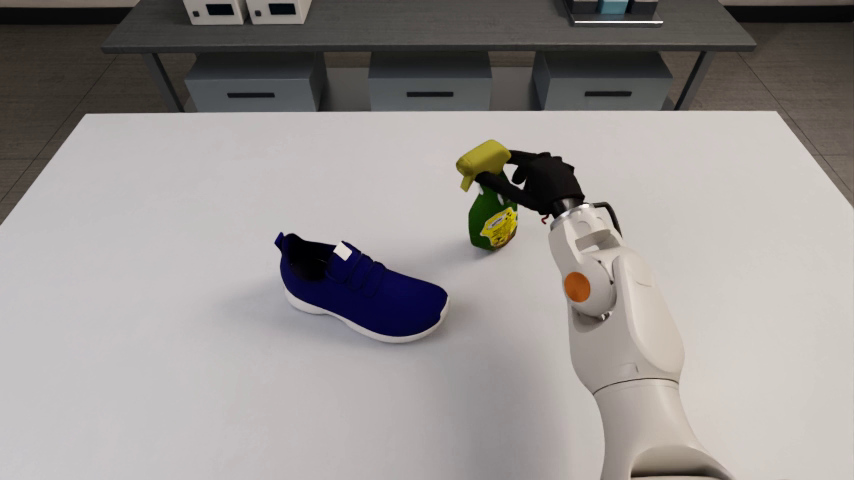}{Spray Shoe}
  {Grasp and aim the spray bottle at the shoe upper before squeezing the trigger. Keep spraying the shoe to complete the task; a spray that entirely misses the shoe upper fails the attempt.}

\DexProfileRow{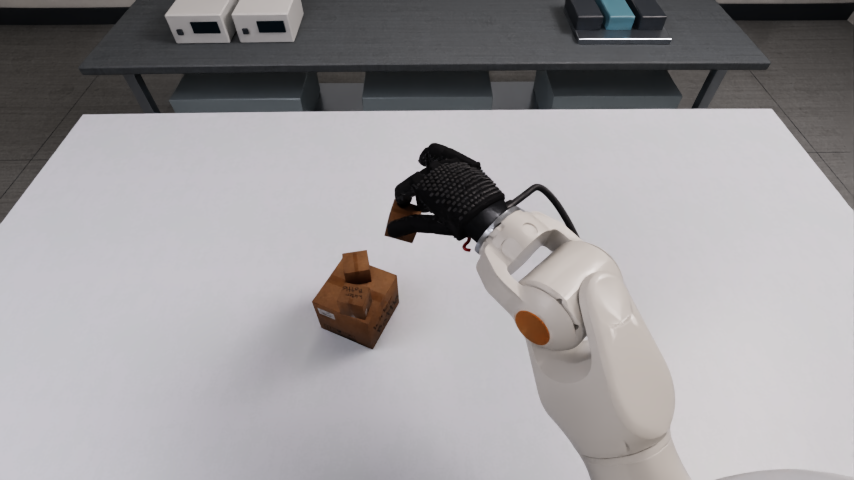}{Stack Boxes}
  {Arrange the four package boxes into four distinct height layers with the large package box on the bottom layer, then keep the hand from touching the boxes for 0.5 seconds.}


\DexProfileRow{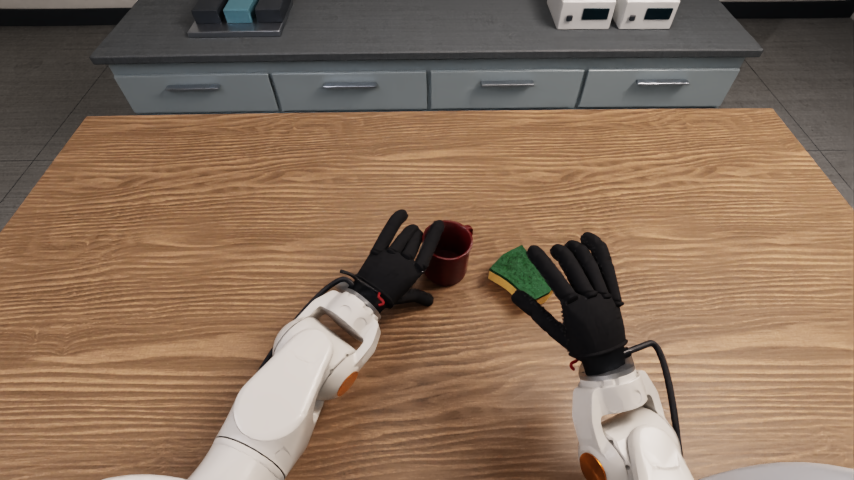}{Compress Sponge into Mug}
  {Use the left hand to stabilize the YCB mug. Sponge 3 uses an isotropic soft-foam material, so clamp it from any reachable pair of opposing sides and compress it along X, Y, or Z; close slowly until it is compressed, keep the pinch while lifting it edge-on through the mug rim until it is mostly inside, then release both objects and leave the mug upright and stable.}

\DexProfileRow{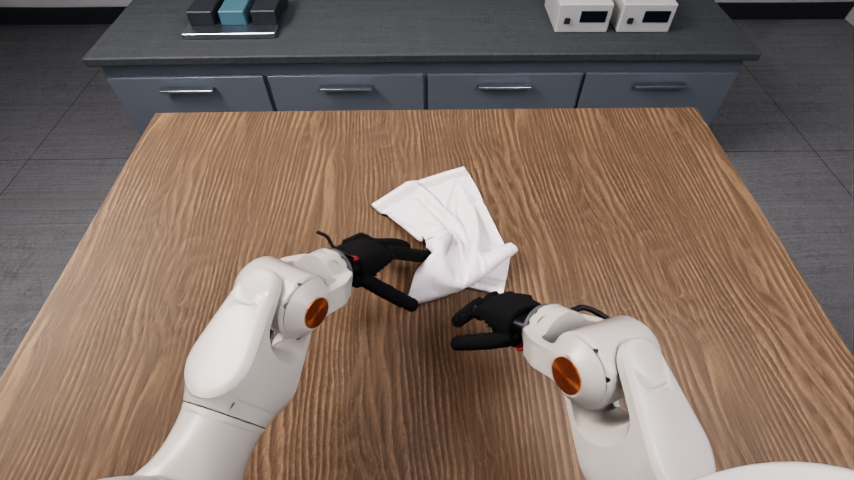}{Fold Towel Twice}
  {Fold the flat towel in half twice, in any directions. The two folds may be in the same direction or different directions. Release the folded towel and leave it stable on the table.}

\DexProfileRow{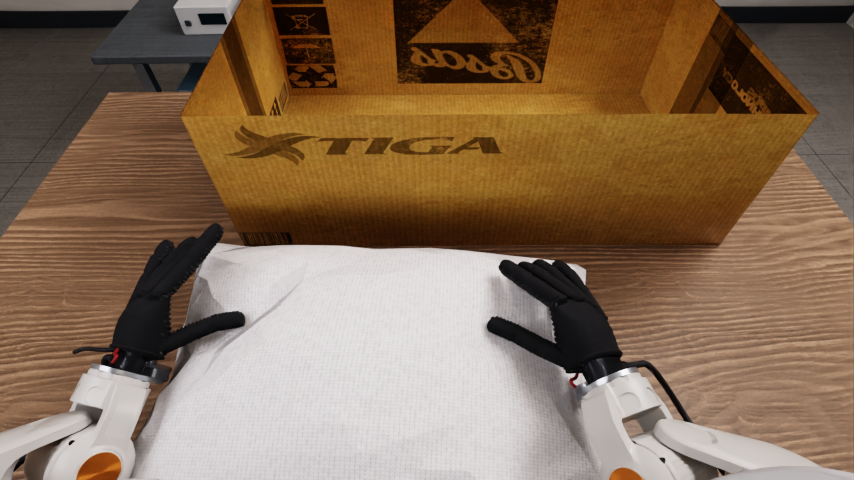}{Pack Pillow}
  {Use both hands to squeeze and lift the foam pillow, lower it into the open carton so it rests inside with most of the pillow below the rim, then release both hands and keep it stable for one second. Tilted placement and raised corners are allowed.}

\end{longtable}
\endgroup

\end{document}